\documentclass[preprint]{elsarticle}
\usepackage{preamble}
\usepackage{commands-private}
\usepackage{booktabs}
\usepackage{multirow}
\usepackage{float}
\usepackage{algpseudocode}
\usepackage{varwidth}
\usepackage{footnote}
\usepackage{amsmath}
\usepackage{amssymb}

\usepackage{soul}

\usepackage{caption}
\usepackage{subcaption}

\algtext*{EndWhile}
\algtext*{EndIf}
\algtext*{EndProcedure}

\usepackage{todonotes}
\usepackage{enumitem,amssymb}
\newlist{todolist}{itemize}{2}
\setlist[todolist]{label=$\square$}

\journal{Artificial Intelligence}

\newcommand{\tick}{\checkmark}
\newcommand{\fznoscar}{fzn-oscar-cbls\xspace}

\algnewcommand{\LineComment}[1]{\State \(\triangleright\) #1} 

\newfloat{algorithm}{t}{lop}

\newcommand{\comment}[1]{}

\begin{document}
\begin{frontmatter}



\title{{\sc Athanor:}\\Local Search over Abstract Constraint Specifications}

\author[sta]{Saad Attieh}
\ead{saad.attieh@gmail.com}
\author[sta]{Nguyen Dang}
\ead{nttd@st-andrews.ac.uk}
\author[dun]{Christopher Jefferson}
\ead{cjefferson001@dundee.ac.uk}
\author[sta]{Ian Miguel}
\ead{ijm@st-andrews.ac.uk}
\author[york]{Peter Nightingale}
\ead{peter.nightingale@york.ac.uk}

\address[sta]{School of Computer Science, University of St Andrews, St Andrews, Fife KY16 9SX, UK}
\address[dun]{School of Science and Engineering, University of Dundee, Dundee, DD1 4HN, UK}

\address[york]{Department of Computer Science, University of York, Heslington, York YO10 5GH, UK}

\begin{abstract}
Local search is a common method for solving combinatorial optimisation problems. We focus on general-purpose local search solvers that accept as input a constraint model --- a declarative description of a problem consisting of a set of decision variables under a set of constraints. Existing approaches typically take as input models written in solver-independent constraint modelling languages like MiniZinc. The {\sc Athanor} solver we describe herein differs in that it begins from a specification of a problem in the abstract constraint specification language \essence, which allows problems to be described without commitment to low-level modelling decisions through its support for a rich set of abstract types. The advantage of proceeding from \essence is that the structure apparent in a concise, abstract specification of a problem can be exploited to generate high quality neighbourhoods automatically, avoiding the difficult task of identifying that structure in an equivalent constraint model. Based on the twin benefits of neighbourhoods derived from high level types and the scalability derived by searching directly over those types, our empirical results demonstrate strong performance in practice relative to existing solution methods.


\end{abstract}

\end{frontmatter}

\section{Introduction}
\label{sec:introduction}

Local search \cite{hoos2004stochastic} is a common method for solving combinatorial optimisation problems. Typically, it operates by generating an initial assignment to the decision variables in a problem and then iteratively modifies this assignment to improve an objective through a sequence of \emph{moves}, selected from a \emph{neighbourhood} of assignments reachable from the current assignment. This approach trades completeness for the ability to make rapid improvements to the objective, and often finds good solutions more quickly than systematic search procedures \cite{FERNANDES2021107592}.

We focus herein on general-purpose local search solvers that accept as input a constraint model: a declarative problem description consisting of a set of decision variables under a set of constraints. This is a general, flexible approach in contrast to local search procedures specialised to individual problems (e.g.,~\cite{FERNANDES2021107592,arnold2019knowledge,ceschia2020solving}). 

Existing approaches typically accept models written in solver-independent constraint modelling languages like MiniZinc~\cite{nethercote07}. The recently-proposed Structured Neighbourhood Search (SNS)~\cite{SNS-IJCAI18}, and the {\sc Athanor} solver we describe in this paper (which was first proposed in~\cite{attieh2019athanor}), differ in that they begin from a problem specification in the abstract constraint specification language \essence~\cite{frisch2005essence,frisch2007design,frisch2008ssence}. \essence allows problems to be described without commitment to low-level modelling decisions through its support for a rich set of abstract types, such as sets, multi-sets, sequences and relations, each of which can be nested arbitrarily (\emt|set of partition|, \emt|multi-set of sequence of tuple|, and so on). \Cref{fig:sonet-spec} presents an example \essence specification of the Synchronous Optical Networking Problem (SONET)~\cite{smith2005symmetry}, where a set of nodes must be connected via a set of fibre-optic rings.

\begin{figure}
\begin{lstlisting}
$ Parameters to the problem:
given nNodes, nRings, capacity : int(1..)
letting Nodes be domain int(1..nNodes)
given demand : set of set (size 2) of Nodes

$ Decision variable for search:
find network : set (maxSize nRings) of
               set (minSize 2, maxSize capacity) of Nodes
$ The problem objective:
minimising sum ring in network . |ring|
$ The problem constraints:
such that  forAll pair in demand .
  exists ring in network .
    pair subsetEq ring
\end{lstlisting}
\caption{Synchronous Optical Networking in \essence. The main parameter of the problem is \emt|demand|, a set of pairs of nodes to be connected to each other. A {\em single} highly structured decision variable \emt|network| (\emt|set of set of int|) suffices to model the fibre-optic rings, each inner \emt|set of int|s representing  a set of connections from a ring to nodes. Rings have a limited number of connections that they can support, hence the \emt|maxSize| attribute on the inner set.  This is important as a pair of nodes in the demand are only connected to each other if they are both on the same ring.  Nodes may connect to multiple rings to facilitate more connections, however the objective is to minimise the total number of connections. The specification can be summarised in mathematical notation as follows.\\
\textbf{given} \(\mathit{nNodes},\:\mathit{nRings},\:\mathit{capacity},\:\mathit{demand}\)\\
\textbf{find} \(\mathit{network}\:\\
\mathbf{where}\:|\mathit{network}|\leq \mathit{nRings}\: \wedge\: \forall r\in \mathit{network}.\:2\leq |r|\leq \mathit{capacity}\:\wedge\: r\subseteq \{1\ldots \mathit{nNodes}\}\)\\
\textbf{minimising} \(\sum_{\mathit{ring}\in \mathit{network}} |\mathit{ring}|\)\\
\textbf{such that} \(\forall \mathit{pair} \in \mathit{demand}. \: \exists \mathit{ring}\in \mathit{network}. \: \mathit{pair}\subseteq \mathit{ring}\)
}
\label{fig:sonet-spec}
\end{figure}

Our approach is to proceed from \essence, where the structure apparent in a concise, abstract problem specification can be exploited to generate high quality \emph{neighbourhood structures}~\cite{sorensen2008multiple} automatically, avoiding the challenging task of identifying that structure in an equivalent constraint model. A neighbourhood structure is a procedure that takes a current assignment to a variable in a specification and applies a random transformation to it, generating a new assignment to the variable from a neighbourhood. \athanor has a set of neighbourhood {\em templates}, which express an abstract concept, such as adding values to a set, or moving a value from one set to another. \athanor takes an abstract problem specification and uses the neighbourhood templates to create neighbourhood structures. In the SONET specification the decision variable is a set of sets. Three illustrative neighbourhood structures produced for this structure by \athanor are:

\begin{itemize}
\item Select a set and remove an element: improves the objective, removes unnecessary connections to the rings.
\item Select a set and add an element: helps with satisfying the constraints, if a demand is not currently met.
\item Select two sets, move an element from one to the other: moves connections to where they may be better used, i.e.\ to connect to more nodes.
\end{itemize}

Given an \essence specification of a problem class and a set of neighbourhood structures, there are two options for solving. One option is SNS~\cite{SNS-IJCAI18}, which uses {\sc Conjure} \cite{conjure-aij} to refine the original specification and neighbourhood structures into a constraint model and solves with an existing constraint solver. Alternatively, we can perform a local search directly on the augmented {\sc Essence} specification. This is the novel approach taken by {\sc Athanor}, described in this paper. Specifically, \athanor supports the direct instantiation of abstract variable types such as \emt|set|, \emt|sequence|, \emt|partition|, \emt|function|, without decomposing the variables and constraints posted on them into low-level representations. For several problem classes, \athanor's high-level representations allow the solver to scale far better than other solvers, solving instances that are infeasibly large for competing solvers.  As explained in \Cref{sec:value-representation}, this is because \athanor dynamically manages memory during the search to support high-level variable types whose cardinality can vary significantly depending on the value assigned. This is in contrast to low-level constraint solvers, which
typically employ a more rigid, conservative approach that immediately requires a substantial memory commitment.

To illustrate, reconsider the single decision variable in the SONET example:
\begin{lstlisting}
find network : set (maxSize nRings) of
               set (minSize 2, maxSize capacity) of Nodes
\end{lstlisting}

While \emt|maxSize capacity| on the inner sets is enforcing a problem constraint (the capacity of each ring), \emt|maxSize nRings| on the outer set is only required because otherwise the deduced maximum cardinality of the outer set \emt|network| is:

$$\sum \limits_{i=2}^{\mathrm{capacity}} \binom{n}{i},$$

Where $n$ is the size of the domain \emt|Nodes|.  Hence, with a capacity of $32$ and a \emt|Nodes| domain of size $65$, the maximum cardinality of \emt|network| is $2^{64}-66$. Existing low-level solvers require this abstract decision variable to be modelled as a constrained collection of more primitive variables, which must be done conservatively so as to be able to represent the largest cardinality value of the set of sets. Clearly, $2^{64}$ is larger than is practically feasible, hence the artificial bound on the outer set. \athanor does not require this bound: it supports the instantiation of the set of sets directly, and is able to assign to it values of different cardinality efficiently (although the worst case still remains). Moreover, our experiments show that even when a maximum cardinality is given, \athanor still scales better to larger instances.

Through the twin benefits of neighbourhood structures derived from high level types and the scalability derived by searching directly over those types, our empirical results demonstrate strong performance relative to existing solution methods.

Our major contributions are as follows:
\begin{itemize}
\item The automatic derivation of neighbourhood structures from the types of the abstract decision variables in \essence.
\item Representations that scale with the size of the value rather than the upper bound of the domain of a variable.
\end{itemize}
To enable the above to be implemented in a practical solver, we also make the following contributions:
\begin{itemize}
\item Time and space-efficient incremental evaluation of expressions that quantify over values of variable size;
\item Efficient generation of random values drawn from extremely large domains;
\item Experimental evaluation of the proposed solver in comparison to other state-of-the-art methods, including local search and systematic solvers. 
\end{itemize}

\section{\label{sec:Overview}An Overview of {\sc Essence} and {\sc Athanor}}

This section provides an overview of the architecture and operation of the \athanor solver, before full details are given in subsequent sections. We also briefly describe the types and structure of the \essence language on which \athanor operates.

\subsection{Background: {\sc Essence}}

We begin with a brief overview of the \essence abstract constraint specification language, from which the neighbourhood templates and neighbourhood structures employed by \athanor are derived (this derivation is discussed in \Cref{neighbourhoods-detail}). \essence was originally conceived as a means of capturing a formal description of a combinatorial problem without committing to a concrete model suitable for input to a particular solving formalism, such as constraint programming or SAT. Hence, the types supported by \essence (summarised in \Cref{tab:EssenceTypes}) are designed to allow a specification to be given in terms of the combinatorial structure of the problem.
We distinguish between atomic types and compound types (as shown in \Cref{tab:EssenceTypes}). There is also a distinction in \essence between \textit{abstract} types (set, multiset, sequence, relation, function, and partition) and \textit{concrete} types (the atomic types, matrix, tuple, and record) \cite{conjure-aij}.

\begin{table}[t]
\small
\centering
\begin{tabular}{lp{0.77\linewidth}}
\toprule
\multicolumn{2}{c}{\textit{Atomic Types}} \\
Boolean    &  A simple Boolean type.\\
Integer    &  Any subset of the integers, specified as a set of intervals.\\
Enumerated &  A type with a finite set of named values.\\
Unnamed    &  A type with a finite set of unnamed values. The values of an unnamed type cannot be referenced in the specification.\\
\cmidrule(lr){1-2}
\multicolumn{2}{c}{\textit{Compound Types}} \\
Set        &  Set of variable size (unless annotated otherwise) of any type \(\tau\). \\
Multiset   &  Similar to set but allows multiple occurrences of values.\\
Sequence   &  A sequence of variable length (an upper bound must be provided) of any type \(\tau\).\\
Relation   &  A relation of arbitrary arity over any types \(\tau_1, \tau_2, \ldots\)\\
Function   &  A function from any type \(\tau_1\) to any other type \(\tau_2\). The function is partial by default but may be made total (or surjective, injective, etc.) with an annotation.\\
Partition  &  A partition from any finite type \(\tau\). A partition may be annotated to control the number and sizes of parts.\\
Matrix     &  A matrix with any number of dimensions, containing any type \(\tau\) and indexed by integer or enumerated types in each dimension.\\
Tuple      &  A container with fields accessed by an integer index, each of which may be of any type. \\
Record     &  A container with named fields, each of which may be of any type.\\
\bottomrule
\end{tabular}
\caption{\label{tab:EssenceTypes}The \essence type constructors supported by \athanor, which may be annotated to impose further structure, such as a total function or maximum cardinality of a set. \essence supports common operators on these types, such as set intersection and union, and arbitrary nesting, such as set of sets of integers. Decision variables must have a finite domain.}
\end{table}

An \essence specification (e.g.\ \Cref{fig:sonet-spec}) comprises formal parameters (\texttt{given}), which may themselves be constrained (\texttt{where}); the combinatorial objects to be found (\texttt{find}); constraints the objects must satisfy (\texttt{such that}); identifiers declared (\texttt{letting}); and an optional objective function (\texttt{min/maximising}). The input to \athanor is an \essence specification of a problem class and values for its formal parameters to derive a particular problem instance to solve.

\subsection{Overview of \athanor{}}
\label{subsec:overview}

Given an \essence specification, \athanor begins by constructing two abstract syntax trees (ASTs).  The first tree encodes the constraints in a problem, the second encodes the objective, if present. Each node in the tree represents an \essence expression, with each leaf being a reference to a variable in the problem (or a constant) and their ancestors representing the operators. Subsequently, \athanor generates a value at random for each of the decision variables according to some restrictions (described in \Cref{sec:neighbourhoods-generating-random-values}). The operators (ancestor nodes) are then evaluated all the way up to the roots of the two trees, yielding a Boolean and an integer value for the constraint and objective trees respectively.

Furthermore, every Boolean expression is associated with an integer termed the {\em constraint violation}.  When the expression evaluates to true, the constraint violation is 0, otherwise it takes a positive value. The constraint violation is a heuristic for the magnitude of the change necessary to the operand values of an expression such that the expression evaluates to true.  In addition, each variable in the problem is associated with a \emph{variable violation}, a heuristic measure of the likelihood of a variable being the cause of violated constraints in the problem.  In the case of a structured type, such as a \emt|set of set of int|, variable violations are associated with the top level structure (here the outermost \texttt{set}), each of its elements, and their elements and so on.  This allows the solver to infer whether a whole or part of a structure is the cause of constraint violations, as described in \Cref{sec:violation_counts}.

\begin{figure}[t]
\centering
\includegraphics[width=0.6\textwidth]{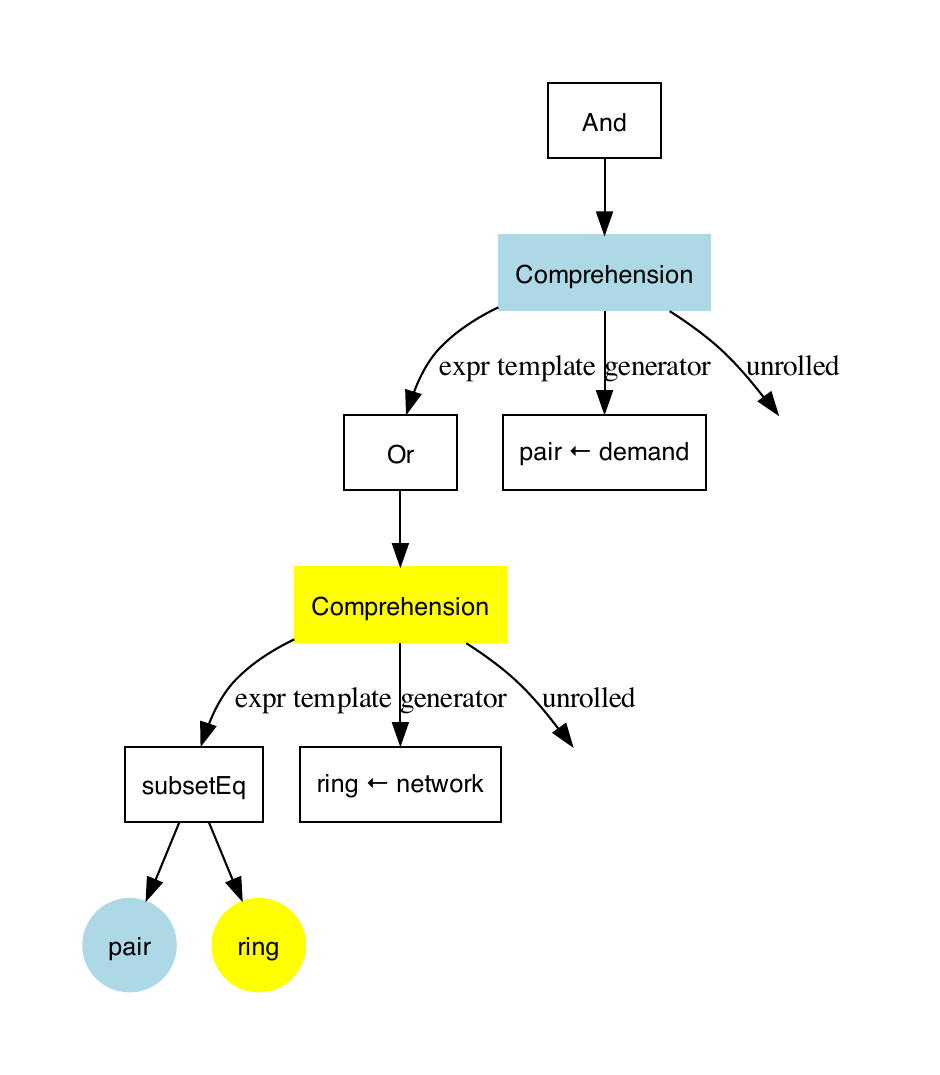}
\caption{\label{fig:SONETASTUnevaled}SONET Abstract Syntax Tree: Unevaluated.}
\end{figure}

\begin{figure}[t]
\centering
\includegraphics[width=0.9\textwidth]{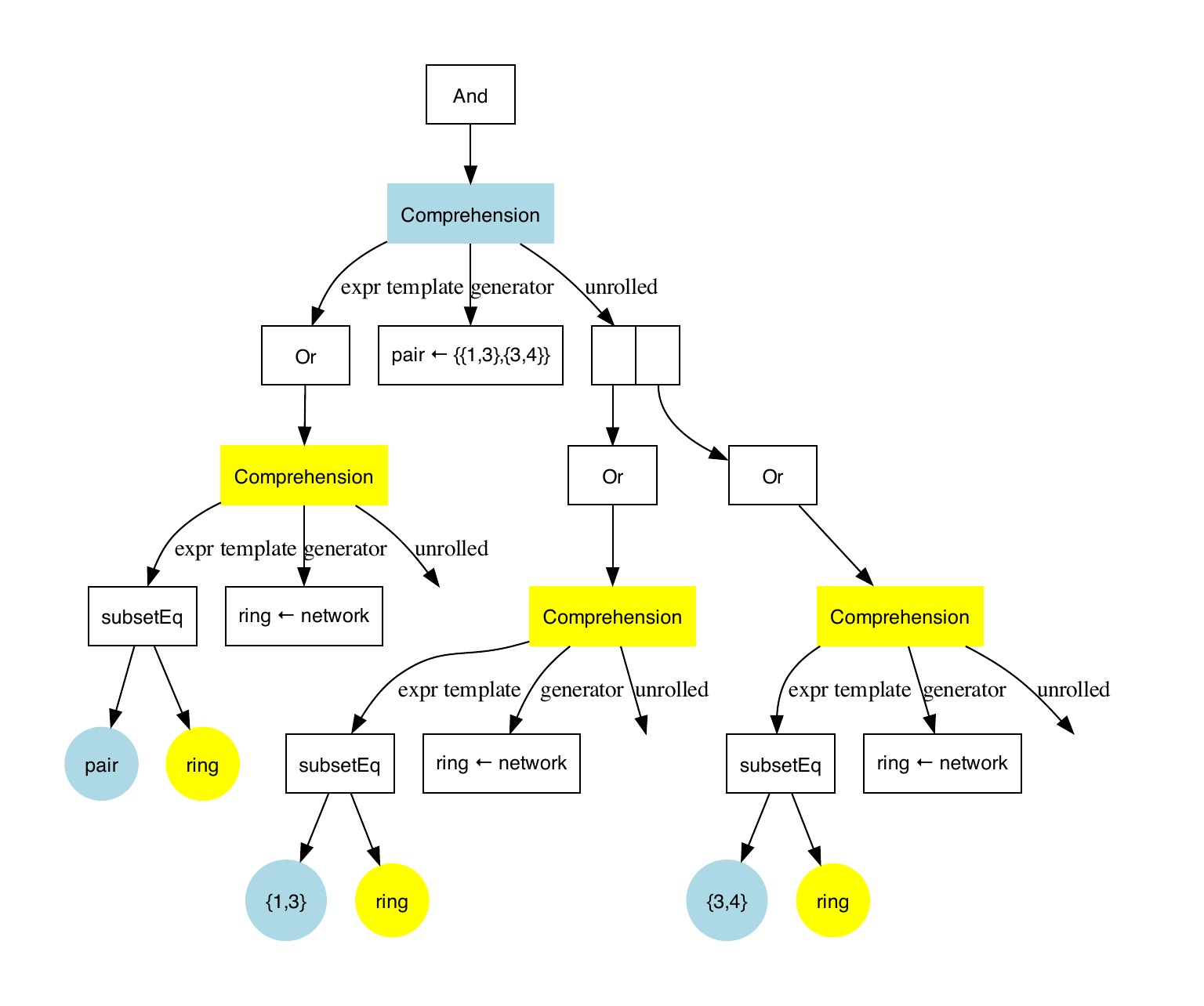}
\caption{\label{fig:SONETASTEvaled1}SONET Abstract Syntax Tree: Outer comprehension (with variable \texttt{pair}) unrolled.}
\end{figure}

\begin{figure}[t]
\centering
\includegraphics[width=\textwidth]{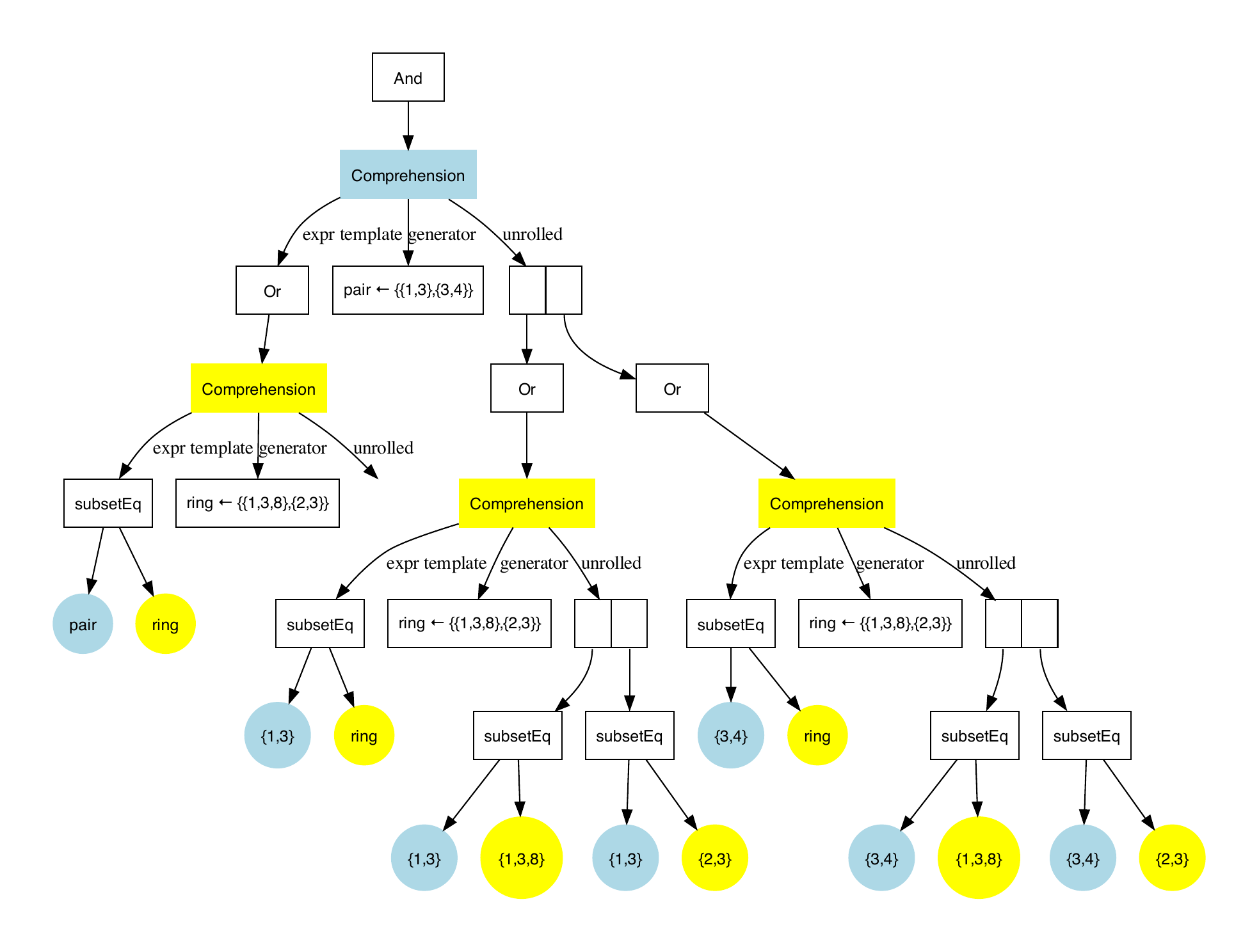}
\caption{\label{fig:SONETASTEvaled2}SONET Abstract Syntax Tree: Inner comprehension (with variable \texttt{ring}) unrolled. }
\end{figure}

\Cref{fig:SONETASTUnevaled,fig:SONETASTEvaled1,fig:SONETASTEvaled2} illustrate the initialisation of the constraint tree for the SONET specification (\Cref{fig:sonet-spec}). The \texttt{forAll} quantifier is represented with an \texttt{and} function, applied to a list generated by a comprehension. Similarly, the \texttt{exists} quantifier becomes an \texttt{or} function containing a comprehension.
\Cref{fig:SONETASTUnevaled} presents the abstract syntax tree prior to the unrolling of the two comprehensions. Each comprehension has three children: the expression template (labelled \texttt{expr template}), the generator, and the unrolled list (initially empty). Comprehensions may also produce sets, but in this example both comprehensions produce lists. The generator supplies a sequence of values to substitute one by one into the expression template to create the elements of the unrolled list.
For this simple example, we consider two demand pairs, $\{1, 3\}$ and $\{3, 4\}$. \Cref{fig:SONETASTEvaled1} shows the constraint tree after the top comprehension has been unrolled with respect to these two pairs, forming a list with two elements.
Finally, a random value for the network variable is generated, $\{\{1, 3, 8\}, \{2, 3\}\}$, and used to unroll the inner comprehension, as presented in \Cref{fig:SONETASTEvaled2}. At this point, the AST can be evaluated, and it evaluates to \texttt{false}. The expression template (\texttt{expr template}) child of a comprehension is always retained in the AST so that the \texttt{unrolled} child can be updated in the event that the \texttt{generator} is changed. 

Unrolling the syntax trees dynamically, as illustrated here, is vital for the scalability of \athanor{}: the sizes of the trees scale linearly with the actual sizes of the values of decision variables, not their largest possible sizes. The values themselves are also represented compactly, as described in \Cref{sec:value-representation}.

After initialisation, the tree structures are used to track subsequent changes to the leaf nodes (\Cref{sec:incremental-evaluation}). \athanor employs two ASTs for both flexibility and performance. Its search strategies benefit from being able to make decisions dependent on access to the root of each tree individually, such as permitting constraint violations while focusing on improving the objective value.

Before commencing the search, \athanor constructs all of the neighbourhood structures that will be used during the search process to generate new assignments within neighbourhoods. These neighbourhood structures are created by combining a set of neighbourhood templates, which represent high-level neighbourhoods. Examples of neighbourhood templates include adding a value to a set, or moving a value from one set to another.

Following initialisation, \athanor iteratively modifies the current assignment in an attempt both to reduce the constraint violation count and to improve the objective value. This is achieved by using the previously constructed neighbourhood structures, which select new values for some variables within a neighbourhood. The two ASTs are used to evaluate the impact of the changes made. Neighbourhood structures are derived directly from the \essence specification (as described in \Cref{neighbourhoods-detail}). \athanor makes use of several search procedures, as described in \Cref{sec:search-procedures}.
The experimental evaluation of \athanor is presented in \Cref{sec:experiments} and \Cref{sec:experiment-globcons}.

\section{Related Work}\label{sec:related-work}

There are a number of existing approaches related to our work, which we summarise in this section.

{\sc Localizer} \cite{michel2000localizer} introduced a modelling language for the manual specification of a local search procedure. One of its key innovations was the concept of an {\em invariant} (also known as one-way constraints in the iOpt toolkit \cite{voudouris2001iopt}), an expression that describes how to compute the values of one set of variables from another. This allows the separation of the problem variables into those over which search is performed, and those whose values are updated incrementally and efficiently via the invariants. {\sc Comet} \cite{hentenryck2009constraint,michel2018constraint} developed these ideas further, providing a full object-oriented programming language that supports declarative modelling followed by local search for solutions.
\textsc{Comet} is able to synthesize local search procedures from a declarative constraint-based local search (CBLS) model~\cite{Van-Hentenryck2007:synthesis} where constraints are annotated as hard (i.e.\ remains satisfied during search) or soft (may be violated during search) and a metaheuristic such as Tabu Search is specified by the user. Neighbourhood structures are synthesized from the constraint semantics (for example, a hard allDifferent constraint may lead to a neighbourhood structure that swaps the values of two variables in the allDifferent).

Several more CBLS solvers~\cite{hentenryck2009constraint,michel2018constraint} have since been developed. The Kangaroo solver~\cite{newton2011kangaroo} employs a lazy approach, increasing efficiency by propagating invariants only as necessary. OscaR-CBLS~\cite{de2013oscar} is an open-source solver that adopts a similar partial propagation scheme, making neighbourhood selections based on the objective value only. Bj{\"o}rdal et al.~\cite{bjordal15,bjordal2021declarative} provide an automated compilation method from the solver-independent constraint modelling language MiniZinc~\cite{nethercote07} to OscaR-CBLS in their solver \fznoscar, performing an analysis of the MiniZinc model to determine invariant structure and which constraints can be implicitly satisfied through neighbourhood structure, as well as synthesising a search strategy. This work was further developed to extend MiniZinc to support the declarative specification of neighbourhoods \cite{bjordal2018declarative}. Yuck\footnote{https://github.com/informarte/yuck} also implements a set of constraint-specific neighbourhoods to support the MiniZinc language and a search strategy based upon simulated annealing. iZplus\footnote{https://www.minizinc.org/challenge2014/description\_izplus.txt} is something of a hybrid, using systematic search to find the first solution to an optimisation problem and then local search to find better solutions. It implements CBLS for decision problems, either alone or in parallel with systematic search. LocalSolver~\cite{benoist2011localsolver} was the basis for the commercial Hexaly Optimizer.\footnote{https://www.hexaly.com} Its emphasis is on allowing the user to focus on modelling while treating the solving process as a black box. Structured Neighbourhood Search~\cite{SNS-IJCAI18} (SNS) generates neighbourhood structures from \essence specifications, as \athanor{} does, then applies \conjure~\cite{conjure-aij} and \savilerow~\cite{sr-journal-17} to refine them (alongside the model) into the input language of a backtracking constraint solver. SNS uses an adapted version of the \minion solver~\cite{gent2006minion} that applies the neighbourhood structures in a similar way to Large Neighbourhood Search (described below). SNS contrasts with the work described in this paper as \athanor directly operates on abstract \essence variables. This leads to a more scalable approach, especially due to the dynamic creation and deletion of values and constraints (as illustrated in~\Cref{subsec:overview}).

{\sc Comet} was extended to support graph variables with specialised neighbourhood structures to search over them \cite{pham2012ls}. Similarly, {\AA}gren et al. \cite{aagren2005set} and Bj{\"o}rdal \cite{bjordalString} showed how to introduce set and string variables in a local search setting respectively. However, these cannot be composed with other abstract types such as those offered by \essence.

Large neighbourhood search (LNS)~\cite{shaw98} is a general framework for solving constrained optimisation problems which is built on a systematic solver. Systematic search is typically used to find an initial feasible solution. In each subsequent iteration, a subset of the decision variables is selected to be unassigned (relaxed) and searched once more by the systematic solver for an assignment with an improved objective value. 
The ability to exhaustively search parts of the search space allows these solvers to perform well on highly constrained and highly structured problems. 
In the context of LNS, a neighbourhood structure is responsible for selecting a set of variables to relax. Highly effective specialised LNS neighbourhood structures have been proposed for many problem classes (such as vehicle routing~\cite{shaw98}); however, developing such neighbourhood structures requires time and in-depth understanding of the problem, motivating research into general-purpose LNS neighbourhood structures. 

Propagation-guided large neighbourhood search (LNS-PG)~\cite{DBLP:conf/cp/PerronSF04} is a general-purpose LNS method that exploits the propagation process of a CP solver to inform its choice of variables to relax. The rationale behind LNS-PG is that the selected variables should form a closely linked sub-part of the problem instance, where the strength of a link between two variables is measured by assigning one of the variables and recording the effect of propagation on the other variable. 
Explanation-based large neighbourhood search (LNS-EB) \cite{DBLP:journals/constraints/PrudhommeLJ14} uses the explanation mechanism of a learning CP solver to find sets of variables that are linked to improving the objective. 
Two methods of LNS-EB were proposed with promising results. 
LNS-PG and LNS-EB are both able to derive neighbourhoods automatically for any problem class, in common with \athanor and SNS, however they do not have access to the high-level structure of \essence.

\section{Neighbourhoods} \label{neighbourhoods-detail}

A {\em neighbourhood structure}~\cite{sorensen2008multiple} is a procedure that takes an assignment to a decision variable and applies a transformation to it, such as randomly reassigning an integer, adding a value to a set, or moving elements between parts of a partition. \athanor has a set of {\em neighbourhood templates}, which are the building blocks from which neighbourhood structures are created. A {\em neighbourhood} is the set of assignments reachable from the current assignment to a decision variable by applying a neighbourhood structure. At each iteration of the search, a neighbourhood structure is selected, and the structure is used to generate a move from within a neighbourhood of the current assignment. The sampling process is described in \Cref{subset:Neighbourhood-Templates}.

Neighbourhoods must respect the types of variables, for example given the value $\{1,3,5\}$ for a variable of type \emt|set (size 3)  of int(1..6)|, we can change $1$ to $2$, $4$ or $6$, but not to $3$ or $5$, nor can we add or remove values from the set, as the set is fixed size. More complex types allow for more interesting neighbourhoods. For example, given a value $\{\{1,2\},\{2,3,4\}\}$ for a variable of type \emt|set of set of int(1..5)|, we could change one of the values inside one of the inner sets, create or remove an entire set, or move values between sets -- as long as we never create an invalid value for our type.

In the next sub-section, we describe the set of neighbourhood templates and how they are combined into neighbourhood structures.

\subsection{Neighbourhood Templates}
\label{subset:Neighbourhood-Templates}

\athanor's neighbourhood templates are inspired by neighbourhood generation in Structured Neighbourhood Search~\cite{SNS-IJCAI18}. Since each neighbourhood is directly linked to a variable type and domain, \athanor's neighbourhoods are not allowed to violate type invariants, such as the uniqueness of elements in a set. As \athanor's neighbourhoods cannot violate type invariants, we reduce the time spent finding assignments to highly structured variables (e.g. \emt|set of partition (regular, numParts 3)|) and can instead focus on satisfying problem constraints or improving the objective.

Where neighbourhood structures can very cheaply check if they would violate a type invariant, for example adding a value to a set which is already maximum size, they do so. However, given the large number of type invariants in \essence we do not require all neighbourhood structures maintain all type invariants. Instead, neighbourhood structures generate a neighbourhood move, which is reverted if any type invariant is violated. If a neighbourhood structure fails to generate a valid neighbourhood move after a fixed number of attempts (currently set to 50), the neighbourhood structure is abandoned and another neighbourhood structure is chosen.

Neighbourhood templates are divided into four categories. We will discuss these categories in turn, using the neighbourhood structures generated for the SONET problem as examples.

\subsubsection{Atomic Neighbourhood Templates}

The first type of neighbourhood templates are \emph{atomic} neighbourhood templates, presented in Table \ref{tab:DirectNeighbourhoods1}. These apply only to a single value of one of the primitive types. For example, \cxxItem{intAssignRandom} reassigns a variable of type \cxxItem{int} to another value in its domain. None of these apply to the SONET problem, as the SONET problem contains no variables of type \cxxItem{bool}, \cxxItem{enum}, or \cxxItem{int}. However, these will be used as building blocks for neighbourhood structures used in the SONET problem later.  Integers have a direct neighbourhood template that restricts the magnitude of the change to be no greater than the violation attributed to that variable (\cxxItem{intAssignRandomFromViolation}). No other type currently has a neighbourhood template that makes use of the violation. We do not provide a general \cxxItem{assignRandom} for any higher-level types, instead considering neighbourhoods that make smaller, structured changes.

\subsubsection{Direct Neighbourhood Templates}

The second type of neighbourhood templates are \emph{direct} templates, presented in Table  \ref{tab:DirectNeighbourhoods2}. These are only dependent on the outermost part of a type.  For example, the \cxxItem{setAdd}\footnote{For clarity, the template name in the text is formed from the name given in \Cref{tab:DirectNeighbourhoods1} or \ref{tab:DirectNeighbourhoods2}, prefixed with the type on which it is instantiated.} neighbourhood template (Table~\ref{tab:DirectNeighbourhoods2}) can only be applied to a set, but can be instantiated for sets of any type. The \cxxItem{Add} templates use the existing infrastructure for creating random values, discussed in \Cref{sec:neighbourhoods-generating-random-values}.   The \emt|relation| type in \essence is equivalent to a \emt|set of tuple|, so variables of type \emt|relation| use the \emt|set| neighbourhoods. 

The attributes of variable domains are considered during the creation of neighbourhood structures. For example, neighbourhood templates that alter a set's cardinality are not used to create neighbourhood structures on a set variable whose domain has a fixed size attribute. Similarly, if a sequence variable has a domain with the \emt|injective| attribute (meaning all values in the sequence are distinct), the neighbourhood structures created for that variable are limited to those that retain injectivity.

In SONET, the set \cxxItem{Add} and \cxxItem{Remove} neighbourhood templates are used to add and remove elements from \cxxItem{network}. These added and removed values are of type \cxxItem{set (minSize 2, maxSize capacity)}.

\subsubsection{Higher-order Neighbourhood Templates}

The third type of neighbourhood template is \textit{higher-order}, as listed in~\Cref{table:lift_neighbourhood}. These take another neighbourhood template and apply it to one or more elements of a container, such as a \emt|set| or \emt|sequence|. \cxxItem{LiftSingle} applies a neighbourhood template on a type $T$ to a single element of a collection of $T$s. For example, in SONET, we can lift \cxxItem{SetAdd} to create \cxxItem{SetLiftSingle\_SetAdd}, which adds an element to a set contained in \cxxItem{network}. We can lift multiple times, so we can lift \cxxItem{intAssignRandom} twice, thus creating a neighbourhood structure that can modify any integer contained in a set that is in \cxxItem{network}.

Neighbourhoods can be lifted for sets, multisets, sequences, and the defined set and the range of a function. Neighbourhoods are not currently lifted for partitions, because none of our currently implemented neighbourhood templates would lift to useful, valid neighbourhood structures on partitions. Neighbourhood structures generated by lifting a \emt|set| illustrate how difficult it is to ensure neighbourhood structures satisfy all type invariants. For example, changing one value in a \emt|set| to be equal to another value in the same \emt|set|, where the \emt|set| has fixed size, would violate the type invariant of the \emt|set|. Rather than require every neighbourhood handle this problem, \athanor runs the neighbourhood structure, then as a final step checks if the set's type invariants are violated -- if so the move is rejected and an alternative move is generated.

\begin{table}[t!]
\centering
\begin{tabular}{p{0.40\columnwidth}p{0.53\columnwidth}}
\toprule
\texttt{\bf bool}\\
\texttt{boolReassign} &
Reassign the variable to the other value in its domain.\\ \cmidrule(lr){1-2}
\texttt{\bf enum}\\
\texttt{enumAssignRandom} &
Randomly reassign the variable to a different value in its domain.\\ \cmidrule(lr){1-2}
\texttt{\bf int}\\
\texttt{intAssignRandom} & Randomly reassign the variable to a different value in its domain. \\
\texttt{intAssignRandom\allowbreak{}From\allowbreak{}Violation} & Assign a value uniformly at random in the range $i \pm v$,  $i$ is the current value of the integer, $v$ is a violation attributed to that variable. \\
\bottomrule
\end{tabular}
\caption{\label{tab:DirectNeighbourhoods1}A complete list of \textbf{atomic neighbourhood templates} in \athanor.}
\end{table}

\begin{table}[t!]
\centering
\begin{tabular}{lp{0.57\columnwidth}}
\toprule
\texttt{\bf set, multiset}\\
\texttt{Remove} & Remove a random element.\\
\texttt{Add} & Add a random element.\\   \cmidrule(lr){1-2}
\texttt{\bf sequence}\\
\texttt{Remove} & Remove a random element from the sequence. \\
\texttt{Add} & Add a random element to the sequence at a randomly chosen position.\\
\texttt{ReverseSub} & Reverse a contiguous subsequence of a random size. \\
\texttt{PositionsSwap} & Swap the position of a random element with another. \\
\texttt{ReassignSub} & Randomly reassign a contiguous subsequence. \\\cmidrule(lr){1-2}
\texttt{\bf function}\\
\texttt{Remove} & Remove a random mapping from the function.\\
\texttt{Add} & Add a random mapping to the function.\\
\texttt{UnifyImages} & Randomly choose two points in the function with distinct images, assign the image of one point to the image of the other. \\
\texttt{SplitImages} & Randomly choose two mappings with equal images, randomly reassign one of the images.\\
\texttt{Swap} & Swap the images of two randomly chosen mappings in the function.\\
\texttt{SwapAlongAxis} & Special case of Swap: with functions that map from a tuple to any type, swap the images of two tuples \(\tau_1\) and \(\tau_2\) where \(\tau_1\) and \(\tau_2\) differ in exactly one place. \\
\cmidrule(lr){1-2}
\texttt{\bf partition}\\
\texttt{MoveParts} & Randomly select two parts $p_1$ and $p_2$ and an element $e_1 \in p_1$ and move \(e_1\) into \(p_2\). \\
\texttt{SwapParts} & Randomly select two parts $p_1$ and $p_2$ and elements $e_1 \in p_1$ and $e_2 \in p_2$ and swap the elements such that $e_1 \in p_2$ and $e_2 \in p_1$. \\
\texttt{MergeParts} & Merge two randomly-selected parts into one.\\
\texttt{SplitPart} & Selects one part \(p_1\) at random, creates a new part \(p_2\) and distributes the elements of \(p_1\) between the two at random, while respecting part size attributes. \\
\bottomrule
\end{tabular}
\caption{\label{tab:DirectNeighbourhoods2}A complete list of \textbf{direct neighbourhood templates} in \athanor.}
\end{table}

\begin{table}
\centering
\begin{tabular}{lp{0.70\columnwidth}}
\toprule
\texttt{LiftSingle} & Randomly select an item from the structure and apply \(t\), keeping all other elements the same.\\
\texttt{LiftMultiple} & Randomly select multiple items from the structure and apply \(t\), keeping all other elements the same. \\
\bottomrule
\end{tabular}
\caption{\label{tab:HigherOrderNeighbourhoods}A complete list of \textbf{higher-order neighbourhood templates}, parameterised with a neighbourhood template $t$.
For \texttt{LiftMultiple}, \(t\) must be a synchronised neighbourhood template.
The \texttt{LiftMultiple} neighbourhood template selects as many elements as required by $t$. At present, this is always two elements.}
\label{table:lift_neighbourhood}
\end{table}

\subsubsection{Synchronised Neighbourhood Templates}

The fourth and final kind of neighbourhood templates operate on multiple members of a container at once. These are known as \emph{Synchronised} neighbourhood templates and are presented in \Cref{tab:SynchronisedNeighbourhoods}. For example, the \texttt{setMove} neighbourhood template moves an element from one set to another. Because these neighbourhood templates require two values to operate, they only work when lifted. \cxxItem{liftMultiple} takes a synchronised neighbourhood and lifts it to any of the container types in \athanor.

In SONET, we generate two synchronised lifted neighbourhoods, \\ \cxxItem{SetLiftMultiple\_SetMove}, which takes two sets in \cxxItem{network} and moves an item from one set to another, and \cxxItem{SetLiftMultiple\_SetCrossover}, which takes two sets in \cxxItem{network} and exchanges two elements from the two sets.

\begin{table}[!]
\centering
\begin{tabular}{lp{0.70\columnwidth}}
\toprule
\texttt{\bf set}\\
\texttt{Move} & Move a random element from one set to another.\\
\texttt{Crossover} & Swap randomly chosen elements between two sets.\\ \cmidrule(lr){1-2}
\texttt{\bf sequence}\\
\texttt{Move} & Move a random element from one sequence to a random position in another sequence.\\
\texttt{Crossover} & Exchange elements between two sequences that are at the same randomly chosen index.\\ \cmidrule(lr){1-2}
\texttt{\bf function}\\
\texttt{Crossover} & Given two functions, select a mapping at random from each one such that they have the same preimage. Exchange the images of these two mappings. \\ \bottomrule
\end{tabular}
\caption{\label{tab:SynchronisedNeighbourhoods}A complete list of \textbf{synchronised neighbourhood templates} in \athanor. These are used in conjunction with higher-order templates to operate on multiple parts of an abstract structure simultaneously.}
\end{table}

\subsection{Neighbourhood Structure Creation}

Neighbourhood structures are created once at the beginning of search, and neighbourhood structures are created independently for each variable in the \essence specification. Parameters do not affect neighbourhood structure generation.

For each variable, we start at the top of the type and attempt to instantiate each type of neighbourhood template appropriate for the outer-most type. For each lifted neighbourhood template that applies to the outer-most type, we remove the outer-most type, recursively create all neighbourhood structures for the inner type, and combine these with the lifted neighbourhood template.

This process of generating all neighbourhood structures is very quick, as it only has to operate on the types and domains of each variable in the \essence specification.

Bringing together the neighbourhood templates discussed in this section, we present the full generation of neighbourhood structures for SONET. In SONET there is only a single decision variable, \cxxItem{network}, which (removing \cxx|minSize| and \cxx|maxSize| annotations) has type \cxx|set of set of Nodes|.

\athanor begins by applying direct neighbourhood templates, which generate \cxxItem{SetAdd} and \cxxItem{SetRemove} neighbour structures on \cxxItem{network}. Next, the lifted neighbourhood templates are considered. These check which templates can be applied to the inner \cxx|set of Nodes| type. Those applicable are \cxxItem{SetAdd}, \cxxItem{SetRemove}, \cxxItem{SetMove}, and \cxxItem{SetCrossover}, which are all lifted to operate on members of \cxxItem{network} to create neighbourhood structures.

Finally, \athanor applies the lifted neighbourhood templates to \cxx|set of Nodes|, which requires the neighbourhood templates on \cxx|Nodes|, \cxxItem{intAssignRandom} and \cxxItem{intAssignRandomFromViolation}. These are both lifted twice, to create template structures that modify a single element in a single element of \cxxItem{network}.


\section{Generating Random Values} \label{sec:neighbourhoods-generating-random-values}
\label{sec:athanor-neighbourhoods-small-values}

\athanor requires  a method to generate a value that belongs to a given domain, both for generating initial variable assignments at the start of search and for generating new elements of a container type (for example, in the \texttt{setAdd} neighbourhood template).

Values are not generated uniformly at random, instead smaller values (i.e.\ values of smaller cardinality) are preferred for two reasons. Firstly, \essence domains can contain values so large that they would not typically fit in memory. Types can be arbitrarily nested, and several types describe variable-sized containers (such as \emt|set| and \emt|sequence|). In \Cref{sec:introduction} we gave an example where a small instance of SONET led to a domain containing values with cardinality \(2^{64}-66\).

Secondly, the unrolling and evaluation of constraints can be computationally expensive when a value has high cardinality. \athanor consistently aims to make small incremental changes that are computationally cheap to evaluate, and may be quickly undone if the new state is not accepted by the search strategy. Generating high cardinality random values is inconsistent with this general aim. When a large value is a necessary part of the solution, it may be reached by adding elements to a container over several iterations of search.

The method for generating random values in \athanor includes a mechanism for limiting the number of steps in the generation process, and therefore limiting the size of the generated values as well as the resources needed to generate a value. 
The resource limit is an essential part of the method. Without the resource limit, when generating a value of a variable-sized compound type, each possible cardinality would be generated with equal probability. 
The resource limit applies to every part of a nested domain and is described in detail below. 

\subsection{Generating Random Values Under Resource Limits} \label{sec:generating-random-values-under-resource-limits}

\begin{algorithm}[t]
\begin{algorithmic}[1]
\State \textbf{Input:} a multiset domain $d$, an integer quantity of resource $r_{\mathrm{in}}$
\If {$r_{\mathrm{in}} \leq 0$}
\State \Return $\mathrm{fail}, 0$ \Comment{no value created, no resource consumed}
\EndIf
\State Let $d_{\mathrm{inner}}$ be the inner domain \Comment{domain of the elements of the multiset}
\State $c \gets \{\}$ \Comment{create $c$, an empty container from the domain $d$}
\State $r \gets 1$ 
\State Let $n_{\mathrm{min}}$ and $n_{\mathrm{max}}$ be min and max cardinalities of $c$, as defined by $d$
\State $n \gets $ value chosen uniformly at random from $\{ n_{\mathrm{min}} \ldots n_{\mathrm{max}} \}$
\While{$\Size{c} < n$}
\LineComment{Calculate reserved resource that cannot be used by recursive call}
\State $r_{\mathrm{res}} \gets \max(0, n_{\mathrm{min}} - \Size{c} - 1) \times \textsc{calcMinResource}(d_{\mathrm{inner}})$
\State $e,r_e \gets $ \Call{generateRandom}{$d_{\mathrm{inner}}$, $r_{\mathrm{in}} - r_{\mathrm{res}} - r$}
\State \Comment{$r_e$ is the resource consumed by the recursive call}
\State $r \gets r + r_e$

\If {$e = \mathrm{fail}$}
\If {$\Size{c} < n_{\mathrm{min}}$}
\State \Return $\mathrm{fail}, r$
\Else
\State \Return $c, r$
\EndIf
\EndIf
\State Add $e$ to $c$
\EndWhile
\State \Return $c, r$
\end{algorithmic}
\caption{\textsc{generateRandomMSet} procedure with resource limit.}
\label{alg:generate-random-mset-with-reserved-resource}
\end{algorithm}

In this section, we describe the algorithm \textsc{generateRandom}. We use the multiset (\emt|mset|) domain as an example because it demonstrates the general method by which random values are produced for variable-sized container types.  Other types, such as \emt|set|, are very similar but require that other constraints (type invariants) are satisfied when generating a value. The procedure \textsc{generateRandomMSet}($d$, $r_{\mathrm{in}}$) is shown in \Cref{alg:generate-random-mset-with-reserved-resource}. \textsc{generateRandom} (for any type) has two parameters: $d$, which is the domain from which a random value is to be generated; and $r_{\mathrm{in}}$, an integer representing the amount of resource (a proxy for computation time) that may be consumed in the attempt to generate a random value.  If the \textsc{generateRandom} procedure determines that there is insufficient resource, the algorithm
has the option of failing.  In this case, no value is produced, only the amount of resource consumed in the attempt is returned.

\textsc{generateRandom} is a recursive procedure.  After initialising an empty container (for example, a multiset of $\tau$), the procedure may add elements to the container by invoking \textsc{generateRandom} with the inner domain $\tau$ (as in \Cref{alg:generate-random-mset-with-reserved-resource}).  However, it is necessary to share the amount of resource given to the \textsc{generateRandom} procedure between the initial invocation and any recursive invocations.  This is because the elements of a container may also be variable-sized and each may therefore consume different amounts of resource when generated. 
For example, consider applying this procedure to the domain \texttt{mset (maxSize 1000) of mset (maxSize 1000) of int(...)}. Not only will the outer multiset have a variable size, but so will each of its elements. In this case, controlling the size of the outer multiset alone is not an effective control of computational cost as  we may spend time producing few but very large inner multisets.  

Generating an atomic type (such as \texttt{int} or \texttt{bool}) has a cost of 1. Four variable-sized compound types (set, multiset, sequence, and relation) have a cost of 1 to initialise the container, while the other compound types (function, partition, matrix, tuple, record) do not. The cost of generating any compound type includes the cost of all attempts to generate an element (regardless of whether the element was successfully added to the container). 

In \textsc{generateRandomMSet} (\Cref{alg:generate-random-mset-with-reserved-resource}), \(r\) represents the resource used so far. It is initialised to 1, and each time an element is added to the multiset \(r\) is increased by the amount of resource used to generate the element. 
Note that the generation of each element of the multiset (by recursive call to \textsc{generateRandom}) must also have a resource limit. A simple resource limit for generating one element would be \(r_{\mathrm{in}}- r\), however it may be necessary to generate several elements to reach the lower cardinality bound \(n_{\mathrm{min}}\). Supposing we have \(|c|\) elements in the multiset, and also that the current iteration of the main loop of \Cref{alg:generate-random-mset-with-reserved-resource} will add one element, then \(n_{\mathrm{min}}-|c|-1\) more elements will be needed to meet the lower cardinality bound. 
The procedure reserves an amount of resource \(r_{\mathrm{res}}\) that is an estimate of the minimum amount required to generate \(n_{\mathrm{min}}-|c|-1\) elements. 
\textsc{calcMinResource} takes a domain and returns an estimate of the minimum resource required to make an element of the domain. 
The resource limit for the recursive call is then calculated as \(r_{\mathrm{in}}-r_{\mathrm{res}}-r\). 
If the recursive call to \textsc{generateRandom} fails, then \textsc{generateRandomMSet} either returns the current multiset, or (if an insufficient number of elements have been generated) it fails and returns \(r\). 

Other types are generated similarly. For sets, lines 12-20 of \Cref{alg:generate-random-mset-with-reserved-resource} are iterated until a value is generated that is not already in \(c\). Sequences are generated identically to multisets, except that adding an element to the container (line 20) appends the element to the end of the sequence (when doing so would not break an attribute such as \texttt{injective}). Relations are treated as sets of tuples. The algorithms for generating the fixed-size types (matrix, tuple, record) are the same as for a multiset of fixed size (i.e.\ with the \texttt{size} attribute), except that with tuples and records the inner domains may differ. Values of the atomic types such as \texttt{int} are chosen uniformly at random.

Functions divide into two cases. The first are total functions where the preimage is straightforwardly enumerable (i.e.\ it is of type \texttt{int}, \texttt{enum}, \texttt{bool}, or tuple of straightforwardly enumerable types). In this case, a fixed number of output values are required, and they are randomly generated in the same way that a fixed-length sequence would be. Otherwise, the process is very similar to generating a set of pairs. For each pair, first the preimage is generated (as in lines 12-14 of \Cref{alg:generate-random-mset-with-reserved-resource}) and this process is iterated until a unique preimage has been generated. Then, the image is generated, and (if successful) the mapping is added to the function. Resources are handled exactly as if generating a set of pairs. 

Partitions are generated in two passes. One pass generates a random sequence of distinct elements using the same method used for injective sequences. 
The other pass (which does not consume any resource) assigns each element in the sequence to a cell in the partition. If the partition is regular, the number of cells is chosen uniformly at random (satisfying any relevant attributes) and the elements are divided equally among the cells uniformly at random. 
Otherwise, the minimum number of cells are created (according to a \texttt{minNumParts} or \texttt{numParts} attribute, if present) and the minimum required number of elements are placed in each. Then, if there are elements remaining, one element \(e\) is chosen and an attempt is made to place \(e\) in an existing cell at random, or build a new cell containing \(e\) with the minimum required number of other elements. This step may fail by exceeding a cell's maximum cardinality, exceeding the partition's maximum number of cells, or having insufficient elements remaining to create a new cell. This step is iterated until there are no elements left from the sequence. 

\subsection{Resource Limits}

The \textsc{generateRandom} procedure is used to generate random values for the initial state and for some neighbourhood structures. In both cases, small values are preferable (as outlined above) and so \textsc{generateRandom} should be given a small value of \(r_{\mathrm{in}}\) initially. However, it may run out of resource and fail to generate a value, requiring it to be called again with a larger value of \(r_{\mathrm{in}}\). We define two constants: \(r_{\mathrm{mul}}=1.1\) and \(r_{\mathrm{min}}=500\). For the first call, we set \(r_{\mathrm{in}} \gets r_{\mathrm{mul}}\times \textsc{calcMinResource}(d)+r_{\mathrm{min}}\). If the first call fails, then we set \(r_{\mathrm{in}} \gets r_{\mathrm{mul}}\times r_{\mathrm{in}}\) and call \textsc{generateRandom} again, repeating until a value is returned. 

In summary, the effect of the resource limit is to introduce a strong bias towards values with low cardinality (for every part of a nested domain) while still allowing some freedom to generate values that are not of the minimal cardinality.

\section{Violation Counts}
\label{sec:violation_counts}

In constraint-based local search solvers, each constraint is often associated with a \emph{constraint violation} value (or violation degree)~\cite{hentenryck2009constraint,Michel2018}, which heuristically measures the number of necessary changes in the current assignment to satisfy the constraint, where the violation is $0$ when the constraint is satisfied. There can be more than one way of calculating a violation for a constraint, and the violation value does not have a single definition across all constraints. For example, the constraint violation $v$ for the constraint $x=y$ can be defined as $v = |x-y|$. One method of measuring violation for the \emph{allDiff} constraint counts the number of repeated values used by its variables. The violation of an arbitrary constraint is calculated inductively based on its structure. For example, the violation of a disjunction \(c_1\vee c_2\) is the minimum of the violations of \(c_1\) and \(c_2\). 
We refer to Michel and Van Hentenryck~\cite{Michel2018} for a full list of violation calculation rules for different types of constraints, on which \athanor's constraint violation rules are based. 

Another key concept in constraint-based local search solvers is \emph{variable violation}, which indicates how much each variable contributes to the violation of a constraint. 
Unlike constraint violations, which are only assigned to booleans, variables of all types can have a variable violation attached to them. 
In previous works~\cite{hentenryck2009constraint,Michel2018}, the violation of a variable $x$ within a constraint $c$ is defined as the largest possible decrease in the constraint violation of $c$ over all possible values in the domain of $x$. However, such calculations can be computationally expensive, especially in \athanor where arbitrarily nested variable types are allowed. 
Furthermore, in a high-level description of a problem, there are cases where only a single high-level decision variable exists in a constraint model. Attaching a violation to only such variable is too coarse. A useful additional feature would be to attach violations to parts of variables, such as the members of a set or sequence. Therefore, we propose a modified version of variable violation calculation, which \emph{heuristically} decides which variables, and parts of variables, are most likely the cause of current constraint violation.

The variable violation calculation in \athanor is done in a top-down manner. The violation of each constraint is assigned to the top operator of the constraint, and then the violation is recursively forwarded to the operands of each operator. Violations can optionally have an explanation label, which is also passed down. At present, the only two explanations are \lstinline{too_small} and \lstinline{too_large}, which can only be applied to expressions of type \cxx|int|. The default behaviour is that the violation count and label are passed unchanged to all operands of an operator, but some operators have a specialised implementation which better assigns the violations to operands.  We demonstrate the process with the following example:

\begin{lstlisting}
given n, y : int
find x: int(1..10)
such that x >= y
find s : set (size 5) of int(1..n)
such that 1 = sum i in s . toInt(i % 2 = 0)
\end{lstlisting}

Assume \lstinline|y = 5| and that the current assignment is \lstinline|x = 1|, \lstinline|s = {1,2,3,4,6}|. Consider the first constraint: \lstinline|x >= y|. \athanor will calculate the constraint violation \lstinline|v|, which can include an explanation. In the case of the \lstinline|>=| operator, \lstinline|v = max(y-x,0)  = 4| (see, e.g., \cite{Michel2018}, for more details). The violation and an explanation are then forwarded to the constraint's operands. For the \lstinline|>=| operator, the left hand side (\lstinline|x|) is assigned the violation of 4 and the explanation label \lstinline{too_small}, while the right hand side (\lstinline|y|) is also assigned the violation of 4 and the explanation label \lstinline{too_large}. This leads to \lstinline|x| being assigned a violation of 4 and \lstinline{too_small}. The violation on \lstinline|y| is dropped, as only variables can have violations, not parameters.

We now consider the second constraint: \lstinline|1 = sum i in s. toInt(i%2 = 0)|. As the right hand side is evaluated to 3, the left hand side receives a violation of \lstinline|v = abs(1-3)  = 2| and an explanation \lstinline{too_small}, although this violation is ignored as it applies to a constant. The same violation, 2, and the explanation \lstinline{too_large} is assigned to the right hand side, and this violation and explanation are subsequently passed to each operand of the \lstinline|sum| operator, \lstinline|toInt(i%2 = 0)|. The violation count is assigned to every member of the quantification, rather than split, under the expectation that usually a single step of local search will tend to change one, or a few, members of the quantification, rather than all members equally.

Rather than pass on the violation count to its operand unchanged, the \lstinline|toInt| operator has a more optimised implementation. The \lstinline|toInt| operator can only take one of two possible values (0 or 1), so we can infer information about the largest possible decrease in violation that can occur. More concretely, if the operand is evaluated to \lstinline|false|, then the \lstinline|toInt| operator is already at its smallest possible value (0), and so cannot take a smaller value. Therefore, any violation labelled as \lstinline{too_large} is ignored. In our example, the violation is dropped when \lstinline[mathescape]|i $\in$ {1,3}|. \lstinline|toInt| is the main operator where violation counts are useful, where they avoid marking expressions as violating when changing their value would not actually reduce the overall violation count.

On the other hand, when \lstinline[mathescape]|i $\in$ {2,4,6}|, the \lstinline|toInt| operand is \lstinline|true|, which means it can become smaller. Therefore, in this case a violation of 1 (as the value of the \lstinline|toInt| operator can change by at most 1) and an explanation \lstinline{too_large} are passed down to the \lstinline|=| operator and subsequently to the operator's non-constant operand: \lstinline|i % 2|. The operator \lstinline|%|, on the other hand, simply passes down the violation to its non-constant operand, variable \lstinline|i|, which is an element of \lstinline|s|. For implementation simplicity, not every operator in \athanor generates an explanation. In this case, the operator \lstinline|%| does not assign any explanation to its operands.

With the presence of nested types, in \athanor the variable violation on any containing structure \lstinline|s| is the sum of the variable violation directly associated with \lstinline|s| and the variable violation associated with all elements of \lstinline|s|. In our previous example, the violation of \lstinline|s| is defined as the sum of variable violations of elements 2, 4, and 6 (there was no variable violation associated with elements 1 and 3 in this example).

\essence, and therefore \athanor, operate on relational semantics \cite{frisch-stuckey-undef-09}. This means Booleans can be true, false, or undefined. Since undefined values can have such a large impact, with one operand potentially causing the entire expression tree to become undefined, \athanor assigns undefined a very large violation count ($2^{32}$). An investigation into other methods of assigning violation to undefined values is left for future work.

Information about constraint violation and variable violation is used frequently during the search in \athanor. As described in~\Cref{sec:search-procedures}, the total constraint violation is used for measuring the magnitude of infeasibility of an assignment (the smaller the better). Variable violation, on the other hand, is used for biasing towards variables with higher contributions to the violation of the current assignment during the application of a neighbourhood structure, i.e., elements with higher variable violations will have a higher probability of being selected.

\section{Incremental Evaluation} \label{sec:incremental-evaluation}

During search, whenever a variable is reassigned we must calculate new values and new violation counts for the constraints. Rather than reevaluating the entire AST, which can be computationally expensive, \athanor only reevaluates the nodes on the paths from the changed variable (leaf node) to the root. The reevaluation is done incrementally and is optimised for efficiency via three mechanisms: caching, triggering, and incremental hashing. \athanor also uses invariants in some cases, as described below. 

\subsection{Caching} \label{subsec:cache}

Every node in the AST can be queried for a value. In the case of a leaf node, this is the value assigned to the variable represented by the leaf. For a non-leaf node, this is the value produced by evaluating the operator encoded by the node. In most cases, operators cache their values after evaluation and simply return the cached values when queried. 
In addition, the operators also cache the values of their operands when needed. For example, a \lstinline|sum| operator will cache the current sum of all its child nodes and the value of each element in the sum. Some operators do not cache their values, but instead forward the request for a value to a descendant node. For example, the sequence index operator, which accepts a sequence $s$ and an integer index $i$, and evaluates to the member $m$ of $s$ whose position matches $i$. Unless $i$ is out of bounds (see below for a discussion of undefinedness) instead of storing its own value, when queried the sequence index operator simply forwards the query to $m$. 

The cached values are used for incremental evaluation whenever possible. This is particularly useful when there are nodes in the AST that have a very large number of children. Consider, for example, summing a list of length 1000. 
As an example, when changing the value of \lstinline|c| from \lstinline[mathescape]|$c_1$| 
to \lstinline[mathescape]|$c_2$| in the expression \lstinline[mathescape]|s = sum({a,b,c,d,e})|, the new value of the sum can be calculated as \lstinline[mathescape]|s' = s - $c_1$ + $c_2$|. This allows the \lstinline|sum| operator to be reevaluated without querying the values of \lstinline|a|, \lstinline|b|, \lstinline|d|, and \lstinline|e|. 

\subsection{Triggering} \label{subsec:triggering}

During incremental evaluation, operators must check if they have to update their current values. Each operator makes use of two pieces of information to update its value during the incremental evaluation: its current cached values and the changes made by its child nodes. Changes in child nodes are communicated to their parents through \emph{triggers} - functions that describe in detail the changes in the child nodes' values. All data types share a common set of three \emph{basic} trigger functions:

\begin{itemize}
\item \lstinline|valueChanged()| notifies that the value of the node has changed. 
\item \lstinline|hasBecomeUndefined()| notifies that the value of the node has become undefined (for example, dividing by 0, or indexing a sequence by an out-of-bounds value, such as $-1$). 
\item \lstinline|hasBecomeDefined()| notifies that the value of a node has changed from undefined to defined.
\end{itemize}

While these three triggers are sufficient to implement incremental evaluation, many abstract types extend the basic set of trigger functions in order to give a more detailed description of the change in the value of a node. This allows a more efficient implementation of incremental evaluation in many cases.  Consider, as an example, the \lstinline|set| type. Notifying a set operator that the value of one of its operands (of type \lstinline|set|) has changed gives very limited information to the operator. Only one member of the set may have changed, or the entire set might have been replaced with a different one.  The basic \mbox{\lstinline|valueChanged()|} trigger function will always need a full reevaluation of the operator.  To improve performance, \athanor defines additional trigger functions for the \lstinline|set| type, including:

\begin{itemize}
    \item \lstinline|valueAdded(index)|: a new element is added to the set at \lstinline|index|.
    \item \lstinline|valueRemoved(index)|: an element is removed from the set at \lstinline|index|.
    \item \lstinline|memberValueChanged(index)|: an element in the set is changed at \lstinline|index|.
\end{itemize}

In addition, some compound types share two additional trigger functions: \lstinline|memberHasBecomeDefined(index)| and \lstinline|memberHasBecomeUndefined(index)|, which indicate that some element has become defined or undefined, respectively. These are useful for operators such as $f(y)$, which evaluates to the image of $y$ w.r.t\ the function $f$, and which only becomes undefined if the image of $y$ becomes undefined.

Note that while types such as sets are unordered in \essence, we treat abstract types as ordered internally, so any particular value remains at the same index where possible.

Each operator makes use of the information provided by the triggers to update its current cached value and violation count (if it is Boolean). Consider the constraint \lstinline|a subsetEq b|. This expression is \lstinline|true| or \lstinline|false|. \athanor also stores the violation count, which is the number of members of \lstinline|a|  not contained in \lstinline|b|. The expression is \lstinline|true| when the violation count is $0$. Assume the violation of the expression, denoted as \lstinline|v|, has been evaluated previously. The following demonstrates how changes to the value of \lstinline|a| change the violation count \lstinline|v|. 

\begin{description}
\item[\cxxItem{a->valueAdded(index)}] If \lstinline[mathescape]|a[index] $\notin$ b| 
then \lstinline[mathescape]|v $\gets$ v + 1|

\item[\cxxItem{a->valueRemoved(index)}] If \lstinline[mathescape]|a[index] $\notin$ b| then \lstinline[mathescape]|v $\gets$ v - 1|

\item[\cxxItem{a->memberValueChanged(index)}] Let $x$ be the old value at index \lstinline|index|, and $y$ be the new value. Then run \cxxItem{valueRemoved(x)} followed by \cxxItem{valueAdded(y)}.

\item[\cxxItem{a->valueChanged()}] Set has changed in an unknown way, perform a full re-evaluation of \lstinline|v|.

\item[\cxxItem{a->hasBecomeUndefined()}] assign the violation count \(2^{32}\) to \lstinline|v| (as discussed in \Cref{sec:violation_counts}).
\item[\cxxItem{a->hasBecomeDefined()}] If \lstinline|b| is defined, perform a full re-evaluation of \lstinline|v|.
\end{description}

A similar procedure is followed for trigger notifications from \lstinline{b}. Another example of triggers and incremental evaluation for the \lstinline|sum| operator and the \lstinline|sequence| type can be found in~\ref{sec:trigger_sum_operator}.

\subsection{Value Representation and Incremental Hashing}\label{sec:value-representation}\label{sec:incremental-hashing}

It is important for scalability that values are represented compactly, particularly for types where the size is
variable. \athanor{} uses representations that scale approximately linearly with the actual size of the value, not its upper-bound size.
For example, values of type \texttt{set of int} are represented with an extensible array of references to the elements, combined with a hash set (i.e.\ a hash table containing only keys) of the elements, both of which scale approximately linearly with the number of elements. Each type has a hash function that is incrementally updated to reflect changes made to a value of that type. The incremental hash functions are important for efficiency because hash values are extensively used to compare values for equality or disequality. \ref{sec:appendix-hashing} describes the value representations and incremental hash functions for each type. In very rare cases a hash collision can affect the behaviour of \athanor, and we also discuss this issue in \ref{sec:appendix-hashing}.

\subsection{Invariants and Partial One-Way Propagation}

Invariants, also known as one-way constraints, are a common feature of constraint-based local search solvers (as discussed in \Cref{sec:related-work}). An invariant defines one set of variables in terms of another set. For example, given $x=y+2$, we can allow the search process to change $y$, and calculate the value of $x$ from $y$ when required.

\athanor implements invariants for integer variables. When an integer variable \(a\) is contained in an equality constraint \(a=e\) (with any expression \(e\)), all other occurrences of \(a\) are replaced with \(e\) throughout the specification and the variable \(a\) is deleted. To ensure that \(e\) takes a value within the domain of \(a\), the constraint \(a=e\) is replaced with \(e\in D(a)\), where \(D(a)\) is the domain of \(a\), unless this constraint is trivially true in which case it is replaced with \textsc{True}. 

Integer elements of compound types are a more difficult case.  Consider an equality involving a member of a compound type, for example $M[2] = y+2$, for a sequence $M$ and integer $y$. In this case we cannot remove $M[2]$ because the sequence $M$ as a whole may appear in other constraints and will have neighbourhood structures. However, we would still like to use this constraint to update $M[2]$ when $y$ changes.

To handle such cases, we introduce a weaker technique which we call \emph{partial one-way propagation}. Given a constraint $b=e$ where $b$ is an integer element of a compound type, partial one-way propagation can assign $b$ the value of $e$ whenever $e$ changes, without deleting $b$ or deleting the constraint $b=e$. 
We implement partial one-way propagation for constraints \(b=e\) when: \(b\) is an integer element of a compound type; the value of \(e\) must always be in the domain of \(b\); and the domain of the decision variable \(x\) containing \(b\) places no restrictions on the value of \(b\) other than the domain of \(b\) itself. 

The final requirement is to ensure that updating \(b\) will not violate the type of \(x\). For example, if \(x\) is an \texttt{injective} sequence, updating \(b\) could break injectivity. Similarly, sets and partitions disallow duplicate values so their elements also cannot be updated by partial one-way propagation. Restrictions on an integer element can arise from any layer of the type of \(x\); for example, while the integers in a \texttt{sequence (maxSize 5) of int} can be updated by partial one-way propagation, the integers in a \texttt{set of sequence (maxSize 5) of int} cannot, because the set cannot contain duplicate sequences.

\athanor keeps track of all constraints of the form \(b=e\) that satisfy the conditions given above. For each such constraint, whenever the value of \(e\) is changed during incremental evaluation the value of \(b\) is updated automatically. To prevent cycles, each element of a compound type may be updated at most once during each pass of incremental evaluation. The constraint $b=e$ is not removed, and may be violated when the cycle detection prevents updating $b$.

An example can be seen in the Minimum Energy Broadcast problem (\Cref{fig:meb-spec}): for a total function named \texttt{depths}, we have the following constraint:
\begin{lstlisting}
depths(child)=depths(parent)+1
\end{lstlisting}
The element \lstinline{depths(child)} is updated when the value of the right-hand side changes. 

The CBLS systems described in \Cref{sec:related-work} have various sophisticated ways of identifying invariants and processing them efficiently. There is clear potential for improvement to \athanor in this area. 


\section{Search Procedures} \label{sec:search-procedures}

In \athanor, as is usual for local search-based metaheuristics, the generated neighbourhood structures are treated as black boxes -- procedures which may improve or worsen the violation count or the objective.  A single iteration of search consists of the selection of one variable and its reassignment via a neighbourhood structure. Built around these iterations is the traditional metaheuristic design; the metaheuristic decides after each iteration of search whether the new solution should be accepted or the change should be reversed.
The metaheuristic used in \athanor follows the well-known Iterated Local Search (ILS) algorithm~\cite{lourencco2019iterated} where a hill climbing phase (for improving the current solution) and an exploration phase (for escaping local optima) are interleaved. During the hill climbing phase, neighbourhood structures are selected \emph{dynamically} to ensure the most effective ones are chosen depending on the current stage of the search.

In this section, we first explain the overall search procedure of \athanor (\Cref{sec:search_archtecture}). The exploration phase is then described in~\Cref{sec:exploration}, followed by details of the hill climbing phase (\Cref{subsec:climb_to_zero} and \Cref{subsec:climb_standard}). Finally, the dynamic neighbourhood structure selection mechanism is explained in~\Cref{subsec:dynamic_neighbourhood_selection}.

\subsection{Search Architecture}
\label{sec:search_archtecture}

The overall search procedure of \athanor is presented in \Cref{alg:athanor-search}.\footnote{All constant values used in \athanor's search procedure were chosen based on manual tuning on some example instances taken from CSPLib~\cite{CSPLib}. A systematic investigation of the influence of those parameters on solver performance and how to select the best values is left for future work.}
The search starts by randomly assigning values to all variables (line $1$), initialising the two best solutions so far (line $2$) and the random walk's length for the exploration phase (line $3$). The two best solutions saved during the search include: (i) $s_{best}$, the current global best during the entire search; and (ii) $s^*$, the local best solution obtained before the random walk's length reaches its limit.

\begin{algorithm}[t]
  \caption{Overall search procedure of \athanor}
  \label{alg:athanor-search}
    \begin{algorithmic}[1]
    \State $s \gets$ random initial assignment \Comment{current solution}
    \State $s_{best} \gets s$, $s^* \gets s $ \Comment{global and local best solutions so far}
    \State $n_r \gets 10$        \Comment{random walk length}
    \While{time limit not reached}
    \If {$\Call{violation}{s} > 0$}
        \State $s \leftarrow \Call{hillClimberToZeroViolation}{s}$ \Comment{repair violations}
    \EndIf
    \State $s \gets \Call{hillClimber}{s}$ \Comment{improve objective function}
    \If {$\Call{strictlyBetter}{s, s^*}$}
        \State $s^* \gets s$ , $n_r \gets 10$ \Comment{update local best \& reset random walk length}
    \If {$\Call{strictlyBetter}{s, s_{best}}$}
        \State $s_{best} \gets s$  \Comment{update global best}
    \EndIf
    \Else \Comment{exploration phase to avoid local optima}
    \State $n_r \gets  n_r \times 1.3$ \Comment{increase random walk length}
    \State $s \gets$ \Call{randomWalk}{$s$, $n_r$} \Comment{random walk for $n_r$ steps}
    \If {$n_r > 500$} \Comment{exploration phase reaches its limit}
    \State $n_r \gets 10$, $s^* \gets s $ \Comment{reset random walk length and local best}
    \EndIf
    \EndIf
    \EndWhile
    \Return $s_{best}$
\end{algorithmic}
\label{alg:overall_search_procedure}
\end{algorithm}

For constrained problems, it is possible that the current solution does not satisfy all constraints. Therefore, in the main loop, \athanor first tries to repair any violations via a hill climbing search (line $6$, procedure \textsc{hillClimberToZeroViolation}) before starting to alternate between two phases: (i) a hill climbing phase (line $7$, procedure \textsc{hillClimber}), which focuses on improving the objective function value; and (ii) an exploration phase (line $14$, procedure \textsc{randomWalk}), which makes a number of random changes to the current solution $s$ obtained from the hill-climbing if $s$ is not better than the local best $s^*$.
The comparator \textsc{strictlyBetter}($s$, $s^*$) (line $8$) is defined as follows:

\begin{algorithmic}
  \Procedure{strictlyBetter}{$s$, $s^*$}
\State \Return $\Call{\textsc{violation}}{s} < \Call{\textsc{violation}}{s^*}\ \lor$ \\
$\hspace{1.5cm}(\ \Call{\textsc{violation}}{s} = 0 \land \Call{\textsc{violation}}{s^*} = 0\ \land$ \\
$\hspace{1.6cm}\Call{\textsc{objective}}{s} < \Call{\textsc{objective}}{s^*}\ )$
  \EndProcedure
\end{algorithmic}

where $\textsc{violation}(s)$ and $\textsc{objective}(s)$ correspond to the total number of constraint violations (as described in~\Cref{sec:violation_counts}) and the objective function value of a solution $s$.

The idea of alternating between the two phases (exploration and hill climbing) is that if the hill climbing fails to find a better solution, this is an indication that \athanor has become stuck in a local optimum.
The exploration phase aims to force \athanor into a different area of the search space by making random changes to the current solution. The number of changes is limited by $n_r$.
If hill climbing still fails to improve on the local best solution after the perturbation, random walk is applied again with an increased value of $n_r$ (lines $13-14$).
Once $n_r$ reaches a certain limit,
\athanor redefines the local best as the current solution after exploration and resets $n_r$ (line $16$). This can be considered as a random restart when the current exploration is no longer effective.

In this work, when a neighbourhood structure is applied on a current assignment, we simply sample a random assignment from the neighbourhood and replace the current assignment with the new one. There are several alternatives used in the literature, such as exploring a random subset of the neighbourhood and choosing the best assignment from the subset, or exploring the neighbourhood until the first improved assignment is found. An investigation of those alternatives is left for future work.

\subsection{Exploration}
\label{sec:exploration}

\begin{algorithm}
  \caption{Exploration phase in \athanor (\sc{randomWalk})}
  \label{alg:athanor-random-walk}
\begin{algorithmic}[1]
\State \textbf{Input:} current solution $s$, random walk length $n_r$
\State $v_{max} \gets \Call{violation}{s} + n_r$ \Comment{cap on violation of current assignment}
\State $i \gets 0$
\While {$i \leq n_r$}
\State $s^{\prime} \gets$ apply a random neighbourhood structure on $s$
\If {$\Call{violation}{s^{\prime}} \leq v_{max}$}
    \State $s \gets s^{\prime}$, $i \gets i+1$
\EndIf
\EndWhile
\Return $s$
\end{algorithmic}
\end{algorithm}

The exploration phase is detailed in~\Cref{alg:athanor-random-walk}.
This phase aims to escape local optima by making a number of unguided random moves. Each move is performed by randomly selecting a neighbourhood structure and applying it to the current assignment. However, we found that simply allowing \athanor to make unrestricted changes to the assignment was unproductive because some neighbourhood structures make much more dramatic changes than others.
For example, given a nested structure such as a \emt|set of set of int|,
one neighbourhood structure deletes a single integer from an inner set, while another neighbourhood structure deletes a set of integers.
An important area of future work is to investigate different strategies to control the magnitude of the changes made by each neighbourhood structure. In this paper, we adopt a simple strategy where we set a limit on the increase in the violation allowed during the current exploration phase (set by $n_r$). A tight restriction on violation increase tends to cause \athanor to make small changes to the assignment, if the specification includes some constraints. The same parameter \(n_r\) also determines the number of random moves made during the exploration phase. Therefore, increasing \(n_r\) allows individual moves to make larger changes to the assignment, and also increases the number of random moves. While \Cref{alg:athanor-random-walk} could hypothetically enter an infinite loop if applying random neighbour structures failed to find any assignments that satisfy the violation requirements, we have never observed this. If it did, the algorithm could be modified to terminate after a specified number of failed attempts to apply a random neighbourhood structure.

\subsection{Hill Climbing for Repairing Violations}
\label{subsec:climb_to_zero}

\begin{algorithm}
  \caption{Procedure \textsc{hillClimberToZeroViolation}}
  \label{alg:athanor_climb_to_zero}
    \begin{algorithmic}[1]
    \State \textbf{Input:} current solution $s$
    \State $i \leftarrow 0$
    \While {$\Call{violation}{s} > 0$}
        \State $s^\prime \leftarrow \Call{applyNeighbourhoodStructure}{s}$
        \If {$\Call{violation}{s^{\prime}} < \Call{violation}{s}$}
            \State $s \leftarrow s^{\prime}$, $ i \leftarrow 0$
        \Else
            \State $i \leftarrow i + 1$
        \EndIf
        \If {$i = 5000$}
            \State $s \leftarrow $ random assignment, $i \leftarrow 0$
        \EndIf
    \EndWhile
    \State \Return $s$
    \end{algorithmic}
\end{algorithm}

The \textsc{hillClimberToZeroViolation} procedure (called on line 6, \Cref{alg:overall_search_procedure}) is described in~\Cref{alg:athanor_climb_to_zero}. This procedure solely focuses on repairing any violations in the current assignment. At each iteration, a neighbourhood structure is dynamically selected (details on how neighbourhood structures are selected are described in \Cref{subsec:dynamic_neighbourhood_selection}) and applied to the current solution $s$ (line $4$). The new solution is assigned to $s$ if it has a smaller number of violations (line $6$). To avoid getting stuck in local optima, we restart the whole search with a random assignment if there is no improvement in terms of violations across $5000$ consecutive iterations (lines $9-10$).

\subsection{Hill Climbing for Improving Objective Function Value}
\label{subsec:climb_standard}

The \textsc{hillClimber} procedure (called on line 7, \Cref{alg:overall_search_procedure}) assumes that the input solution is feasible and focuses on improving the objective value. This procedure is comprised of two different versions of hill climbing. The first one, namely \textsc{hillClimberStandard}, is a simple and standard hill climbing algorithm where only strictly better solutions are accepted, while the second one, namely \textsc{hillClimberWithViolations}, is more sophisticated and temporarily allows assignments with constraint violations. The latter one was introduced to address the issue where heavily constrained variables exist and the search must temporarily violate constraints to improve the objective.

Each version of hill climbing is given a fixed budget of $5000$ solution evaluations during each iteration of the overall search procedure of \athanor. We first describe the sophisticated hill climbing version (\Cref{subsec:hc_with_temporary_violations}), followed by the dynamic selection mechanism to choose between \textsc{hillClimberStandard} and \textsc{hillClimberWithViolations} (\Cref{subsec:dynamic_selection_hc}).

\subsubsection{Hill Climbing with Temporary Violations}
\label{subsec:hc_with_temporary_violations}

\begin{algorithm}[!h]
  \caption{Procedure \textsc{hillClimberWithViolations}}
  \label{alg:athanor_climb_with_violations}
    \begin{algorithmic}[1]
    \State \textbf{Input:} current solution $s$, budget $b$
    \State $i \leftarrow 0$ \Comment{number of evaluations used by the whole procedure}
    \State $n_{vio} \leftarrow 20$ \Comment{current violation limit}
    \State $s^{\prime} \leftarrow s$ \Comment{current best solution}
    \State $o \leftarrow \Call{objective}{s}$ \Comment{current best objective value (without violation)}
    \Repeat
        \LineComment{We first attempt to improve the objective while allowing violations}
        \State $k_1 \leftarrow 0$ \Comment{number of evaluations used by the first loop}
        \While {$\Call{objective}{s^{\prime}} \geq o$ \textbf{and} $k_1 \leq \mathrm{min}(b-i, 500)$ }
            \State $s^{\prime\prime} \leftarrow \Call{applyNeighbourhoodStructure}{s^{\prime}}$
            \State $k_1 \leftarrow k_1 + 1$
            \If {$\Call{violation}{s^{\prime\prime}} \leq n_{vio}$ \textbf{and} $\Call{objective}{s^{\prime\prime}} \leq o$}
                \State $s^{\prime} \leftarrow s^{\prime\prime}$
            \EndIf
        \EndWhile
        \State $i \leftarrow i + k_1$
        \LineComment{Now attempt to repair violations}
        \State $k_2 \leftarrow 0$ \Comment{number of evaluations used by the second loop}
        \While {$\Call{violation}{s^{\prime}} > 0$ \textbf{and} $k_2 \leq \mathrm{min}(b-i, 500)$ }
            \State $s^{\prime\prime} \leftarrow \Call{applyNeighbourhoodStructure}{s^{\prime}}$
            \State $k_2 \leftarrow k_2 + 1$
            \If {$\Call{violation}{s^{\prime\prime}} \leq \Call{violation}{s^{\prime}}$ \textbf{and} $\Call{objective}{s^{\prime\prime}} < o$}
                \State $s^{\prime} \leftarrow s^{\prime\prime}$
            \EndIf
        \EndWhile
        \State $i \leftarrow i + k_2$
        \If {($\Call{violation}{s^{\prime}}=0$ \textbf{and} $\Call{objective}{s^{\prime}} < o$) \textbf{or} $n_{vio} \geq 20 \times 1.2^{10}$}
            \State $n_{vio} \leftarrow 20$
            \State $o \leftarrow \Call{objective}{s^{\prime}} $
        \Else
            \State $n_{vio} \leftarrow n_{vio} \times 1.2$
        \EndIf
    \Until {$i \geq b$}
    \Return $s^{\prime}$
    \end{algorithmic}
\end{algorithm}

The \textsc{hillClimberWithViolations} procedure is shown in \Cref{alg:athanor_climb_with_violations}. Starting from a current solution $s$ and an initial limit $n_{vio}$ for temporary violations (lines $3-4$), the main loop includes two stages. First, \athanor searches for an assignment with a better objective value while allowing violations up to the limit of $n_{vio}$ (lines $8-13$). The second stage attempts to repair the introduced violations while maintaining the improved objective value (lines $16-21$). Following the second stage, if the search found a feasible solution with improved objective value then we reset $n_{vio}$ and the target objective, essentially re-starting the procedure from the improved assignment \(s'\) (lines $23-25$).
Also, when the violation limit reaches a certain value (line $23$), the search is reset but with an assignment \(s'\) that could be worse than the original assignment \(s\). Otherwise, we increase the violation limit $n_{vio}$ and continue to the next iteration (line $27$).

\subsubsection{Dynamic Selection of Hill Climbing Procedure}
\label{subsec:dynamic_selection_hc}

Each time \textsc{hillClimber} is called from \Cref{alg:overall_search_procedure}, a choice is made between \textsc{hillClimberStandard} and \textsc{hillClimberWithViolations}. It is not known a priori which one is most suitable at a given stage of the search, so the choice is made dynamically.
The choice is treated as a Multi-Armed Bandit (MAB) problem~\cite{berry1985bandit}. The MAB problem considers a bandit with multiple independent arms, each of which when pulled returns a random reward. The reward distribution of each arm is unknown, and the aim is to choose which arms to pull such that the total reward is maximised. The most important point when solving a MAB problem is to have a balance between exploitation (pulling the best-so-far arm) and exploration (trying a new or less frequently selected arm to gather more information). One of the most common strategies for achieving a balanced exploitation-exploration trade-off is the Upper Confidence Bound (UCB) algorithm~\cite{agrawal1995sample}, where the selected arm $a^*$ for a time step $t$ is defined as:

\begin{equation}
    a^* = \operatorname*{argmax}_{a \in A}
    \left(\frac{R_t(a)}{n_t(a)}
    + c \times \sqrt{\frac{\ln t} {n_t(a)}}
    \right)
\label{ucb_equation}
\end{equation}

where $A$ is the set of arms, $n_t(a)$ is the number of times arm $a$ has been pulled until time $t$, and $R_t(a)$ is the current total reward received by pulling arm $a$. The formula offers a balance between exploitation (the first term) and exploration (the second term), and $c$ is a parameter of the algorithm~\footnote{$c$ is set as $1$ in all experiments in this paper}.

In our context, each version of hill climbing is an arm. The individual reward received at each time step (i.e., when an arm is pulled) is defined as 
the number of times that the objective value is improved during the hill climbing call without introducing constraint violations.

\subsection{Dynamic Neighbourhood Structure Selection}
\label{subsec:dynamic_neighbourhood_selection}

In this section, we describe in detail how neighbourhood structures are selected during each hill climbing variant (i.e., the \textsc{applyNeighbourhoodStructure} procedure in \Cref{alg:athanor_climb_to_zero} and \Cref{alg:athanor_climb_with_violations}).

A typical \essence{} specification with high-level nested types usually leads to the instantiation of several neighbourhood structures.
Some neighbourhood structures may be better at improving the objective, while others may be better at reducing violations when looking for a feasible solution.  Of course there may be some that are not favourable for either. The aim of dynamic neighbourhood structure selection is to quickly classify these neighbourhood structures and bias the search towards the neighbourhoods appropriate to the current search stage.

Consider, for example, the SONET problem and its objective,  the minimisation of the sum of the inner sets' cardinalities.  One of the matching neighbourhood structures, \verb=setRemove=, as the name suggests, closely fits the objective by explicitly focusing on reducing the sizes of the inner sets.  The inverse, \verb=setAdd=, is not useful with regards to the objective as it always leads to a worse solution, but can aid with satisfying the constraints of the problem.

Similar to the dynamic selection of hill climbing procedures, we make use of the UCB algorithm to select neighbourhood structures during the search, i.e., each neighbourhood structure is an arm. There are three UCB controllers for neighbourhood structure selection, each of which targets a different stage of the search and therefore has a different definition for the reward function. 
The first controller assigns an individual reward of one to a neighbourhood structure when the total constraint violation is reduced. This controller is used when the main focus is on repairing the violations (line $4$ of \Cref{alg:athanor_climb_to_zero} and line $18$ of \Cref{alg:athanor_climb_with_violations}).
The second controller focuses on improving the objective value. It gives each neighbourhood structure an individual reward of one when the objective value is improved, irrespective of changes in constraint violations. This controller is used in line $10$ of \Cref{alg:athanor_climb_with_violations}.
The third controller gives each neighbourhood structure an individual reward of one only when the objective value is improved without introducing constraint violations. This last one is used in the \textsc{hillClimberStandard} procedure described at the beginning of ~\Cref{subsec:climb_standard}.

The cost to execute and evaluate a neighbourhood structures can vary significantly, some neighbourhood structures can be much more computationally expensive than others. The original UCB formula (\Cref{ucb_equation}) simply counts the number of times each neighbourhood structure is selected by a UCB controller, neglecting the computational cost of applying the neighbourhood structure. Here, we use an approximation of the total amount of computational resources consumed by the application of the neighbourhood, \(\mathrm{cost}(a)\). This includes the total resources of all the inner loops executed during the neighbourhood structure application.
If an iteration of an inner loop involves generating new random values for a variable within the current assignment, the resources are calculated as detailed in~\Cref{sec:generating-random-values-under-resource-limits}. For all other cases, the resource for each iteration is simply counted as 1 because the implementation involves simpler operations (e.g., reassigning a pointer).
Each time a neighbourhood structure is selected by a UCB controller, $n_t(a)$ is increased by \(\mathrm{cost}(a)+1\).

\section{Experimental Evaluation} \label{sec:athanor-experiments} \label{sec:experiments}

In this section we evaluate the performance of \athanor in comparison with a collection of other solvers. We compare with constraint-based local search solvers that are able to generate neighbourhoods automatically (since \athanor generates neighbourhood structures automatically), and with systematic constraint solvers that have performed well in recent competitions. We have two hypotheses: first, that the structure available in a high-level problem specification can be exploited to generate effective local search neighbourhoods; and second, that \athanor's use of variable-sized data structures (for both values and expressions) will allow it to scale gracefully to large problem instances.

In the first experiment we compare \athanor{} with local search and systematic solvers on a collection of benchmark problems, ranging from simple to complex in structure.
The results show that \athanor performs well overall but comparatively better for problem classes with nested structure (such as a set of partitions). Our results broadly confirm the first hypothesis.

In the second experiment, we use new sets of large instances of four problem classes (for example, new knapsack instances with 80,000 objects) to explore how \athanor scales compared with the other solvers. The results demonstrate \athanor's ability to scale gracefully to large instances (through its support for variable-sized data structures) on three of the four problem classes, largely confirming the second hypothesis.

All models and instances used for the experiments and the instance generators we created, together with experimental results are all available in the repository \url{https://github.com/athanor/athanor-experiments/}. The source code of \athanor is available at \url{https://github.com/athanor/athanor}. The constraint modelling tool \conjure~\cite{conjure-aij}\footnote{\url{https://github.com/conjure-cp/conjure}} is used by \athanor to parse \essence. 

\subsection{Solvers}\label{sec:othersolvers}

We compare \athanor with six other solvers in seven configurations, together with a simplified version of itself, as follows:

\begin{description}

\item[LNS-PG:] propagation-guided large neighbourhood search \cite{DBLP:conf/cp/PerronSF04} as implemented in Choco \cite{Prud'homme2022} version 4.10.10. Details are provided below.
\item[LNS-EB:] explanation-based large neighbourhood search \cite{DBLP:journals/constraints/PrudhommeLJ14} as implemented in Choco version 4.0.9.  At the time of performing the experiments, 4.0.9 was the most recent version of Choco to support explanation-based LNS.
\item[Yuck:] a constraint-based local search solver.\footnote{\url{https://github.com/informarte/yuck} (the release dated November 1st, 2022)}
\item[\fznoscar:] a constraint-based local search solver~\cite{bjordal15,bjordal2021declarative} based on OscaR-CBLS~\cite{de2013oscar}.\footnote{Downloaded from \url{http://user.it.uu.se/~gusbj192/fzn-oscar-cbls/latest/oscar-cbls-flatzinc.zip}, version dated August 22nd, 2021, which is the latest release as of 8th June 2023.}
\item[Chuffed:] a systematic CP solver with conflict learning~\cite{chuffedGithub}, version 0.10.4 with the free search option enabled.
\item [OR-Tools:] a systematic CP solver with conflict learning,\footnote{\url{https://github.com/google/or-tools}} version 9.4.1874 with the free search option enabled. 
\item[SNS:] structured neighbourhood search \cite{SNS-IJCAI18}, a local search solver for \essence{}.
\item[Athanor-Reduced:] a simplified version of \athanor itself where the more complex neighbourhood templates are removed. This is to validate the  impact of high-level neighbourhoods on the solver's performance.
\end{description}

The other solvers and configurations were chosen to include: both prominent approaches to automatically generating neighbourhoods for LNS (propagation-guided and explanation-based); 
\emph{open-source} CBLS solvers that were entered in the Local Search track of the MiniZinc Challenge Series (Yuck and \fznoscar); 
two systematic CP solvers, including OR-Tools (the overall winner of several recent MiniZinc Challenges) and Chuffed (a solver with competitive performance to OR-Tools);
and Structured Neighbourhood Search, the only solver other than \athanor that uses the structure available in \essence{}.

LNS-PG and LNS-EB were both implemented using the API of Choco. The initial search for a feasible solution uses a binary depth-first search with the dom/wdeg variable ordering heuristic and random value ordering. The search restarts when a backtrack limit is reached, initially restarting after 50 backtracks; the backtrack limit is increased by a multiple of 1.5 at each restart. The search used within LNS is also a binary depth-first search with dom/wdeg variable ordering. The value ordering is minimum value first, and search is limited to 50 backtracks.

Yuck, \fznoscar, Chuffed, and OR-Tools are called via MiniZinc~\cite{nethercote07}\footnote{\url{https://www.minizinc.org/}} 2.6.4.
The version of \athanor used in the experiments is release 0.9.9.\footnote{\url{https://github.com/athanor/athanor/releases/tag/release_v0.9.9}}

\athanor makes use of high-level structure in two ways: variables are represented internally in a compact form, rather than flattened to a lower-level representation; and neighbourhood structures are automatically created from neighbourhood templates and high-level types of decision variables. To investigate these two ideas separately, \athanorreduced is a version of \athanor with many high-level neighbourhood templates removed. \athanorreduced contains only the atomic neighbourhood templates, neighbourhood templates to add or remove a value for variable-sized containers, and \texttt{LiftSingle}, so atomic templates can be applied to members of higher-order types. It also contains all the \emt|partition| templates, except \texttt{SwapParts}.

\subsection{Essence Specifications and Constraint Models}\label{sec:choco-mzn-models}

Each problem class used in the experiments has an \essence specification (used by \athanor and SNS) and two constraint models: a MiniZinc model (used by Yuck, \fznoscar, Chuffed, and OR-Tools); and a Choco model (for LNS-PG and LNS-EB, written in Java using the Choco API). To ensure (as far as possible) a fair comparison of all solvers, we have chosen or written constraint models that correspond closely to the \essence specifications. In each case, we describe how the abstract decision variables in the \essence specification are represented in the MiniZinc and Choco models. We also ensure that the MiniZinc and Choco models are as close as possible given the differences between the two systems.

In some cases the representation of an abstract decision variable introduces symmetry (such as representing a set of $\tau$ as a matrix of $\tau$, for any \essence type $\tau$). For each problem where this occurs, we have two versions of the MiniZinc and Choco models: one with symmetry-breaking constraints and the other without.  All experiments are conducted with both versions of the model, because we do not know in advance whether symmetry-breaking constraints will help or hinder a given solver. 

The MiniZinc and Choco models use common global constraints such as allDifferent and element. In this set of experiments, we do not utilise specialised global constraints that encapsulate an entire problem or a key part of it (such as knapsack~\cite{trick2003dynamic} and bin packing~\cite{shaw2004constraint} constraints). The propagators for global constraints such as knapsack and bin packing are essentially special-purpose solvers for a problem class, and our aim is to compare the general-purpose solver \athanor to other general-purpose solvers. In~\Cref{sec:experiment-globcons}, we conduct a separate set of experiments where those specialised global constraints are employed.

Both MiniZinc and Choco support set variables (representing a set of integers). We use set variables in MiniZinc for any \essence variable with type \texttt{set of int}. However, in the Choco models we represent \texttt{set of int} with an array of Boolean variables (representing the characteristic function of the set). Decomposing set variables gives the LNS methods the ability to search parts of the set representation while freezing the rest, and also avoids having a single variable in the model of the Knapsack problem. 

All MiniZinc and Choco models minimise or maximise an integer variable \texttt{optVar}, the value of which is a function of the values of primary decision variables. Choco requires a single variable to optimise and we follow the same convention in MiniZinc for consistency.

\subsection{Benchmark Problems}\label{sub:problems}

We benchmark \athanor against other solvers on seven problem classes. As detailed in the subsequent sections, we make use of standard benchmark instance sets whenever they are available. For problems where there are only a small number of instances available, or the existing instances are trivially solved by the solvers considered in our experiments, we create new benchmark instance generators and randomly generate new instances for the evaluation. All models, generators and instances used in our experiments are provided in the accompanying repository.

\subsubsection{Bin Packing}

\begin{figure}
\begin{lstlisting}
given items new type enum
given weights : function (total) items --> int
given binSize : int
find packing : partition from items
minimising |parts(packing)|
such that  forAll p in parts(packing) .
    binSize >= sum i in p . weights(i)
\end{lstlisting}
\caption{\essence specification of the bin packing problem.}
\label{fig:binpacking-spec}
\end{figure}

The specification of the classic bin packing problem is shown in \Cref{fig:binpacking-spec}. Each item must be allocated to exactly one bin, so we are able to use the \texttt{partition} type to capture the entire problem in a single decision variable. The optimisation function (to be minimised) is the number of parts (cells) in the partition. The constraints state that the weight limit of each bin is respected.

We use a 0/1 model by Håkan Kjellerstrand\footnote{\url{https://github.com/hakank/hakank/blob/master/minizinc/bin_packing_me.mzn}} for the Choco and MiniZinc solvers. The model extends the basic model of R\'egin and Rezgui \cite{regin2011discussion}. For each item and bin, the model has a 0/1 variable indicating whether the item is placed in the bin. Also, for each bin, we have a load variable that is the sum of weights of items in the bin. The upper bound of each load variable is the weight limit of the bins. Constraints ensure that each item is packed exactly once, and an implied constraint states that the sum of the load variables is equal to the total weight of items. Symmetry breaking constraints place the load variables in non-increasing order.
The optimisation function (to minimise) is the number of load variables that are greater than 0.

We use $80$ instances from a set of $160$ generated by Falkenauer~\cite{falkenauer-binpack} and available from the  OR-Library~\cite{beasley1990or}.\footnote{\url{http://people.brunel.ac.uk/~mastjjb/jeb/orlib/binpackinfo.html}} Falkenauer generated $8$ sets of $20$ instances; from each set we took the 10 even-numbered instances. The number of items in this instance set range from $60$ to $1000$.

\subsubsection{Travelling Salesperson Problem}

\begin{figure}
\begin{lstlisting}
given nCities : int
given distances : function (total)
    tuple (int(1..nCities), int(1..nCities)) --> int
letting maxDistance be max([i | ((_,_), i) <- distances])
find tour : sequence (size nCities, injective)
    of int(1..nCities)
minimising sum i : int(2..nCities) .
    distances((tour(i-1),tour(i)))
    +  distances((tour(nCities),tour(1)))
\end{lstlisting}
\caption{\essence specification of the Travelling Salesperson Problem}
\label{fig:tsp-spec}
\end{figure}

In the classic Travelling Salesperson Problem (TSP) \cite{flood1956traveling} we are given a set of \(n\) cities, and the cost of travelling between each pair of cities. The objective is to find the lowest-cost cycle that visits all cities exactly once. The \essence specification is given in \Cref{fig:tsp-spec}. The cycle is represented using a single decision variable with a sequence domain of fixed length. Notice that no constraints are needed because the injective attribute on the sequence requires all elements to be distinct, capturing the main constraint of the TSP. The objective function is simply the sum of the costs of each journey between two cities within the tour.

For Choco and MiniZinc solvers we use a model that matches the \essence specification as closely as possible. The model has an array of integer decision variables to represent the sequence, and an allDifferent constraint to ensure each city is visited exactly once. In MiniZinc the objective remains the same as in \Cref{fig:tsp-spec} (except that an optimisation variable is introduced as described in \Cref{sec:choco-mzn-models}). In Choco each two-dimensional matrix lookup is translated to a one-dimensional lookup using an \texttt{element} constraint with additional index and result variables.

In our experiments we use a subset of the TSPLIB set of symmetric instances~\cite{reinelt1995tsplib95}.
We extract instances with $1000$ or fewer cites ($68$ instances in total), as instances of larger sizes could not go through MiniZinc within the time and memory limit (more details in~\Cref{sec:expdetails}).

\subsubsection{Capacitated Vehicle Routing Problem}

\begin{figure}
\begin{lstlisting}
given n : int(1..) $ number of locations
letting L0 be domain int (0..n) $ 0 is the depot
letting L1  be domain int(1..n)
given weights : function (total) L1 --> int(1..)
given costs : function (total)  tuple (L0,L0) --> int(0..)
given cap : int(1..) $ vehicle capacity
letting totalW be sum([weight | (_,weight) <- weights])
letting mV be totalW/cap+toInt(totalW%cap != 0) $ lowerbound
find plan : set (minSize mV, maxSize n) of
    sequence (maxSize n, injective, minSize 1) of L1
minimising sum r in plan .
  (sum([costs(tuple(r(i-1), r(i))) | i : int(2..n), i<=|r|])
  + costs((0, r(1))) + costs((r(|r|), 0)))
such that forAll route in plan .   $ vehicle capacity
  (sum (_,order) in route . weights(order)) <= cap,
$ all orders delivered once
allDiff([loc | route <- plan, (_,loc) <- route]),
(sum p in plan . |p|) = n
\end{lstlisting}
\caption{\essence specification of the Capacitated Vehicle Routing Problem.}
\label{fig:cvrp-spec}
\end{figure}

The Vehicle Routing Problem (VRP) without capacities \cite{toth2002vehicle} is a variation of TSP with multiple tours (performed by a set of vehicles) that all include one depot location. Each location must be visited exactly once, with the exception of the depot. Capacitated vehicle routing (CVRP) adds a weight to each location, and a capacity that applies to each vehicle separately. For each vehicle, the sum of weights of its visited locations must not exceed the capacity. The \essence specification of CVRP is shown in \Cref{fig:cvrp-spec}. The decision variable is a set of sequences, where each sequence contains locations (and the depot is implicitly the first and last location). Constraints ensure that vehicles do not exceed their capacities, and that all locations are visited by one vehicle. 

The models used by Choco and MiniZinc solvers are derived from the \essence specification in \Cref{fig:cvrp-spec}. The outer type of the \texttt{plan} variable is a set with maximum size \(n\) (the number of locations). The set is represented by a matrix named \texttt{planMat} with rows indexed \(1\ldots n\), where some rows may be unused in any given solution. The inner type of \texttt{plan} is a sequence of maximum length \(n\), and each row of \texttt{planMat} directly represents one sequence. A row is indexed \(0\ldots n+1\), and the first and last variables are fixed to zero (representing the depot). For each row, an additional variable is declared for the length of the sequence. Variables in the inactive part of each row in \texttt{planMat} are fixed to zero, and unused rows are fixed to zero. An \texttt{alldifferent\_except\_0} global constraint is applied to \texttt{planMat}, and the sequence lengths are constrained to ensure each location is visited exactly once. Symmetry-breaking constraints order the first elements of the sequences.

We use $88$ instances of CVRP available from VRP-REP~\cite{mendoza2014vrp} that have size $n \leq 100$.\footnote{Instances of larger sizes either could not go through MiniZinc within the time and memory limits, or result in a memory exception for several solvers called via MiniZinc.}\footnote{The instances are from Augerat 1995 -- Sets A, B, and P, Christofides and Eilon 1969 -- Set E., Fisher-1994-Set-F, Christofides-et-al.-1979, Christofides-et-al.-1979-Set-M}

\subsubsection{Knapsack Problem}

\begin{figure}
\begin{lstlisting}
given items new type enum
given gain : function (total) items --> int
given weight : function (total) items --> int
given capacity : int
find picked : set of items
maximising sum i in picked . gain(i)
such that (sum i in picked . weight(i)) <= capacity
\end{lstlisting}
\caption{\essence specification of the knapsack problem.}
\label{fig:knapsack-spec}
\end{figure}

The classic knapsack problem is straightforward to state in \essence (\Cref{fig:knapsack-spec}). The single decision variable is the set of items in the knapsack, and the constraint enforces the weight capacity of the knapsack. The optimisation function is to maximise the sum of the values of items in the knapsack.
The MiniZinc model has one decision variable of type \texttt{set of int} and closely follows \Cref{fig:knapsack-spec} except that a second variable is introduced representing the optimisation value.  The Choco model represents the set with an array of Boolean variables, one per item.

We sampled 54 instances from a benchmark set generated by Pisinger~\cite{pisinger05}.\footnote{Available from \url{www.diku.dk/~pisinger/hardinstances_pisinger.tgz}} The full benchmark set has subsets of 100 instances for each one of 54 generator configurations. We took instance number 50 from each subset. The number of items in our selected instances ranges from 20 to 10,000.

\subsubsection{Minimum Energy Broadcast}

\begin{figure}
\begin{lstlisting}
given n: int(1..)
given initNode: int(1..n)
letting dNodes be domain int(1..n)
given linkCosts:
    function (total) (dNodes,dNodes) --> int(0..)
find parents: function (total) dNodes --> dNodes
find depths: function (total) dNodes --> int(1..n)
minimising sum parent : dNodes .
  max([ linkCosts((parent,child)) * toInt(parentI = parent)
        | (child,parentI) <- parents])
such that
parents(initNode) = initNode,
forAll (child,parent) in parents .
  (child != initNode) ->
    (parent != child /\ linkCosts((parent,child)) != 0),
forAll (child,parent) in parents .
  (child != initNode) -> depths(child) = depths(parent) + 1
\end{lstlisting}
\caption{\essence specification of the Minimum Energy Broadcast problem.}
\label{fig:meb-spec}
\end{figure}

The Minimum Energy Broadcast (MEB) problem~\cite{meb,csplib:prob048} is to connect a set of wireless devices (nodes) to form a broadcast network while minimising the energy required to broadcast a message. The structure of the network is a tree where each non-leaf node is responsible for broadcasting messages to its set of children, thus a message broadcast by the root will eventually reach all nodes. The root node is given as a parameter, and each pair of nodes has a given link cost. The objective is to minimise the total energy to broadcast a message, where the energy used by a node is the maximum of the set of link costs between the node and its children.

The \essence specification of MEB is shown in \Cref{fig:meb-spec}. The specification has a total function representing the parent relationship of the tree, and it ensures acyclicity using a second total function representing the depth of nodes in the tree. Constraints connect the depth of each non-root node to the depth of its parent. The MiniZinc and Choco models closely follow the \essence specification. The total functions are represented straightforwardly using arrays (indexed by the function domain) of integer decision variables. Although the \essence specification is similar to the MiniZinc and Choco models, MEB allows us to evaluate whether neighbourhoods designed specifically for functions can outperform generic neighbourhoods applied to arrays of integer variables.

We use a set of 56 instances from previous work~\cite{attieh2019athanor}. 50 of the instances were created with a random instance generator that has 4 parameters: the length of the square area where the devices are placed, the number of devices, the maximum power each device can broadcast at, and the rate at which the radio signal attenuates. The generator is parameterised according to the description provided in~\cite{meb,csplib:prob048}. The maximum power determines whether two devices can communicate with each other, while the link cost between each pair of devices is determined by the Euclidean distance between them and the attenuation rate. The parameters of the generator were tuned with the automatic algorithm configuration tool irace~\cite{lopez2016irace} to find instances of appropriate difficulty. The random instances have between 73 and 297 nodes (within the range 70 to 300 imposed on the generation process).
The remaining $6$ instances were sampled at random from an existing set provided in~\cite{SNS-IJCAI18}, those instances were of smaller sizes, where the number of devices is in the range $20$ to $60$. All instances and the instance generator are provided in the experimental repository.

\subsubsection{Progressive Party Problem}
\label{subsec:ppp_problem}

\begin{figure}
\begin{lstlisting}
given nBoats, nPeriods : int(1..)
letting Boat be domain int(1..nBoats)
given capacity, crew : function (total) Boat --> int(1..)
find hosts : set (minSize 1) of Boat
find sched : set (size nPeriods) of partition from Boat
minimising |hosts|
$ Hosts stay on their own boat
such that forAll p in sched . |parts(p)| = |hosts| /\
    forAll part in parts(p) . |part intersect hosts| = 1,
$ Host boats have sufficient capacity
forAll p in sched . forAll h in hosts .
    (sum b in party(h,p) . crew(b)) <= capacity(h),
$ Pairs of crews that meet in the schedule are all distinct
allDiff([ (i,j) | p <- sched, part <- p, i,j <- part, i<j ])
\end{lstlisting}
\caption{\essence specification for the Progressive Party Problem.}
\label{fig:ppp_spec}
\end{figure}

The Progressive Party Problem (PPP) \cite{smith1996progressive,csplib:prob013} is to arrange a social event involving boat crews. Given a set of boats (each with a capacity and crew size) and a number of periods, a set of host boats are chosen and a schedule is designed where the crews of non-host boats visit the host boats over several periods. In each period, every non-host boat will visit one of the host boats. The capacity of the host boats must be respected, and no pair of boat crews are allowed to meet more than once.

The \essence specification for \athanor is shown in \Cref{fig:ppp_spec}.
The key decision variable is \texttt{sched}, which is a set of partitions of the boats, with one partition for each time period. Within each partition, each part contains exactly one host boat. For each period and each host, the total number of crew members on-board the host boat is bounded by the capacity of the host. The final constraint states that every ordered pair \texttt{(i,j)} of crews that meet is distinct throughout the schedule. Unfortunately we cannot use the same specification for SNS because it does not support enough of the current \essence language. For SNS we use a less compact specification (available in the experimental repository) where the \texttt{allDiff} constraint is decomposed.

For MiniZinc and Choco we use a standard model from the literature \cite{leo-globalizing-aij,smith1996progressive}. In MiniZinc the main decision variables are the set of hosts (exactly as in \Cref{fig:ppp_spec}) and an array named \textit{visit}, indexed by boat \(b\) (row) and time period \(t\) (column), representing the host boat visited by the crew of \(b\) at time \(t\). Hosts always visit themselves, and other boats are required to visit hosts. For each pair of crews, they meet at most once (i.e. are assigned the same host at most once) so we have a \(\mathrm{sum} \leq 1\) constraint.  Symmetry breaking constraints order the columns of \textit{visit}. The Choco model is very similar but explicitly adds Boolean variables for each pair of boats \(b_1,b_2\) and time period \(t\) indicating whether \(visit[b_1, t]=b_2\), required for the capacity constraints. In Choco the \texttt{hosts} variable is represented with an array of Boolean variables indicating whether each boat is in the set.

There are a small number of existing benchmark instances available for this problem. Therefore, we created a new instance generator based on the problem description. 
We set the number of boats, number of periods, and maximum boat capacity (\(\mathit{maxC}\)) to be within the ranges $[10, 80]$, $[5, 30]$, and $[10, 100]$, respectively (each chosen at random with uniform distribution). The capacity and crew size of each boat are generated at random with uniform distribution in the range [$0.3 \times maxC$, $0.8 \times maxC$].
We generated $500$ random instances with the given parameter ranges and used the OR-Tools solver to identify unsatisfiable instances. The remaining instances (including the ones that were not solved by OR-Tools within the time and memory limits given in \Cref{sec:expdetails}) were taken and $54$ instances were randomly sampled from the set. We also sampled $6$ instances at random from an existing benchmark set used elsewhere~\cite{SNS-IJCAI18}. The total number of instances used for this experiment is $60$.

\subsubsection{Synchronous Optical Networking}
\label{subsec:sonet_problem}

The Synchronous Optical Networking (SONET) problem \cite{smith2005symmetry,csplib:prob056} is to design a fibre-optic network comprised of multiple \textit{rings}, where two network nodes are connected if there exists a ring that both nodes are connected to.  Each instance has a set of demand pairs \((i,j)\) where node \(i\) must be connected to node \(j\) on at least one ring. The objective is to minimise the total number of node to ring connections.  We use the simplest variation of the problem, where each ring has an upper bound on the number of nodes connected to it, but there is no upper bound on the amount of network traffic carried by a ring.
The \essence specification of SONET is shown in \Cref{fig:sonet-spec}. The decision variable \texttt{network} is a set of sets of nodes, where each inner set represents a ring. The constraint ensures that each demand pair is a subset of at least one of the rings. The capacity constraint on the rings is expressed as a domain attribute of \texttt{network} (namely \texttt{maxSize capacity}).

The MiniZinc model uses an array of decision variables of type \texttt{set of int} to represent the rings. The demand constraint for each demand pair \((i,j)\) states that there exists a ring that is a superset of \(\{i,j\}\). The capacity constraints and the objective are straightforwardly stated using the cardinality of sets. Symmetry-breaking constraints order the sets in the array. The Choco model is similar to the MIP model of Sherali et al~\cite{sonet-mip-sherali}. It represents each ring with an array of Boolean variables (indicating whether each node is in the ring), and also introduces variables for the cardinality of each ring. Demand constraints are represented with Boolean logic, while the capacity constraints and objective are stated on the cardinality variables.

There are a number of SONET instances available in CSPLib~\cite{csplib:prob056}. However, those instances are of very small sizes ($7$ to $13$ nodes) and are mostly trivially solvable by the solvers considered in our experiments. Therefore, we generated 50 random instances with number of nodes \(\mathit{nNodes} \in \{30,40,\ldots,120\}\), between $15$ and $90$ rings, and with the capacity ranging from $15$ to $90$. For each unordered pair of nodes, the pair is chosen to be a demand pair with probability 0.5.  We also include 8 existing instances~\cite{SNS-IJCAI18} (the parameters of those instances are within similar ranges to the newly generated ones). In total, there are $58$ instances for the first experiment.

\subsection{Experimental Details}\label{sec:expdetails}

For each solver, model, and problem instance, we ran the solver 10 times with 10 distinct random seeds, and with a time limit of 10 minutes for each run. The elapsed time and objective value were recorded whenever a solver found a new best solution. All experiments were run on the compute nodes of the Cirrus High Performance Computing cluster.\footnote{\url{https://www.cirrus.ac.uk/}} Each node has two 2.1 GHz, 18-core Intel Xeon E5-2695 processors. 
Each solver is given a memory limit of 7GB, imposed using the \texttt{runsolver} tool~\cite{roussel2011controlling}.\footnote{\url{https://www.cril.univ-artois.fr/~roussel/runsolver/runsolver-3.4.1.tar.bz2}}

The MiniZinc Challenge incomplete scoring method\footnote{\url{https://www.minizinc.org/challenge/2020/rules/}} was used to compute a performance score for each problem class and solver (aggregating instances and runs of each instance). The scoring method is based on the quality of solutions found and the time taken to find them, but takes no account of proof of optimality by a systematic solver.

The MiniZinc Challenge incomplete score is computed from the quality $q(x,i,j)$ of the best solution found (within the time limit) by solver $x$ on instance $i$ and run $j$, and the time $t(x,i,j)$ taken by solver $x$ on instance $i$ and run $j$ to find the best solution. Given a problem class, each pair of solvers $s$ and $s'$ are scored on every problem instance $i$ and run $j$. Solver $s$ is awarded $1$ if $q(s,i,j)$ is better than $q(s',i,j)$; $0$ if $q(s,i,j)$ is worse than $q(s',i,j)$; or in the case that $q(s,i,j) = q(s',i,j)$, $\frac{t(s',i,j)}{t(s',i,j) + t(s,i,j)}$. If $s$ failed to find a solution then $s$ is awarded $0$ points, regardless of whether \(s'\) found a solution. For each solver $s$, its total score is the sum of its scores relative to every other solver $s'$ for every problem instance and run.

Note that the score depends critically on the time limit. To show how the relative performance of solvers evolve over time, we calculated the score for every integer time limit in the range \(\{1\ldots 600\}\). The score of a solver $s$ can either increase or decrease over time based on its progress relative to the other solvers.

\subsection{Experiment 1: Evaluation of Athanor Neighbourhoods} \label{sec:experiments-all-solvers}

The first experiment compares \athanor to the other eight solver configurations (described in \Cref{sec:othersolvers}) using all seven benchmark problems. In this experiment we are testing the first hypothesis: that \athanor will generate effective neighbourhoods and neighbourhood structures from the high-level structure available in the \essence specifications. If the hypothesis is true, we expect to see \athanor performing well compared to the other solvers in general, but particularly when the \essence specification contains nested structures such as a \emt|set of set| that allow powerful neighbourhood structures to be constructed.

\subsubsection{Symmetry Breaking}\label{sec:symbreak-results}

First we determined whether to include symmetry-breaking constraints for the four problem classes where they are available (Bin Packing, CVRP, PPP, and SONET), and when using a solver other than \athanor or SNS (i.e.\ any solver using a MiniZinc or Choco model). The symmetry-breaking constraints are described in the relevant subsections of \Cref{sub:problems}. For each solver and problem class we computed the score with a time limit of 600 seconds, with and without the symmetry-breaking constraints.
\Cref{tab:sym-or-nosym} shows the outcome for each of the four problem classes and each relevant solver.

For all subsequent experiments, we show results only for the version with the better score as reported in \Cref{tab:sym-or-nosym}. However, both versions (with symmetry-breaking constraints and without) are included in each competition and therefore both versions affect the scores of other solvers. For example, when comparing \athanor to systematic solvers (OR-Tools and Chuffed) on the PPP, the scores are calculated from all five solver and model combinations and the results are reported for three: \athanor{}, Chuffed without symmetry-breaking, and OR-Tools with symmetry-breaking.

\begin{table}[]
\begin{tabular}{lllllll}
\toprule
           & Chuffed & LNS-EB  & LNS-PG  & OR-Tools & \fznoscar & Yuck   \\
\midrule
Bin Packing & Sym     & ---    & ---    & Sym      & ---         & ---    \\
CVRP        & Sym     & Sym    & ---    & Sym      & ---         & ---    \\
PPP         & ---     & Sym    & Sym    & Sym      & ---         & ---    \\
SONET       & ---     & ---    & ---    & ---      & ---         & ---    \\
\bottomrule
\end{tabular}
\caption{Whether symmetry-breaking constraints are valuable, for each problem class where they are available and for each solver that uses a MiniZinc or Choco model. \textbf{Sym} indicates that the model with symmetry-breaking constraints outperforms the model without them. }
\label{tab:sym-or-nosym}
\end{table}

\subsubsection{Local Search Solvers}

\begin{figure}[t]
\begin{center}
\caption{Performance of \athanor compared to other local search solvers using the scoring system described in \Cref{sec:expdetails}. Symmetry is broken for some combinations of problem class and solver (as described in \Cref{sec:symbreak-results}). Higher scores indicate better relative performance.}
\label{fig:experiments-athanor-vs-local}
\includegraphics[width=\columnwidth]{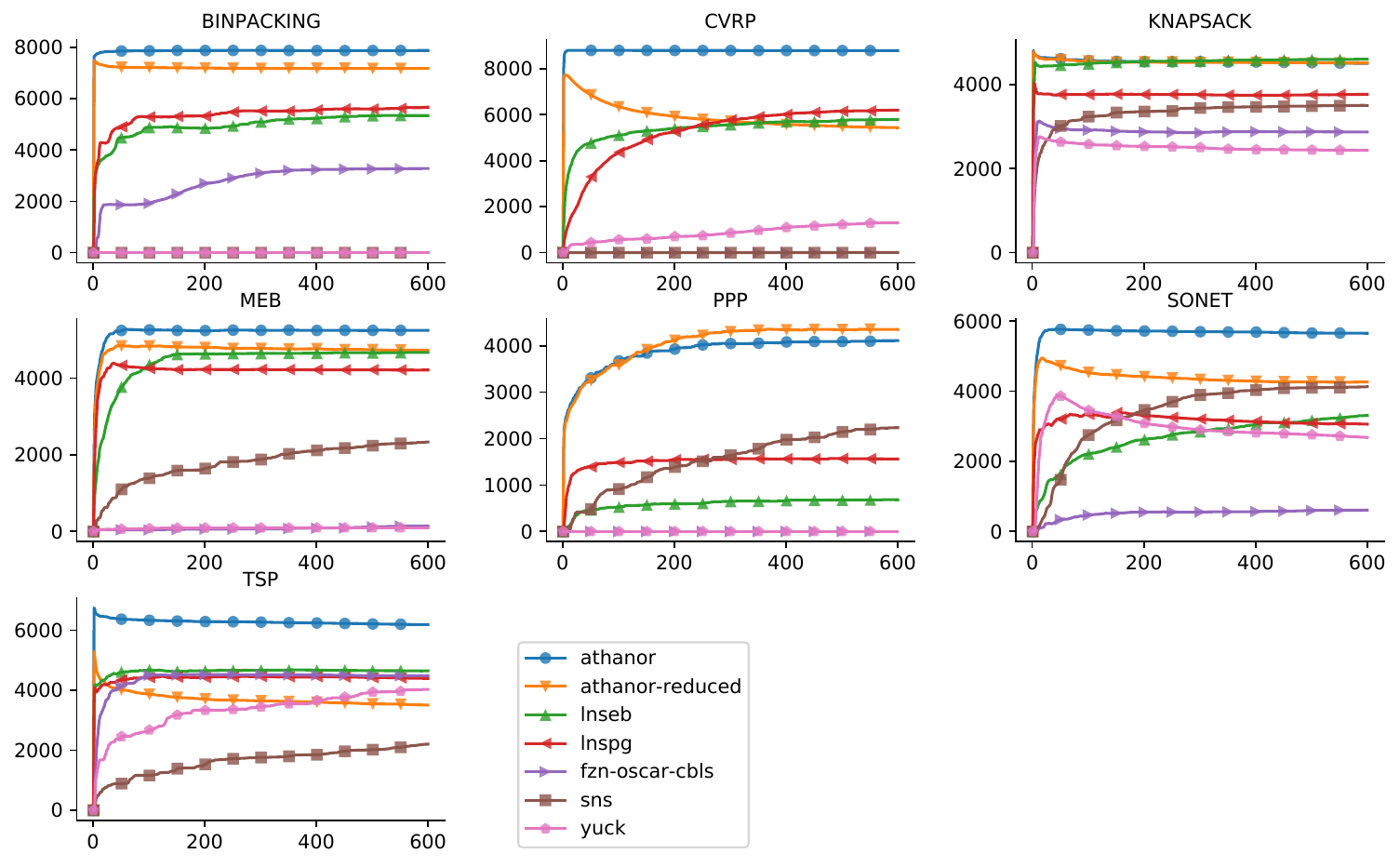}
\end{center}
\end{figure}

\begin{figure}[t]
\begin{center}
\caption{Cumulative plots of the number of runs for which a feasible solution (of any quality) has been found, for each solver and each problem class.}
\label{fig:solutions-non-large}
\includegraphics[width=\columnwidth]{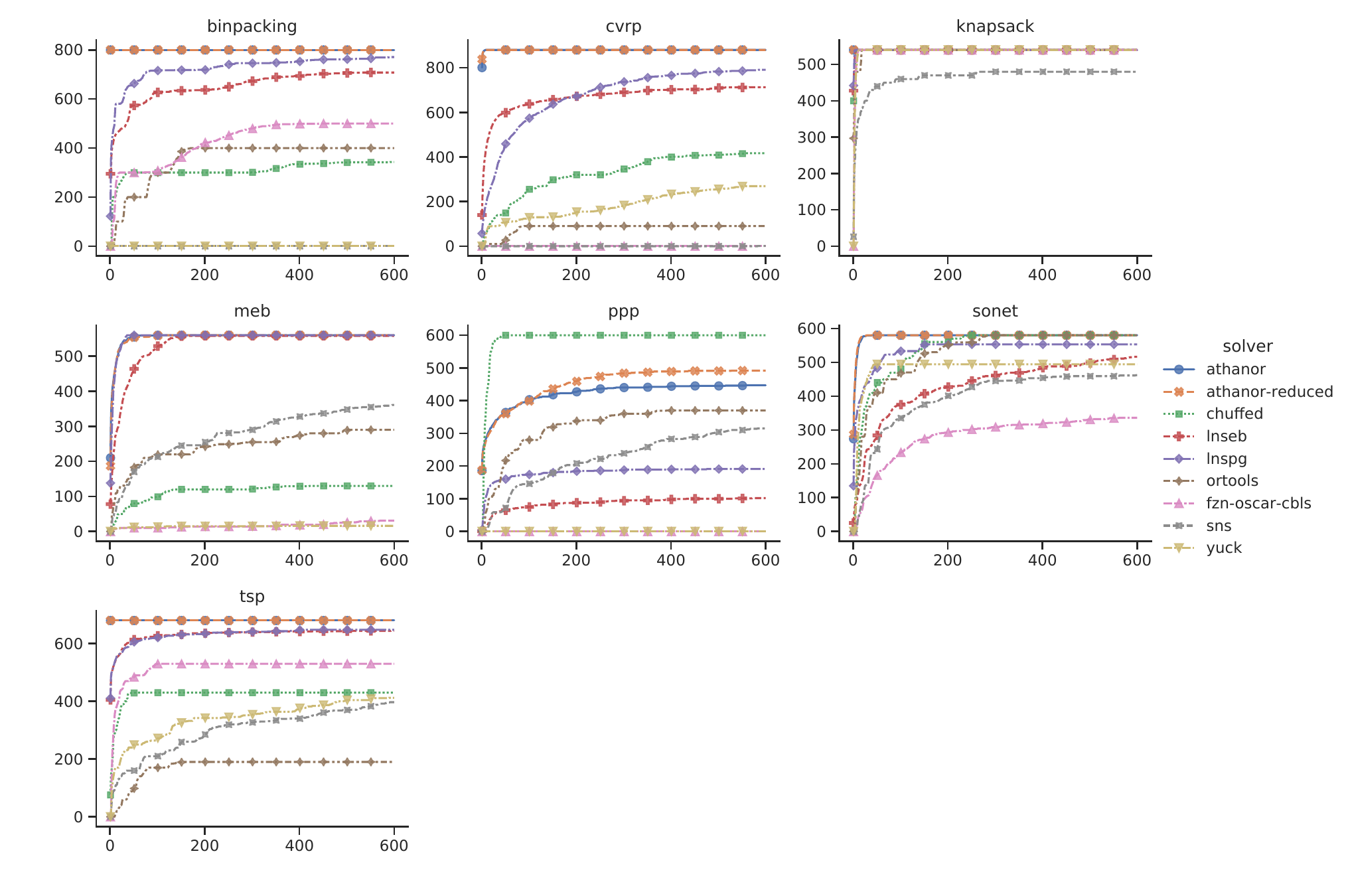}
\end{center}
\end{figure}

\begin{figure}[tbp]
\centering
    \begin{subfigure}[b]{0.49\textwidth}
         \centering
         \caption{Bin Packing}
         \includegraphics[trim={0 1.8cm 0 0.8cm},clip,height=3.2cm]{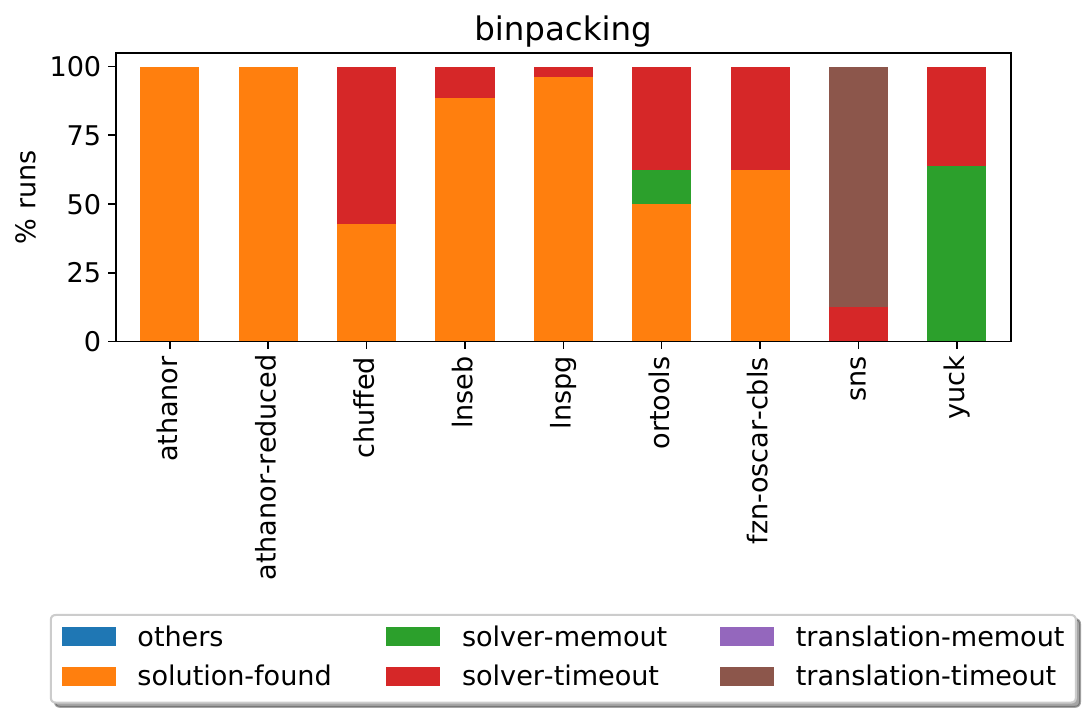}
     \end{subfigure}
     \begin{subfigure}[b]{0.49\textwidth}
         \centering
         \caption{CVRP}
         \includegraphics[trim={0 1.8cm 0 0.8cm},clip,height=3.2cm]{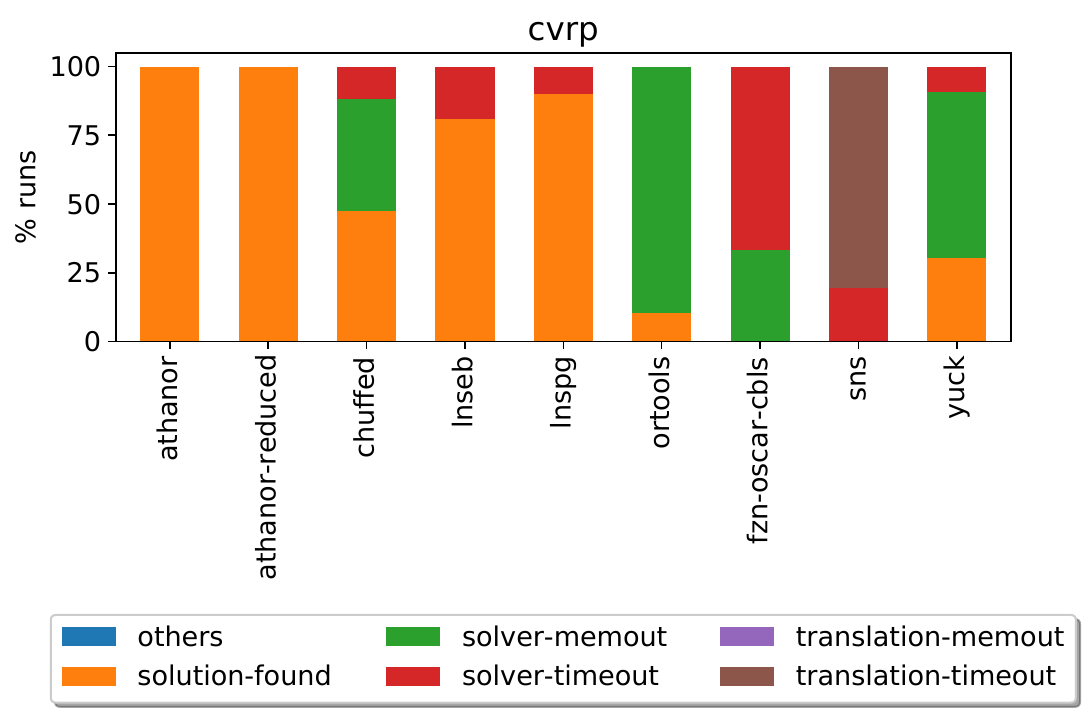}

     \end{subfigure}

     \begin{subfigure}[b]{0.49\textwidth}
         \centering
         \caption{Knapsack}
         \includegraphics[trim={0 1.8cm 0 0.8cm},clip,height=3.2cm]{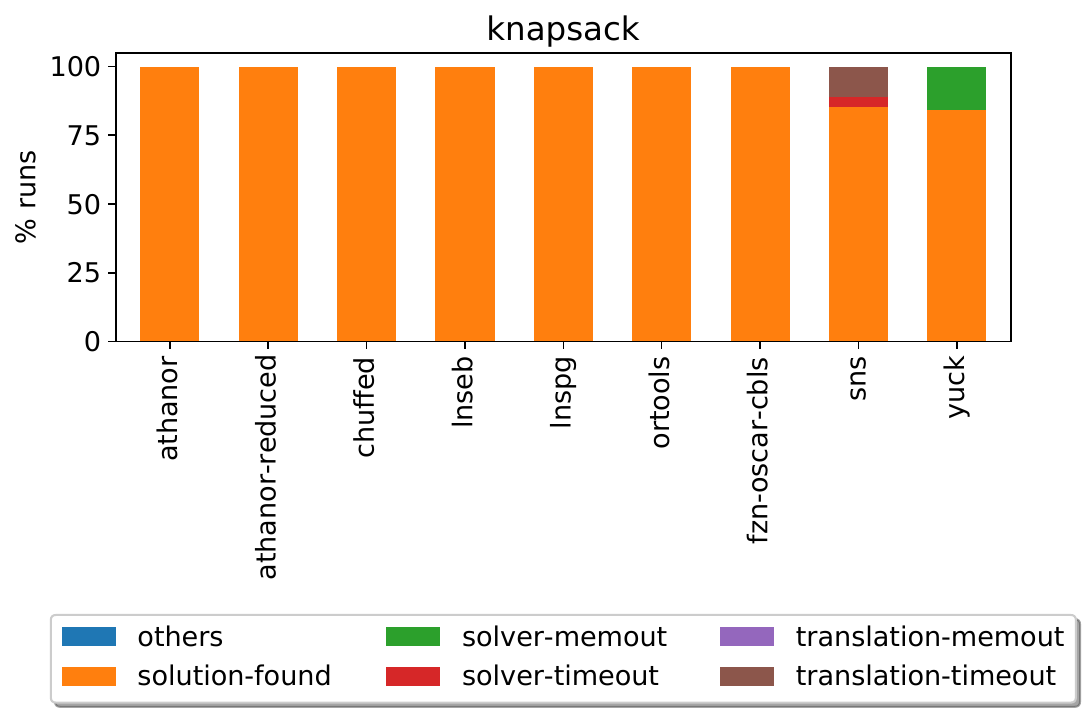}

     \end{subfigure}
     \begin{subfigure}[b]{0.49\textwidth}
         \centering
         \caption{MEB}
         \includegraphics[trim={0 1.8cm 0 0.8cm},clip,height=3.2cm]{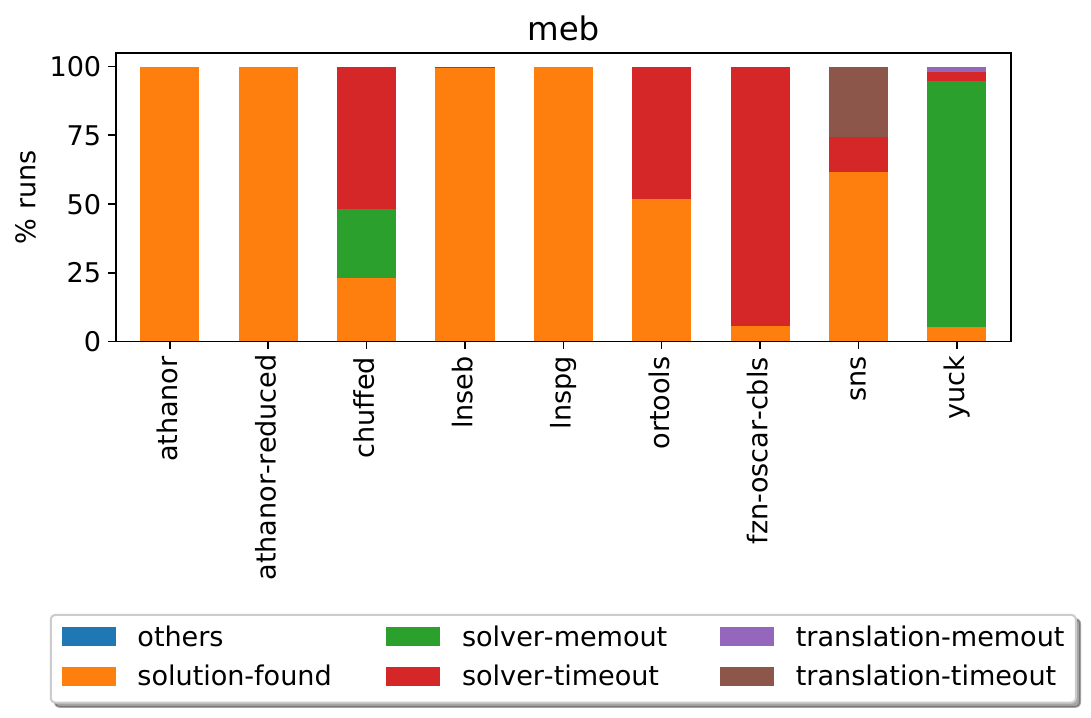}

     \end{subfigure}

     \begin{subfigure}[b]{0.49\textwidth}
         \centering
         \caption{PPP}
         \includegraphics[trim={0 1.8cm 0 0.8cm},clip,height=3.2cm]{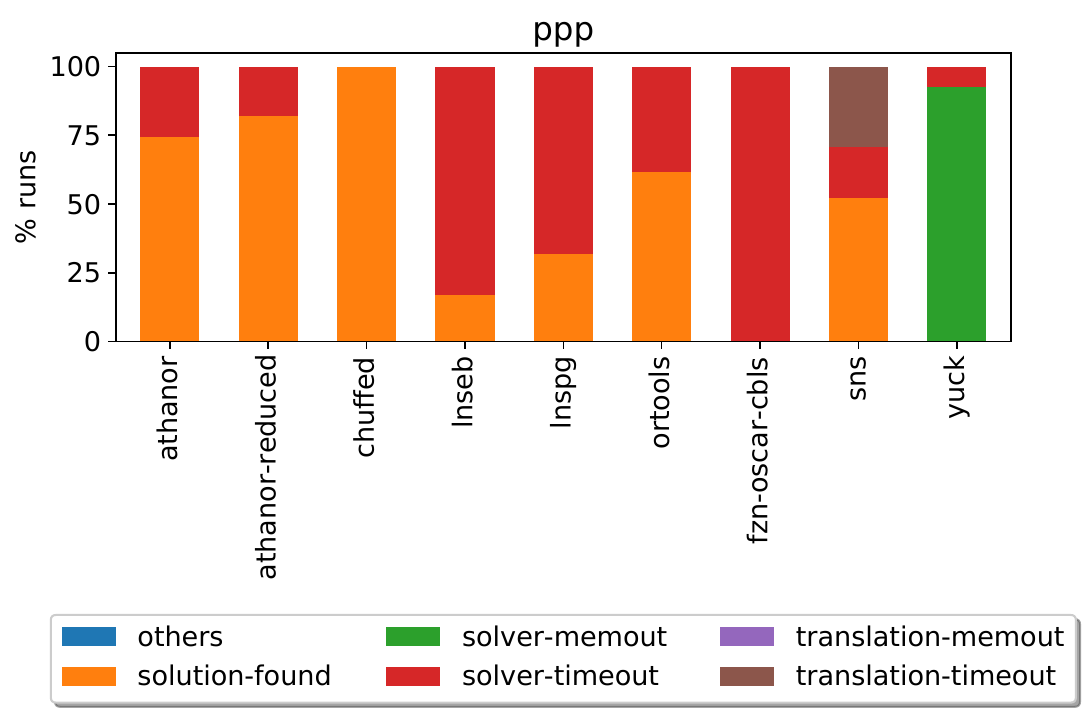}

     \end{subfigure}
     \begin{subfigure}[b]{0.49\textwidth}
         \centering
         \caption{SONET}
         \includegraphics[trim={0 1.8cm 0 0.8cm},clip,height=3.2cm]{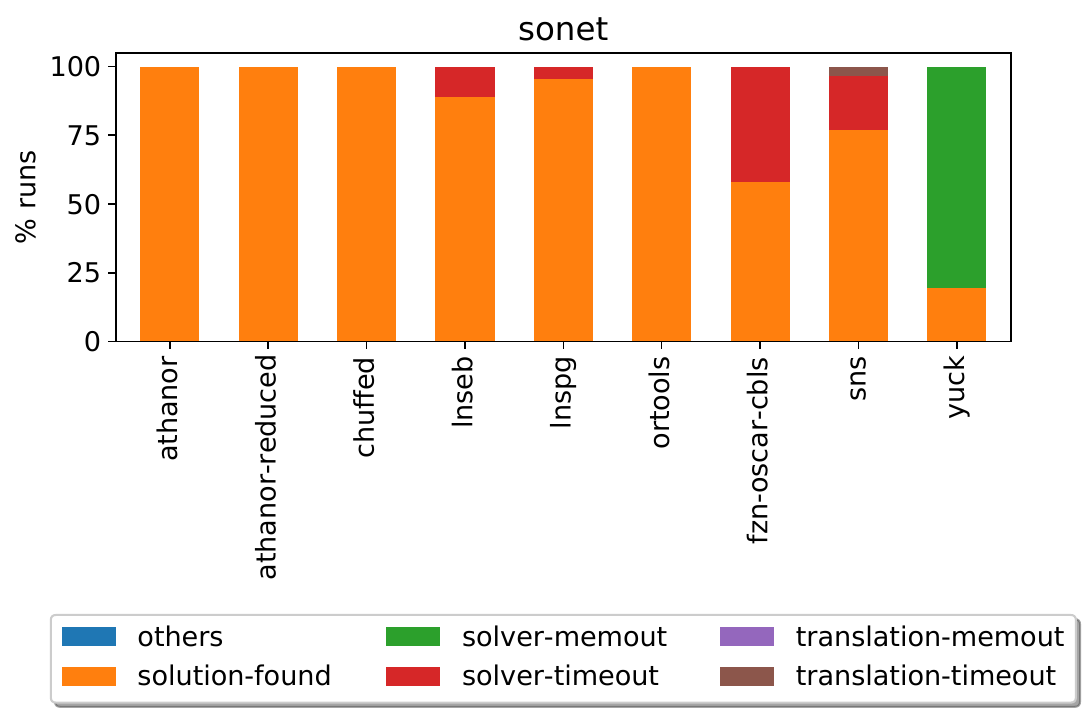}

     \end{subfigure}

\begin{subfigure}[b]{0.49\textwidth}
         \centering
         \caption{TSP}
         \includegraphics[trim={0 1.8cm 0 0.8cm},clip,height=3.2cm]{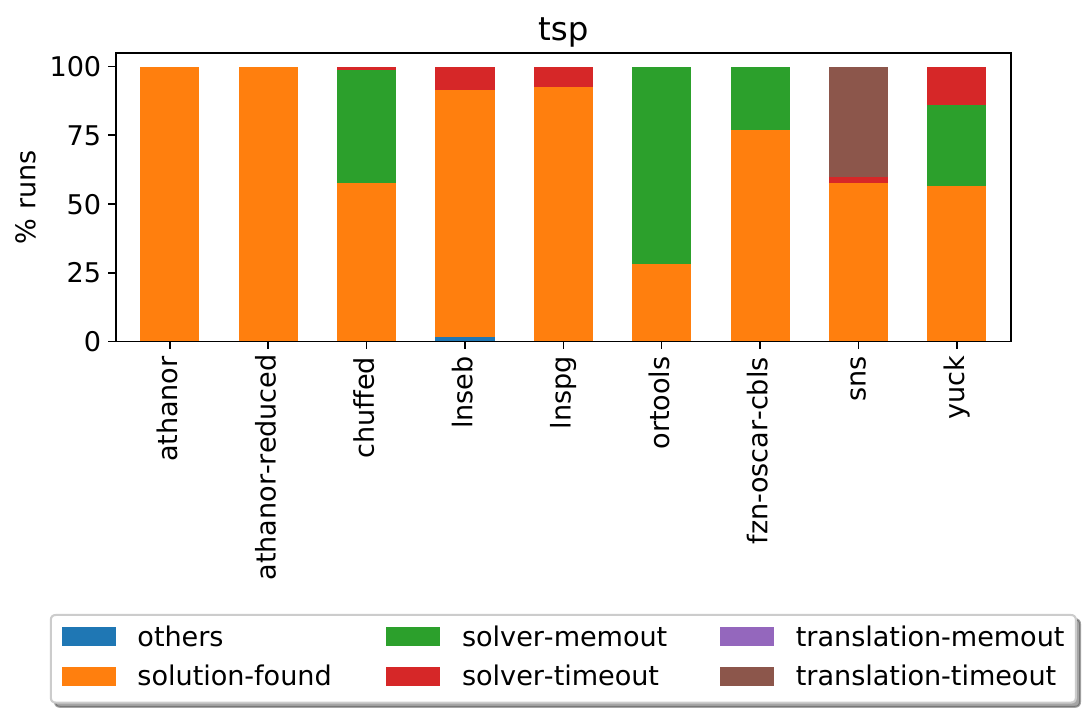}

     \end{subfigure}

\centering
\includegraphics[trim={0 0 0 10cm},clip,width=0.7\textwidth]{figures/revised-version/run-status-sonet-edited.pdf}
\caption{Status of solver runs for each problem class and each solver. Note that \texttt{solution-found} indicates that a solver finds a feasible solution within the time limit and that it does not exceed the memory limit during the run.}
\label{fig:experiments-runstatus}
\end{figure}

In \Cref{fig:experiments-athanor-vs-local} we compare the performance of \athanor to the local search solvers SNS, Yuck, \fznoscar, LNS-EB, LNS-PG, and \athanorreduced (as described in \Cref{sec:othersolvers}). Also, \Cref{fig:solutions-non-large} plots the time taken to find the first feasible solution (regardless of quality) for each solver and each problem class as a cumulative curve. \Cref{fig:experiments-runstatus} plots the status of solver runs (i.e.\ whether a feasible solution was found, and if not how the solver failed) for each problem class and each solver.
We will discuss the problems with simpler (non-nested) \essence types first.

The Knapsack model uses one of the simplest types in \essence{}: a set of integers. Even so, \athanor is able to perform well compared to other solvers. Both \athanor and LNS-EB achieve the best overall performance on this problem, which indicates the effectiveness of the local search algorithm adopted by \athanor (independent of the neighbourhood derivation contribution). 
The LNS-EB version of Choco LNS also performed very well, and slightly outperformed \athanor at the end of the 10 minutes. Surprisingly, the CBLS solvers Yuck and \fznoscar are less strong, despite supporting sets of integers natively. All solvers except SNS find a first feasible solution rapidly (\Cref{fig:solutions-non-large}).
SNS is hindered by the overhead of representing neighbourhood structures as part of the constraint model, and for some runs the translation process (instantiating the model and neighbourhood structures) timed out at 600s (\Cref{fig:experiments-runstatus}). Due to the simplicity of this problem, \athanor and \athanorreduced have an identical set of neighbourhoods and therefore their performance is the same (with very marginal difference due to fluctuation in time measurement).

The MEB specification consists of two total functions that are tightly constrained together. As with Knapsack, \athanor finds good solutions quickly and performs comparatively well over the 10 minute timespan. Compared with Knapsack, MEB has a more complex type (i.e., \texttt{function}) and \athanor is able to outperform \athanorreduced on this problem, showcasing the effectiveness of the derived high-level neighbourhoods. \athanor's performance is closely followed by the LNS solvers, which are able to edit both functions together. The CBLS solvers are relatively weak, possibly because they cannot use neighbourhoods designed specifically for functions. Both \fznoscar and Yuck fail to find a feasible solution for the vast majority of runs (\Cref{fig:solutions-non-large}) due to either timeout or exceeding the memory limit (\Cref{fig:experiments-runstatus}). SNS was more successful but in some cases the translation process timed out.

The TSP is represented as a fixed-length sequence of integers, and \athanor is able to use neighbourhood structures generated from the direct neighbourhood templates for sequences. For example, the well-known 2-opt move \cite{croes1958method} (where a contiguous subsequence is reversed) is used by \athanor and SNS, but (to our knowledge) is not available in the other solvers. In particular, OscaR-CBLS (the CBLS library used by \fznoscar) has specialised sequence neighbourhoods including 2-opt \cite{oscar-cbls-sequence-neighbourhoods}, but (to the best of our knowledge) these are not available in \fznoscar because the sequence is represented as an array of integers in MiniZinc. 
The effective neighbourhood derivation results in \athanor's winning performance compared with all other local search solvers, including \athanorreduced.
Yuck and \fznoscar have a large number of cases where the memory limit is exceeded. In addition, Yuck has a number of cases where the solver times out, while many SNS runs time out in the translation stage.

The \essence model of Bin Packing is not nested, however it makes use of the partition type which could otherwise be expressed as a set of sets. \athanor will maintain the partition structure at all times, and is able to use neighbourhood structures generated from the direct neighbourhood templates for partition.  As a result, \athanor performs well on Bin Packing, followed by the LNS solvers then \fznoscar (which finds a feasible solution for over 60\% of runs). 
Yuck times out or exceeds the memory limit for almost all runs, while SNS fails by translation timeout for the majority.
Compared with \athanor, \athanorreduced is lacking only the \texttt{SwapParts} neighbourhood template, which results in its reduced performance. Nevertheless, \athanorreduced still performs relatively well compared with other local search solvers, demonstrating the effectiveness of the derived neighbourhood structures for partition.

The SONET problem is represented in \essence using a set of sets of integers. \athanor{}'s higher-order and synchronised neighbourhood templates allow moves such as moving an element from one inner set to another, or swapping elements between two of the inner sets (in addition to moves involving only one of the inner sets). The effectiveness of these high-level neighbourhoods is validated via the significant performance difference between \athanor and \athanorreduced. 
\athanor also significantly outperforms the other local search solvers. 
There are a number of cases where each of the other local search solvers do not find a feasible solution within the time limit. Yuck is also out of memory for a large number of runs, while a small proportion of SNS runs timed out during translation.

CVRP is represented in \essence as a set of sequences of integers (representing locations), allowing \athanor{} to make higher-order moves (by combining higher-order and synchronised neighbourhood templates) such as moving a location from one sequence to another, in addition to moves involving only one sequence (such as reversing a contiguous subsequence). Similar to SONET, 
The performance of \athanor is well ahead of all other solvers, including \athanorreduced, with LNS being the only competitive alternative. 
\Cref{fig:solutions-non-large} shows that \athanor and \athanorreduced are able to find a feasible solution quickly, followed by the two LNS solvers.
SNS and \fznoscar are not able to find feasible solutions due to exceeding solver time or memory limits, or timeout during the translation (SNS).

PPP also has nested structure in the form of a set of partitions. However, there are no synchronised neighbourhood templates for partition so only the direct neighbourhood templates (operating on one partition) are available for this type in \athanor. Nevertheless, \athanor has access to six direct neighbourhood templates for partition, and it convincingly outperforms the other local search solvers. 
Interestingly, \athanorreduced performs slightly (but consistently after 180 seconds) better than \athanor on this problem. \athanorreduced lacks only the \texttt{SwapParts} neighbourhood template. The reason could be that this neighbourhood template is not useful for PPP, and it takes time for the UCB neighbourhood selectors to eliminate it from the search of \athanor.
For the other local search approaches, a majority of runs fail due to either timeout (LNS-EB, LNS-PG and \fznoscar), exceeding the memory limit (Yuck), or timeout during the translation (SNS).

In summary, \athanor is the clear winner for all three classes with a nested type (CVRP, PPP and SONET)
and also for TSP and Bin Packing where powerful direct neighbourhood templates are available. \athanor also performed well on MEB and Knapsack despite a relative lack of structure. 
The results firmly
confirm our hypothesis that \athanor is able to exploit the structure available in high-level problem specifications to generate effective neighbourhoods.

\subsubsection{Systematic Solvers}

\begin{figure}[t]
\begin{center}
\caption{Performance of \athanor compared to systematic solvers using the scoring system described in \Cref{sec:expdetails}. Symmetry is broken for some combinations of problem class and solver (as described in \Cref{sec:symbreak-results}). Higher scores indicate better relative performance.}
\label{fig:experiments-athanor-vs-systematic}
\includegraphics[width=\columnwidth]{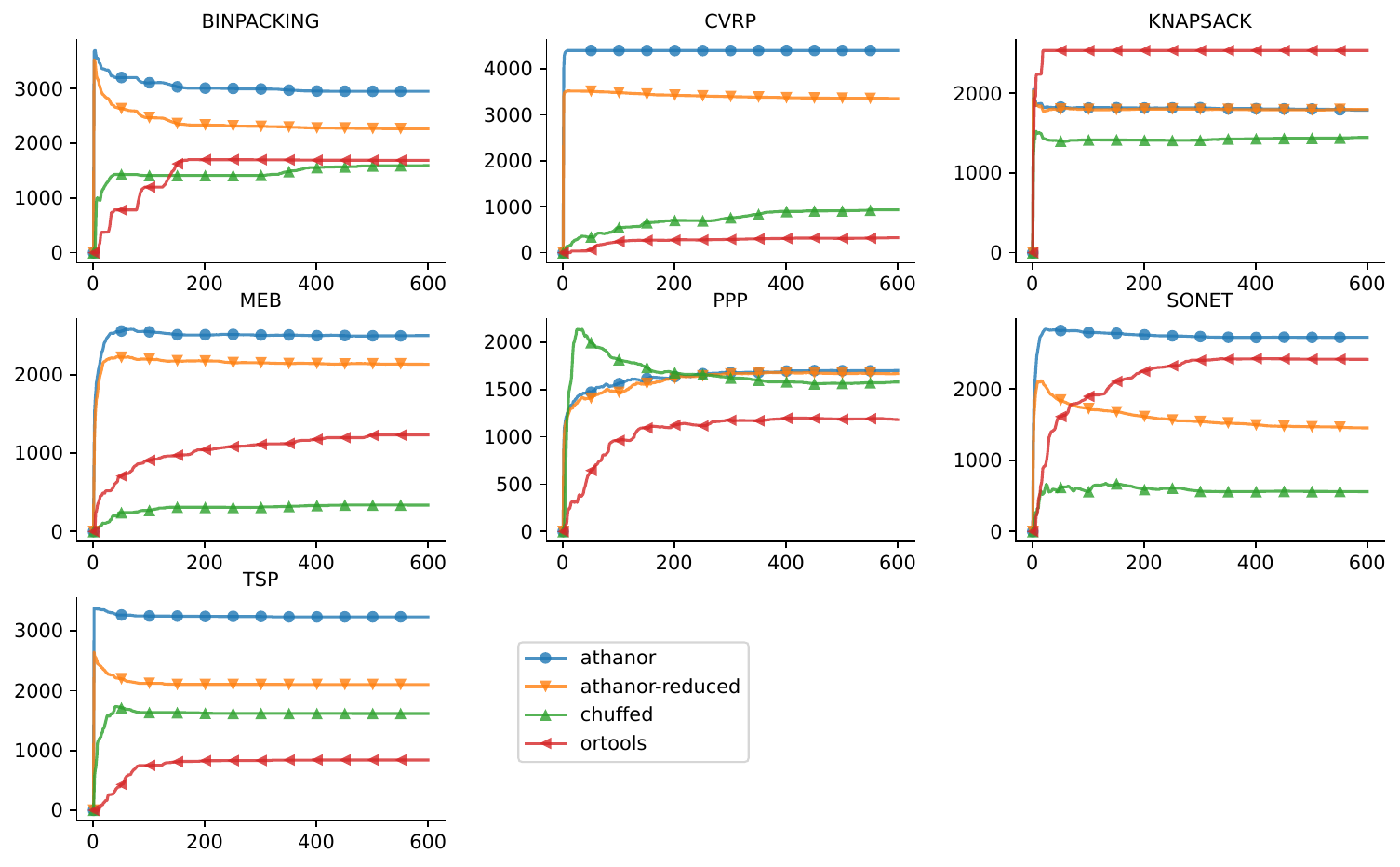}
\end{center}
\end{figure}

In \Cref{fig:experiments-athanor-vs-systematic} we compare \athanor and \athanorreduced with Chuffed and OR-Tools (as described in \Cref{sec:othersolvers}), both of which are systematic solvers with conflict learning and activity-based heuristics. The two systematic solvers have quite different behaviour on the 7 problem classes, with Chuffed performing better on CVRP, PPP and TSP,  while OR-Tools dominates on Knapsack, MEB and SONET. The two solvers are competitive on Bin Packing, where OR-Tools performs slightly better towards the end of the time limit.

Knapsack has a very simple type without any nested structure. Therefore, it is not surprising that \athanor is not competitive on this problem. All solvers are able to find a feasible solution very quickly for all runs (\Cref{fig:solutions-non-large}). They are then competing on solution quality, and OR-Tools quickly takes the lead and retains it.

For MEB and TSP, we found that \athanor performed strongly in comparison to the systematic solvers. 
\athanor is able to quickly find a first feasible solution for all runs on all instances of MEB and TSP, while Chuffed and OR-Tools do not scale as well. For the MEB problem, Chuffed and OR-Tools timed out for a large proportion of runs, while Chuffed ran out of memory for some runs. For TSP, we found that both Chuffed and OR-Tools ran out of memory for a significant proportion of the runs, and also Chuffed timed out for some runs (see \Cref{fig:experiments-runstatus}). Finding a feasible solution for TSP would ordinarily be trivial but the size of the instances (up to 1000 cities) poses a challenge for both systematic solvers.

For Bin Packing, \athanor and \athanorreduced were the only solvers that were able to find feasible solutions for all runs of all instances. However, Chuffed and OR-Tools are able to compete in terms of solution quality (when they find a solution), reducing the scores of \athanor and \athanorreduced over time. OR-Tools is able to find a feasible solution for 50\% of the runs and Chuffed for somewhat less than 50\% (as shown in \Cref{fig:experiments-runstatus}).

On the SONET problem \athanor and \athanorreduced quickly find a feasible solution for all runs on all instances, while the other two solvers are slower to do so. OR-Tools is able to compete with \athanor on solution quality, and \athanor's score declines slightly over time, however after 600 seconds \athanor still has a clear lead in terms of solution quality. As noted above, there is a clear difference between \athanor and \athanorreduced caused by the use of higher-order and synchronised neighbourhood templates in \athanor. 
For both OR-Tools and Chuffed, the sets of integers in the MiniZinc model are represented with Boolean variables, producing a model where most variables and constraints are Boolean and which should be well-suited to conflict learning solvers.

\athanor{} is the clear winner for CVRP, it is able to find feasible solutions quickly for all runs and produce the highest quality solutions at every time step.
OR-Tools and Chuffed do not scale as well.  Both solvers run out of memory for many runs and Chuffed also times out for some runs, as shown in \Cref{fig:experiments-runstatus}.

Finally, PPP challenges all four solvers in different ways.
Chuffed is able to quickly find feasible solutions for all runs, but its score declines over time as the other solvers produce better quality solutions. \athanor{} finds feasible solutions for just over 70\% of runs, and \athanorreduced for just over 80\%. OR-Tools finds feasible solutions on approximately 60\% of runs, timing out for the rest. \athanor tends to find higher quality solutions than Chuffed and \athanorreduced, and as a consequence has a slightly higher score at 600 seconds. This is in contrast to \Cref{fig:experiments-athanor-vs-local}, where \athanorreduced has a better score than \athanor because \athanorreduced is able to find feasible solutions for more runs, and thus gain points from other local search solvers that struggle to find feasible solutions. 
For all instances, a solution exists where all boats are in the \texttt{hosts} set (and every boat crew stays on their own boat throughout the schedule), however this solution is difficult for \athanor{} to find. The reasons for this are explored below where we report experiments with larger instances of PPP.


\subsection{Experiment 2: Scalability} \label{sec:experiments-scalability}

The second experiment focuses on scalability of solvers when given very large instances. The hypothesis here is that \athanor{}'s use of variable-sized data structures for both values and expressions (described in Sections \ref{sec:incremental-hashing} and \ref{subsec:overview} respectively) will allow it to scale gracefully and therefore outperform the other solvers for sufficiently large instances of a given problem class.

We focus on four problem classes in this experiment: Bin Packing, Knapsack, PPP, and SONET. For the Knapsack problem, other solvers outperformed \athanor in the first experiment. For PPP the results were inconclusive. 
For the Bin Packing problem we found that other solvers are able to compete with \athanor in terms of solution quality, despite the \essence specification using a partition domain for the decision variable. The SONET problem is the only other one with a nested type where another solver (OR-Tools) is able to compete with \athanor on solution quality.

\subsubsection{Benchmark Instances}

We generated new large instances for each of the four problem classes as follows.

\noindent\textbf{Bin Packing:} we made use of the Bin Packing generator proposed in~\cite{schwerin1997bin}. The generator (BPPGen) was written in Fortran and can be downloaded from the Bin Packing library BPPLib~\cite{delorme2018bpplib}. The problem size parameter is the number of objects. For each problem size in $\{1000, 1500, 2000\}$ we generated $50$ instances using the default parameters provided in the original generator, i.e., the lower bound and upper bound for the relative weight of items (as a fraction of the bin capacity) are set to $0.2$ and $0.7$, respectively. The bin capacity is set to $10000$.

\noindent\textbf{Knapsack:} we use Pisinger's hard instance generator~\cite{pisinger05} to generate $20$ instances for each problem size (i.e., the number of items) in the set $\{10000,\allowbreak 20000,\allowbreak 30000,\allowbreak 40000,\allowbreak 50000,\allowbreak 60000,\allowbreak 70000,\allowbreak 80000\}$ ($160$ instances in total). The parameters of the generator are set to their default values (range of coefficients: $1000$, instance type: uncorrelated spanner instances of type $11$, number of tests in series: $1000$).

\noindent\textbf{Progressive Party Problem (PPP):} we make use of the new generator described in~\Cref{subsec:ppp_problem} and generate $20$ random instances for each problem size (expressed as \(\langle \mathit{number\ of\ boats}, \mathit{number\ of\ periods} \rangle\)) of $\langle 80, 20\rangle$, $\langle 120, 30\rangle$, $\langle 160, 40\rangle$, and $\langle 200, 50\rangle$ ($80$ instances in total). The capacity is kept in the range $\{10\ldots 100\}$.

\noindent\textbf{SONET:} we make use of the new generator described in~\Cref{subsec:sonet_problem} and  generate $20$ random instances for each of the following problem sizes (written as \(\langle \mathit{number\ of\ nodes},  \mathit{number\ of\ rings}\rangle\)): $\langle 160,80\rangle$, $\langle 180,90\rangle$, and $\langle 200,100\rangle$.

\subsubsection{Results}

\begin{figure}[tbp]
\centering
\begin{subfigure}[b]{0.4\textwidth}
         \centering
         \includegraphics[trim={0 1cm 4.5cm 0.55cm},clip,width=\textwidth]{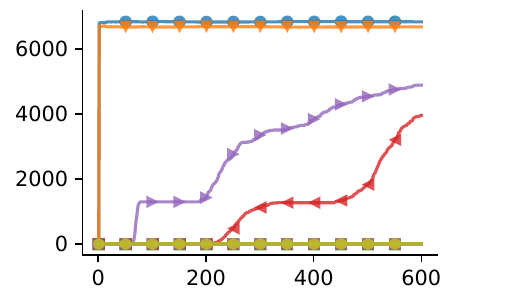}
         \caption{Size 1000}
     \end{subfigure}
     \begin{subfigure}[b]{0.4\textwidth}
         \centering
         \includegraphics[trim={0 1cm 4.5cm 0.55cm},clip,width=\textwidth]{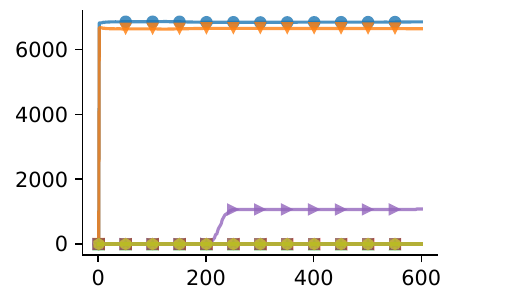}
         \caption{Size 1500}
     \end{subfigure}

\begin{subfigure}[b]{0.4\textwidth}
         \centering
         \includegraphics[trim={0 1cm 4.5cm 0.55cm},clip,width=\textwidth]{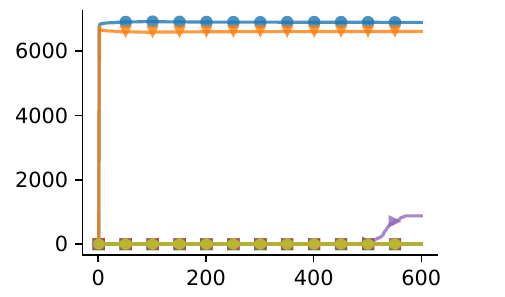}
         \caption{Size 2000}
     \end{subfigure}
     \begin{subfigure}[b]{0.4\textwidth}
         \centering
         \includegraphics[trim={8cm 0 0 0.55cm},clip,width=0.6\textwidth]{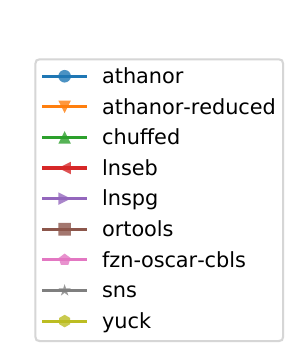}
     \end{subfigure}
\caption{Comparing all solvers on large instances of the Bin Packing problem (with 1000, 1500, or 2000 objects), using the scoring system described in \Cref{sec:expdetails}. Symmetry is broken for some combinations of problem class and solver (as described in \Cref{sec:symbreak-results}). Higher scores indicate better relative performance.}
\label{fig:experiments-large-bp}
\end{figure}

Results for the Bin Packing problem are plotted in \Cref{fig:experiments-large-bp}, showing a clear progression as the instance size increases from 1000 to 2000 objects. \athanor quickly finds feasible solutions for all instances of all three sizes, whereas other solvers either cannot find a feasible solution within the time limit or take significantly longer to do so.  
LNS-PG and LNS-EB are able to find feasible solutions in some cases, and all other solvers are unable to solve any instance. In summary, \athanor clearly scales better than any of the other solvers on this problem class.

\begin{figure}[tbp]
\centering
    \begin{subfigure}[b]{0.49\textwidth}
         \centering
         \caption{Bin Packing}
         \includegraphics[trim={0 1.8cm 0 0.8cm},clip,height=3.2cm]{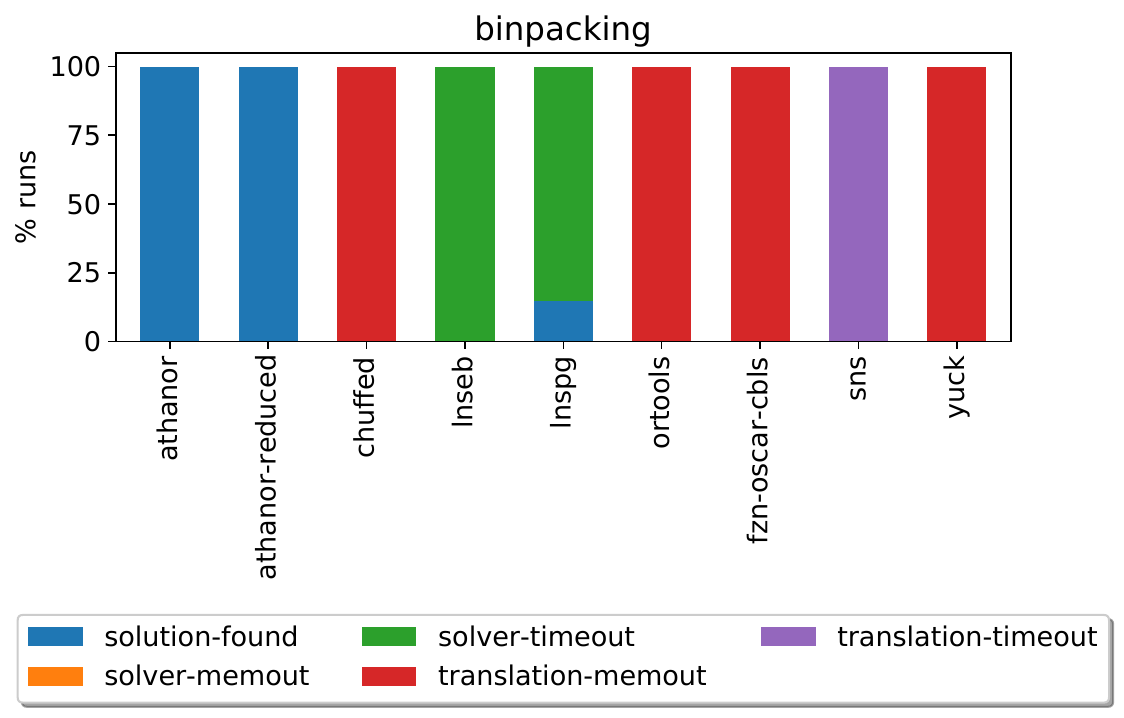}

     \end{subfigure}
     \begin{subfigure}[b]{0.49\textwidth}
         \centering
         \caption{Knapsack}
         \includegraphics[trim={0 1.8cm 0 0.8cm},clip,height=3.2cm]{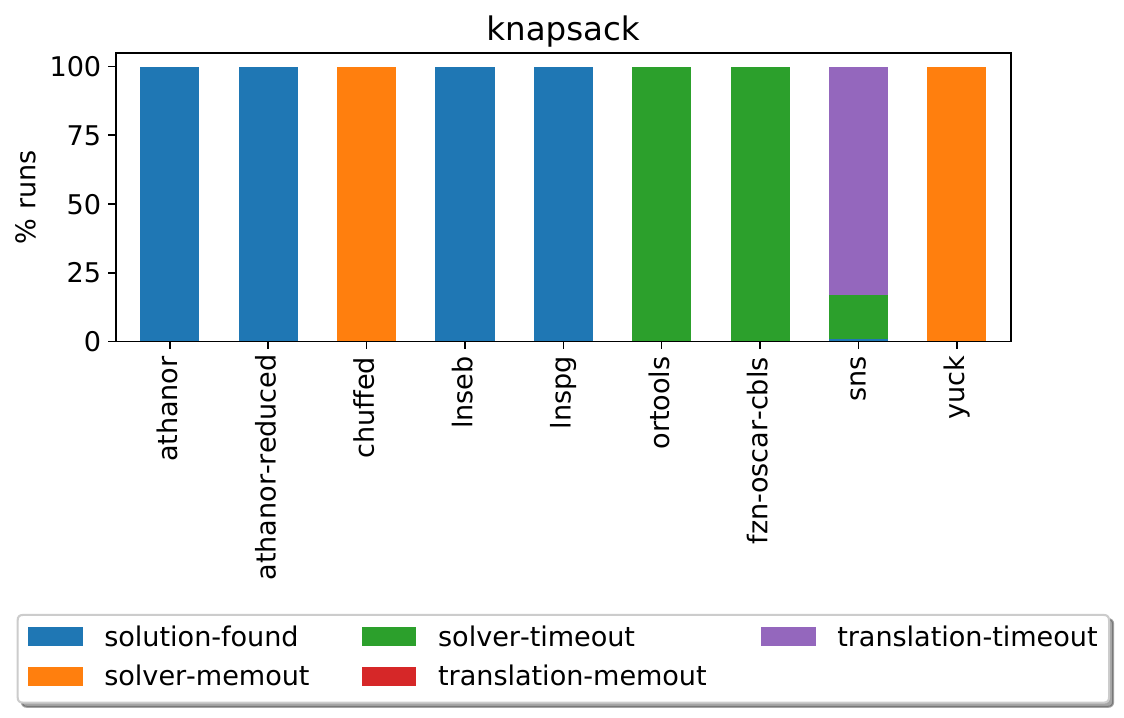}

     \end{subfigure}

     \begin{subfigure}[b]{0.49\textwidth}
         \centering
         \caption{PPP}
         \includegraphics[trim={0 1.8cm 0 0.8cm},clip,height=3.2cm]{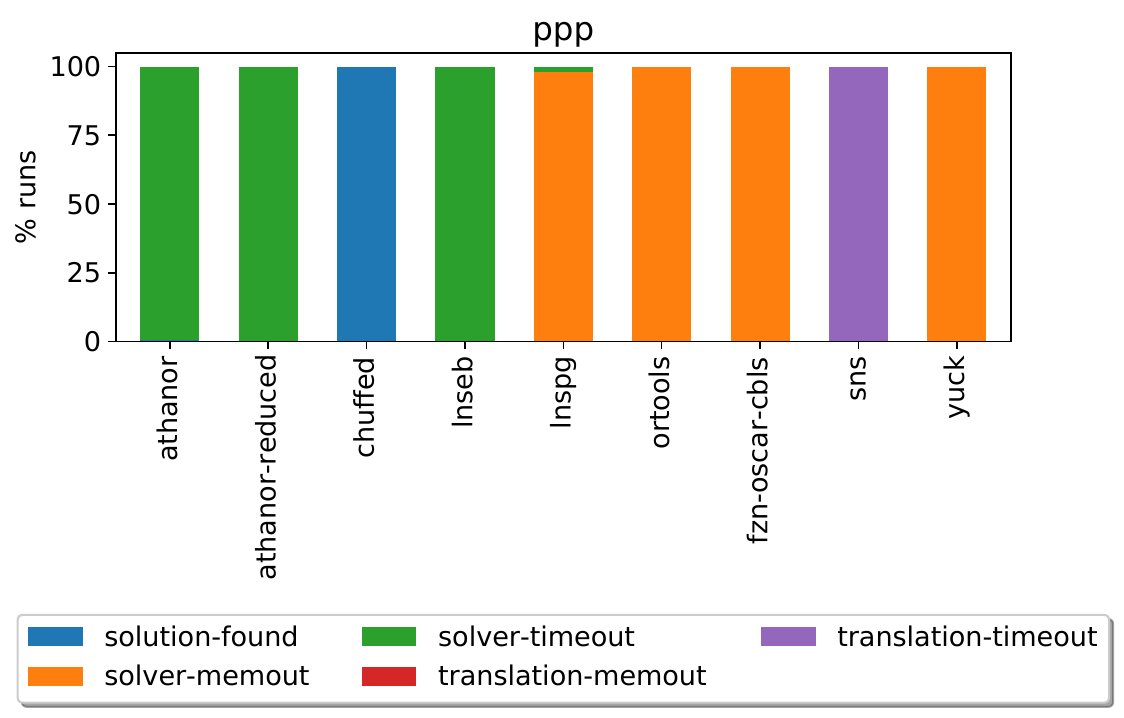}

     \end{subfigure}
     \begin{subfigure}[b]{0.49\textwidth}
         \centering
         \caption{SONET}
         \includegraphics[trim={0 1.8cm 0 0.8cm},clip,height=3.2cm]{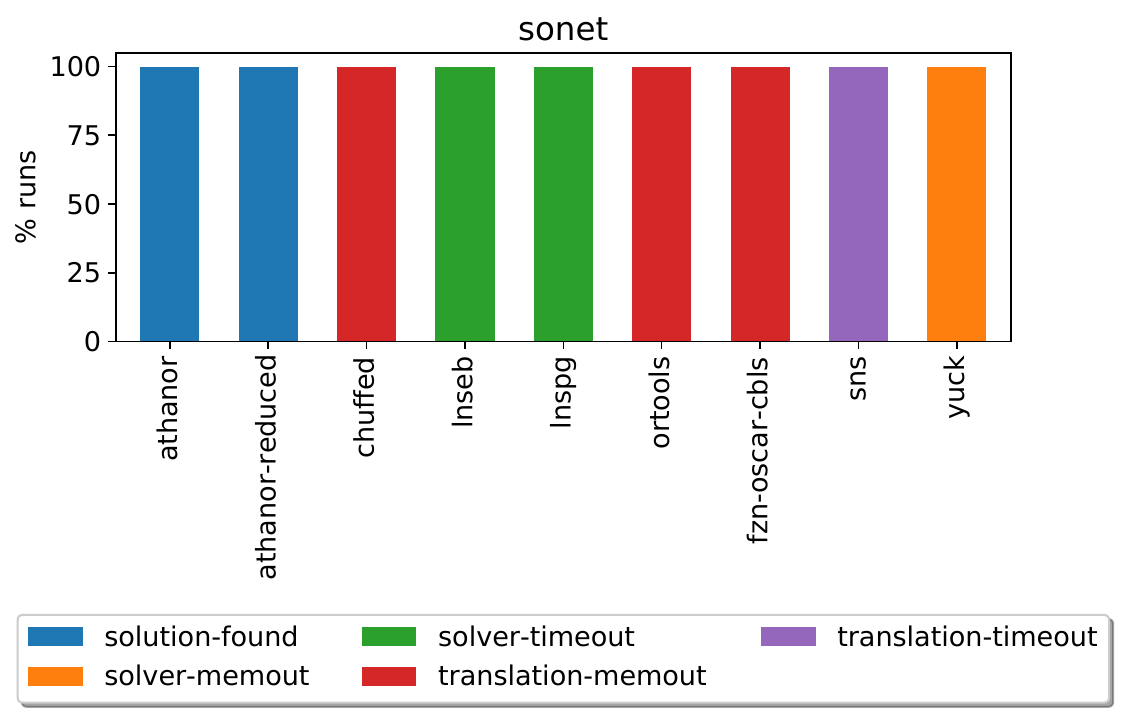}

     \end{subfigure}

\centering
\includegraphics[trim={0 0 0 10cm},clip,width=0.7\textwidth]{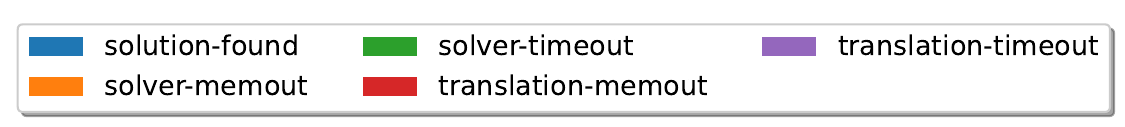}
\caption{Status of solver runs for Bin Packing (size 2000), Knapsack (size 80,000), PPP (size 160), and SONET (size 200), for each solver in the scalability experiment.
Note that \texttt{solution-found} indicates that a solver finds a feasible solution within the time limit and that it does not exceed the memory limit during the run.}
\label{fig:experiments-scal-runstatus}
\end{figure}

\Cref{fig:experiments-scal-runstatus} summarises the reasons for failure of each solver on the largest Bin Packing instances. All the MiniZinc solvers and SNS ran out of memory or time during translation, while the LNS solvers were able to start searching but in most cases they timed out before finding a feasible solution. Only LNS-PG finds any feasible solutions for the size 2000 instances.

\begin{figure}[tbp]
\centering
    \begin{subfigure}[b]{0.3\textwidth}
         \centering
         \includegraphics[trim={0 1.2cm 4.5cm 0.55cm},clip,width=\textwidth]{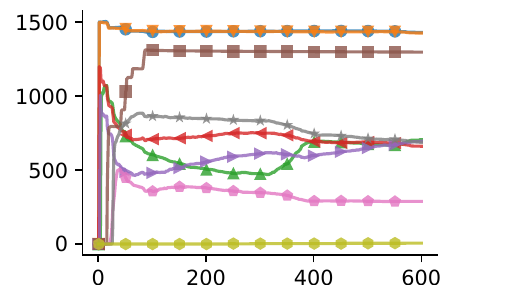}
         \caption{Size 10,000}
     \end{subfigure}
     \begin{subfigure}[b]{0.3\textwidth}
         \centering
         \includegraphics[trim={0 1.2cm 4.5cm 0.55cm},clip,width=\textwidth]{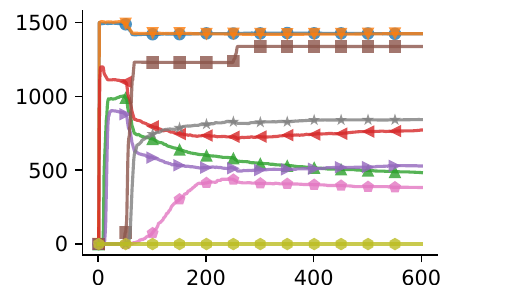}
         \caption{Size 20,000}
     \end{subfigure}
     \begin{subfigure}[b]{0.3\textwidth}
         \centering
         \includegraphics[trim={0 1.2cm 4.5cm 0.55cm},clip,width=\textwidth]{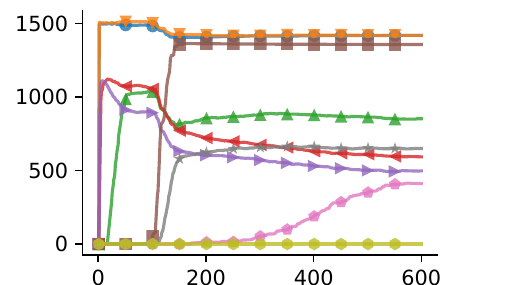}
         \caption{Size 30,000}
     \end{subfigure}

     \begin{subfigure}[b]{0.3\textwidth}
         \centering
         \includegraphics[trim={0 1.2cm 4.5cm 0.55cm},clip,width=\textwidth]{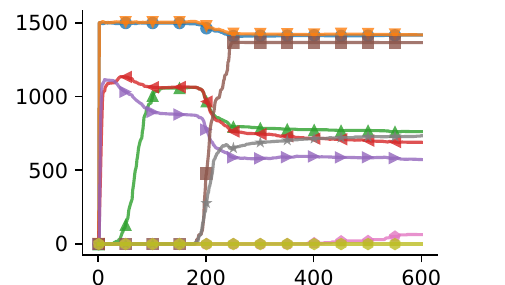}
         \caption{Size 40,000}
     \end{subfigure}
     \begin{subfigure}[b]{0.3\textwidth}
         \centering
         \includegraphics[trim={0 1.2cm 4.5cm 0.55cm},clip,width=\textwidth]{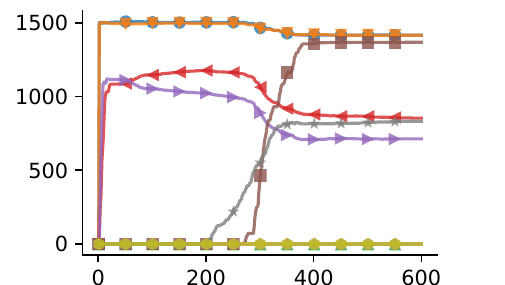}
         \caption{Size 50,000}
     \end{subfigure}
     \begin{subfigure}[b]{0.3\textwidth}
         \centering
         \includegraphics[trim={0 1.2cm 4.5cm 0.55cm},clip,width=\textwidth]{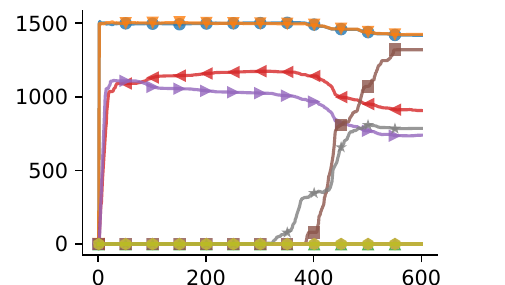}
         \caption{Size 60,000}
     \end{subfigure}

\begin{subfigure}[b]{0.3\textwidth}
         \centering
         \includegraphics[trim={0 1.2cm 4.5cm 0.55cm},clip,width=\textwidth]{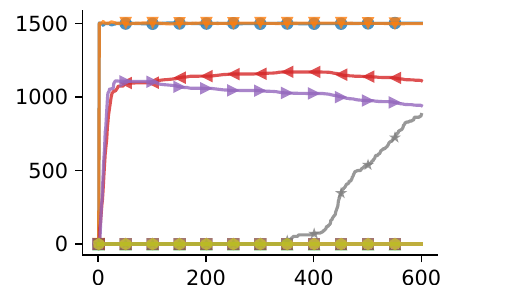}
         \caption{Size 70,000}
     \end{subfigure}
\begin{subfigure}[b]{0.3\textwidth}
         \centering
         \includegraphics[trim={0 1.2cm 4.5cm 0.55cm},clip,width=\textwidth]{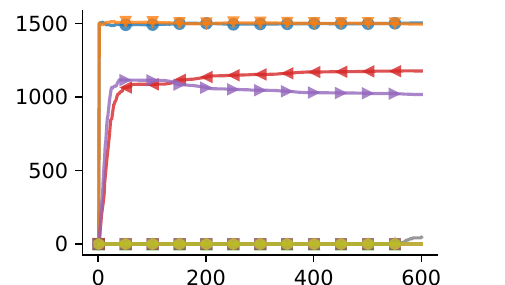}
         \caption{Size 80,000}
     \end{subfigure}
     \begin{subfigure}[b]{0.3\textwidth}
         \centering
         \includegraphics[trim={8.5cm 0 0 0.55cm},clip,width=0.55\textwidth]{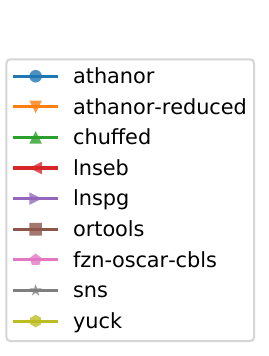}
     \end{subfigure}
\caption{Comparing all solvers  on large instances of the Knapsack problem (with 10,000 to 80,000 objects), using the scoring system described in \Cref{sec:expdetails}. Higher scores indicate better relative performance.}
\label{fig:experiments-large-knap}
\end{figure}

For the Knapsack problem, we found that \athanor is able to find feasible solutions quickly and outperform the other solvers for all 8 sizes, as shown in \Cref{fig:experiments-large-knap}. OR-Tools performs well up to size 60,000, but does not scale to the larger sizes.
For the size 80,000 instances, \athanor, LNS-EB, and LNS-PG are able to find an initial feasible solution for all runs of all instances.  \athanor has a higher score based on solution quality. 
The Knapsack problem contains a single decision variable with type \emt|set of enum|, and so can benefit from \athanor's  compact representation of sets, as well as dynamic unrolling of expressions for the objective and constraint (all of which scale with the actual size of the set, rather than its maximum possible size).
\Cref{fig:experiments-scal-runstatus} shows that Chuffed and Yuck are limited by their memory use, while OR-Tools and \fznoscar time out before finding a feasible solution. SNS times out during translation for the majority of the instances of size 80,000.

\begin{figure}[tbp]
\centering
    \begin{subfigure}[b]{0.3\textwidth}
         \centering
         \includegraphics[trim={0 1.2cm 4.5cm 0.55cm},clip,width=\textwidth]{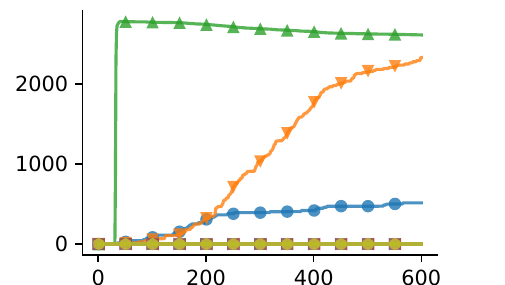}
         \caption{Size 80}
     \end{subfigure}
     \begin{subfigure}[b]{0.3\textwidth}
         \centering
         \includegraphics[trim={0 1.2cm 4.5cm 0.55cm},clip,width=\textwidth]{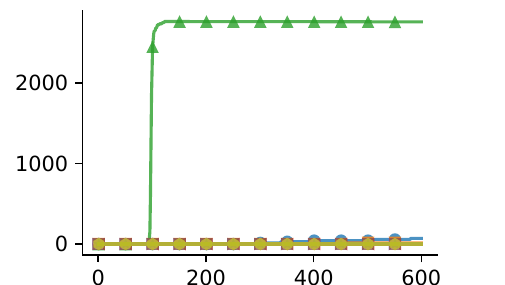}
         \caption{Size 120}
     \end{subfigure}
     \begin{subfigure}[b]{0.3\textwidth}
         \centering
         \includegraphics[trim={0 1.2cm 4.5cm 0.55cm},clip,width=\textwidth]{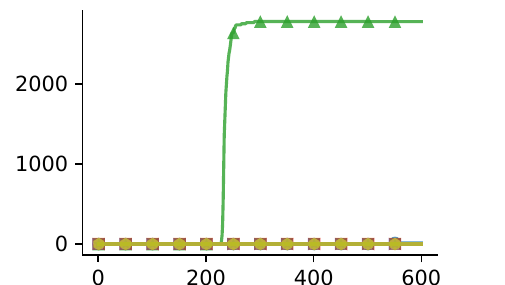}
         \caption{Size 160}
     \end{subfigure}

\begin{subfigure}[b]{0.3\textwidth}
         \centering
         \includegraphics[trim={0 1.2cm 4.5cm 0.55cm},clip,width=\textwidth]{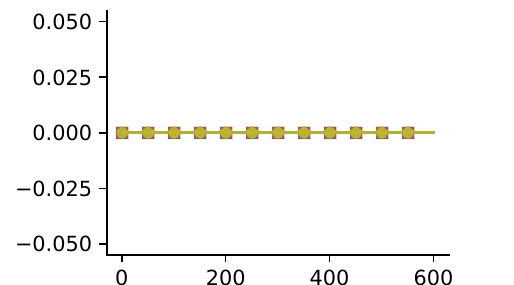}
         \caption{Size 200}
     \end{subfigure}
     \begin{subfigure}[b]{0.3\textwidth}
         \centering
         \includegraphics[trim={8.5cm 0 0 0.55cm},clip,width=0.55\textwidth]{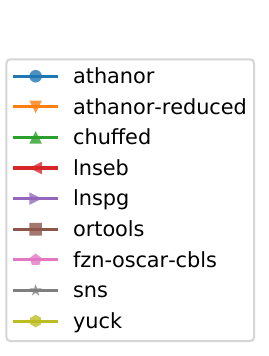}
     \end{subfigure}
\caption{Comparing all solvers  on large instances of PPP (with 80 to 200 boats), using the scoring system described in \Cref{sec:expdetails}. Symmetry is broken for some combinations of problem class and solver (as described in \Cref{sec:symbreak-results}). Higher scores indicate better relative performance.}
\label{fig:experiments-large-ppp}
\end{figure}

\begin{figure}[tbp]
\centering
\begin{subfigure}[b]{0.4\textwidth}
         \centering
         \includegraphics[trim={0 1.2cm 4.5cm 0.55cm},clip,width=\textwidth]{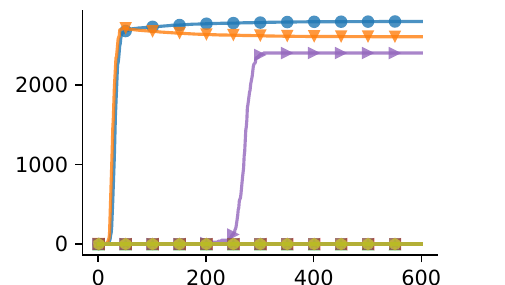}
         \caption{Size 160}
     \end{subfigure}
     \begin{subfigure}[b]{0.4\textwidth}
         \centering
         \includegraphics[trim={0 1.2cm 4.5cm 0.55cm},clip,width=\textwidth]{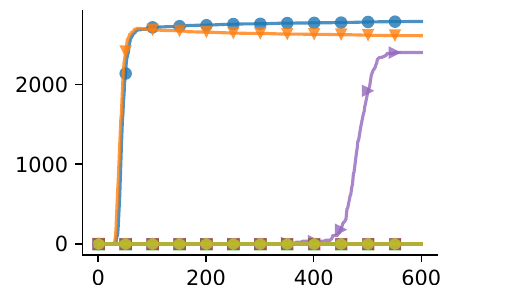}
         \caption{Size 180}
     \end{subfigure}

\begin{subfigure}[b]{0.4\textwidth}
         \centering
         \includegraphics[trim={0 1.2cm 4.5cm 0.55cm},clip,width=\textwidth]{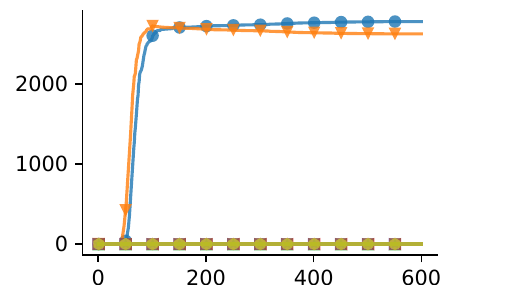}
         \caption{Size 200}
     \end{subfigure}
     \begin{subfigure}[b]{0.4\textwidth}
         \centering
         \includegraphics[trim={8.5cm 0 0 0.55cm},clip,width=0.55\textwidth]{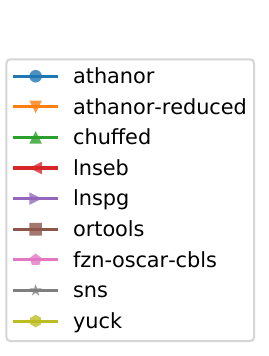}
     \end{subfigure}
\caption{Comparing all solvers  on large instances of the SONET problem (with 160 to 200 nodes), using the scoring system described in \Cref{sec:expdetails}. Symmetry is broken for some combinations of problem class and solver (as described in \Cref{sec:symbreak-results}). Higher scores indicate better relative performance.}
\label{fig:experiments-large-sonet}
\end{figure}

The results for large PPP instances are shown in \Cref{fig:experiments-large-ppp}. Only Chuffed, \athanor, and \athanorreduced were able to find feasible solutions for any of the four sizes. On the size 160 instances, the solvers other than Chuffed failed in several different ways (as shown in \Cref{fig:experiments-scal-runstatus}). All solvers fail on instance size 200, indicating the scalability challenge of this problem. As discussed in \Cref{sec:experiments-all-solvers}, \athanor is hindered by the partition \texttt{SwapParts} neighbourhood structure and \athanorreduced is more effective.
Most constraints of the PPP are easy to solve if we have a large set of hosts (for example, adding a host increases the total host boat capacity so capacity constraints become easier to solve). 
However, the solutions with large host sets are not easily reachable for \athanor{}. The second constraint in particular (\Cref{fig:ppp_spec} lines 11-12) is quantified with \texttt{forAll h in hosts}, therefore its violation may be reduced by taking an element out of \texttt{hosts}. As a consequence, the second constraint will discourage \athanor from exploring areas with large sets of hosts. 
By examining the UCB neighbourhood selector's statistics across PPP instances we found that in general
the most frequently used neighbourhood templates all make changes to partitions within the schedule (using the \texttt{LiftSingle} higher-order neighbourhood template to lift a partition from \texttt{sched}), while changes to the set of hosts are less frequent. Removing an element from \texttt{hosts} is more frequent than adding an element to it, and other neighbourhood structures do not change the size of \texttt{hosts}.  Also, if the final constraint (\Cref{fig:ppp_spec} line 14, pairs of crews meet at most once) is removed, we found that  \athanor finds a feasible solution for 53 of 200 runs on the 20 instances of size 200. Removing a very significant constraint on the schedule does make the problem easier but not to the extent that \athanor can straightforwardly solve the largest instances.
Taken together, these observations suggest that \athanor is changing the set of hosts relatively infrequently and spending most of its time attempting to satisfy constraints on the schedule for a fixed set of hosts.

Diversifying the search to include large sets of hosts could potentially help a future version of \athanor to find feasible solutions to PPP. It is worth noting that each neighbourhood structure is treated as an independent action by UCB, and there is no mechanism for UCB to learn useful sequences of actions. To reduce the violation of the capacity constraints, \athanor would need to first add a new host and then redistribute the guest crews among the hosts. However, \athanor does not have any incentive to apply the first action alone.
Adding a host does not improve the violation of the capacity constraints, might break the first constraint (that \texttt{|parts(p)| = |hosts|} for all partitions in the schedule), and increases the objective value.

For the SONET problem (\Cref{fig:experiments-large-sonet}) we found that \athanor can quickly produce high-quality solutions for all three sizes of instances (including all runs of all instances of size 200). LNS-PG is also able to find feasible solutions up to size 180 but it is not competitive with \athanor with respect to solution quality. Other solvers are failing for a variety of reasons, as shown in \Cref{fig:experiments-scal-runstatus}.

In summary, our hypothesis that \athanor will scale better than the other solvers is largely supported by the results. The exception is the Progressive Party Problem where \athanor was unable to find feasible solutions for large instances. For sufficiently large instances of Bin Packing, Knapsack, and SONET, \athanor performs substantially better than the other solvers.


\section{Experiments with Specialised Global Constraints}\label{sec:experiment-globcons}

Some of our benchmark problems can be modelled using global constraints that represent the entire problem or a substantial part of it (such as the knapsack constraint). We refer to these global constraints as \textit{specialised} to distinguish them from others (such as allDifferent and element) that are widely used across otherwise entirely distinct problem classes. Implementations of specialised global constraints can include sophisticated reasoning that is specific to a problem class (for example, Shaw's propagator for the bin packing constraint \cite{shaw2004constraint}), therefore we consider them to be similar to problem-class-specific solvers and excluded them from the experiments reported in \Cref{sec:experiments}. In this section we compare \athanor to seven other solvers (as in \Cref{sec:experiments}) on four problem classes with specialised global constraints. The MiniZinc and Choco models for the four problems are briefly described here, and are available in the experimental repository. 

\begin{itemize}
\item The Knapsack problem is entirely captured with a knapsack global constraint \cite{katriel2007propagating} stated on 0/1 variables. 
\item Bin Packing is modelled with a bin packing global constraint that includes load variables for each bin \cite{shaw2004constraint}. The load variables are used in the objective (to minimise the number of non-empty bins). 
\item TSP is modelled with a global circuit constraint \cite{caseau1997circuit} stated on a successor representation of the sequence. The objective is modelled using element constraints to look up the distances between adjacent locations. 
\item CVRP is converted to a single sequence of locations using multiple dummy locations to mark the start and end of the delivery routes.\footnote{The model used here is very closely based on the MiniZinc benchmark: \url{https://github.com/MiniZinc/minizinc-benchmarks/}} The successor representation is used with a circuit constraint, combined with a redundant predecessor representation with a second circuit constraint. Other decision variables represent load and arrival time for each location (accumulated as a vehicle traverses a delivery route), enabling the capacity constraint and the objective (minimise total travel time) to be stated straightforwardly. The MiniZinc model includes a custom search order that branches on the successor variables first. 
\end{itemize}

We use the same benchmark instances as in \Cref{sec:experiments}. For Bin Packing, some larger instances have been generated for the scalability experiment. 
\Cref{tab:native-global-cts} shows which of the solvers implement the specialised global constraints. \athanor and SNS use the same \essence specification as in \Cref{sec:experiments}. 

\begin{table}
\centering
\begin{tabular}{lcccccc}
\toprule
Problem     & Chuffed & OR-Tools  & \fznoscar & Yuck  & Choco LNS \\ 
\midrule
Knapsack    &         &           &            &       & \tick     \\
Bin Packing &         &           &            & \tick & \tick     \\
TSP, CVRP   &         &  \tick    &  \tick     & \tick & \tick     \\
\bottomrule
\end{tabular}
\caption{Implementation of specialised global constraints by each solver. A tick indicates that the solver implements the constraint natively (and, for MiniZinc solvers, that the native implementation is used by MiniZinc). Otherwise the global constraint is decomposed before reaching the solver.}\label{tab:native-global-cts}
\end{table}

Among the four problems, Bin Packing is the only one where symmetry breaking is available (with both MiniZinc and Choco models). We evaluated the two Choco LNS solvers, Yuck, OR-Tools, Chuffed, and \fznoscar both with and without symmetry breaking. Experimental results indicate that the version without symmetry breaking outperforms the one with symmetry breaking for all solvers on this problem except Chuffed. In all plots shown in this section, we present results for Chuffed with symmetry-breaking constraints and the rest without those constraints. 

\subsection{Experiment 1: Evaluation of Athanor Neighbourhoods}

As in \Cref{sec:experiments-all-solvers}, the first experiment compares \athanor with the other seven solver configurations using all four problem classes for which we have a specialised global constraint model. We are testing the first hypothesis: that \athanor will generate effective neighbourhood structures from the high-level structure available in the \essence specifications. If the hypothesis is true, we expect \athanor to be competitive with the specialised global constraint models. 

\begin{figure}[h]
\begin{center}
\caption{Performance of \athanor compared to other local search solvers \textbf{with specialised global constraints} using the scoring system described in \Cref{sec:expdetails}. Higher scores indicate better relative performance.}
\label{fig:experiments-athanor-vs-local-gc}
\includegraphics[width=\columnwidth]{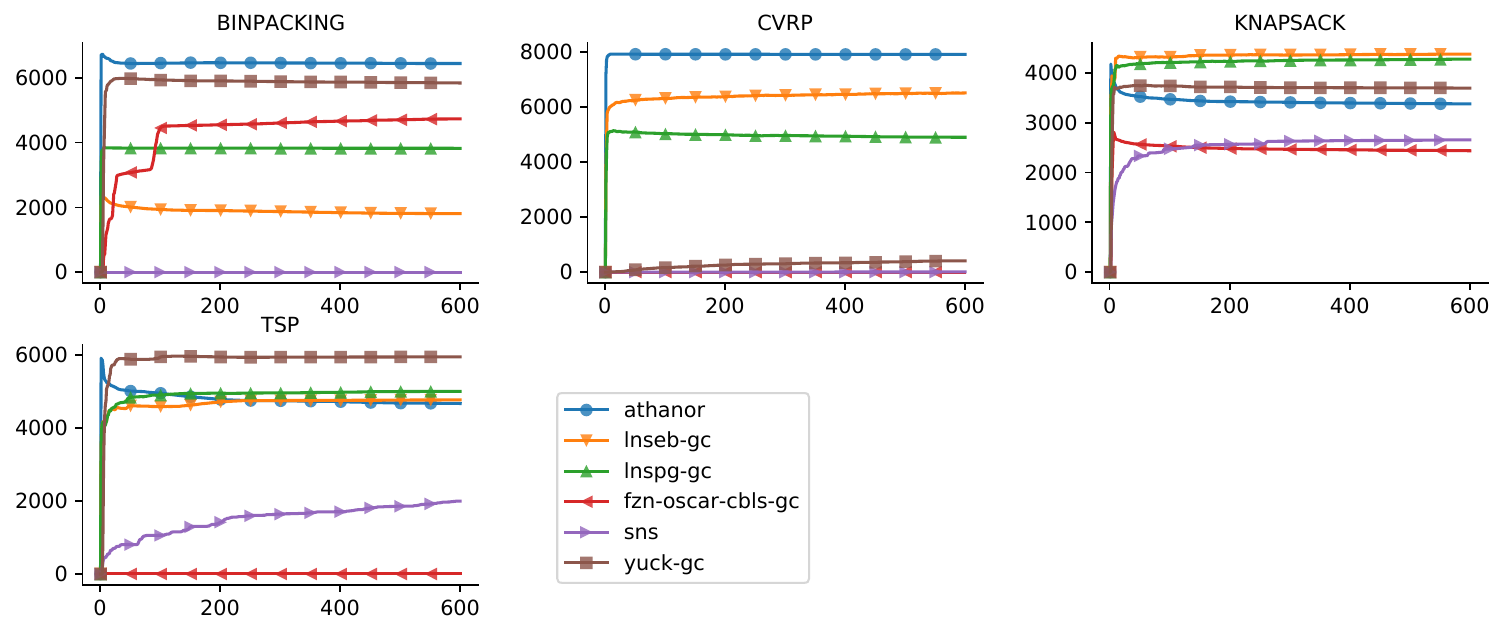}
\end{center}
\end{figure}

\athanor is compared with other local search solvers in \Cref{fig:experiments-athanor-vs-local-gc}. For the Bin Packing problem, \athanor remains the leading solver but by a smaller margin than in \Cref{sec:experiments}, and the relative performance of Yuck in particular is much improved by using the specialised global constraint. For CVRP, once again it is the LNS solvers that are the closest challengers. \athanor remains the best-performing solver but by a smaller margin. On the Knapsack problem, the two LNS solvers and Yuck perform better than \athanor, with all three showing a big improvement when using the specialised global constraint model.\footnote{The knapsack global constraint is decomposed for Yuck. The MiniZinc model of Knapsack used in \Cref{sec:experiments} has a set variable, while in this section we use an array of 0/1 variables as required for the global constraint. Yuck exhibits better performance on the 0/1 model.} Finally, for TSP the LNS solvers and Yuck are substantially improved by using the specialised global constraint model, and \athanor is not the best-performing solver for this problem class. 

\begin{figure}[h]
\begin{center}
\caption{Performance of \athanor compared to systematic solvers \textbf{with specialised global constraints} using the scoring system described in \Cref{sec:expdetails}.  Higher scores indicate better relative performance.}
\label{fig:experiments-athanor-vs-systematic-gc}
\includegraphics[width=\columnwidth]{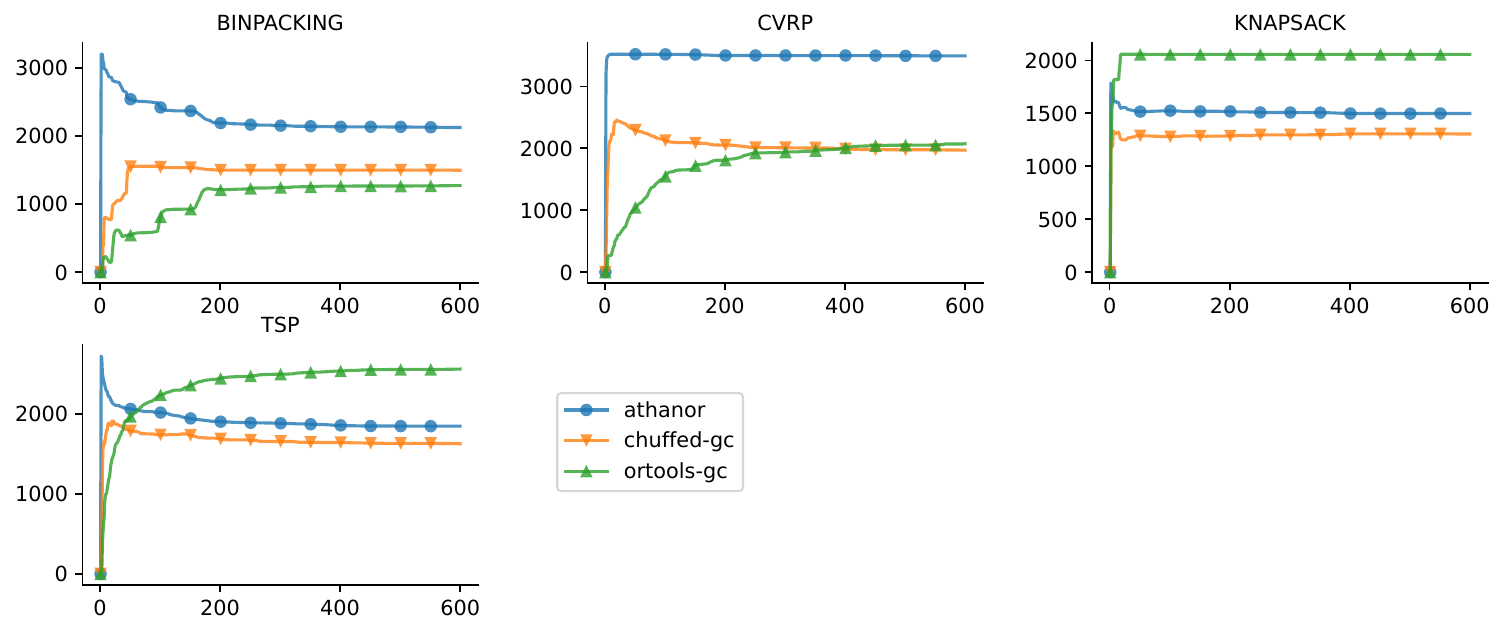}
\end{center}
\end{figure}

\athanor is compared to the systematic solvers in \Cref{fig:experiments-athanor-vs-systematic-gc}. For Bin Packing and CVRP, the results are qualitatively similar to those in \Cref{sec:experiments} but \athanor is leading by a smaller margin. On the Knapsack problem, the results are very similar to those without the specialised global constraint. Finally, for TSP the performance of OR-Tools is far better with the specialised global constraint, and in this experiment it outperforms \athanor. 

\subsection{Experiment 2: Scalability}

As in \Cref{sec:experiments-scalability}, the second experiment focuses on scalability of solvers when given very large instances. The hypothesis is that \athanor{}'s use of variable-sized data structures for both values and expressions will allow it to scale gracefully and therefore outperform the other solvers for sufficiently large instances of a given problem class. For Bin Packing we found that other solvers were able to compete with \athanor and reduce its score over time in the first experiment (although \athanor still has the highest score at 600 seconds). For the Knapsack problem, several other solvers were able to outperform \athanor on instances with 20 to 10,000 objects (Figures \ref{fig:experiments-athanor-vs-local-gc} and \ref{fig:experiments-athanor-vs-systematic-gc}). We perform scalability experiments for Bin Packing and Knapsack, and we also discuss TSP below. 

\begin{figure}[!h]
\centering
\begin{subfigure}[b]{0.4\textwidth}
         \centering
         \includegraphics[trim={0 0.5cm 4cm 0.55cm},clip,width=\textwidth]{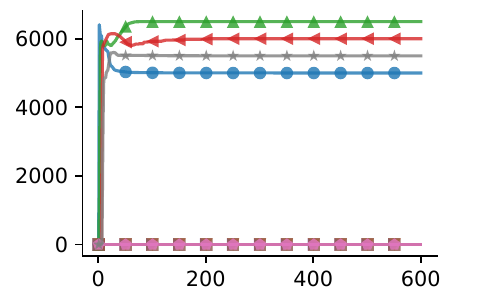}
         \caption{Size 1000}
     \end{subfigure}
 \begin{subfigure}[b]{0.4\textwidth}
     \centering
     \includegraphics[trim={0 0.5cm 4cm 0.55cm},clip,width=\textwidth]{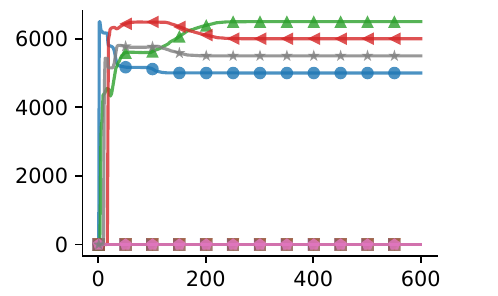}
     \caption{Size 1500}
 \end{subfigure}
\begin{subfigure}[b]{0.4\textwidth}
         \centering
         \includegraphics[trim={0 0.5cm 4cm 0.55cm},clip,width=\textwidth]{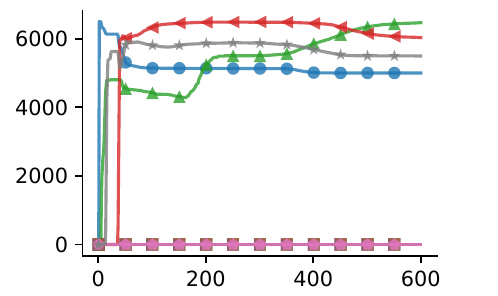}
         \caption{Size 2000}
     \end{subfigure}
\begin{subfigure}[b]{0.4\textwidth}
         \centering
         \includegraphics[trim={0 0.5cm 4cm 0.55cm},clip,width=\textwidth]{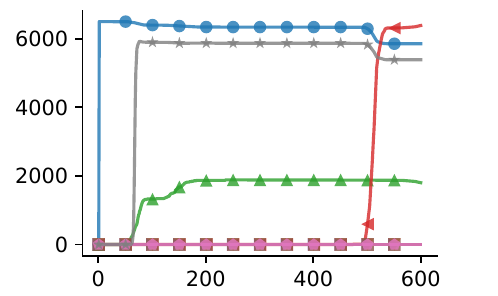}
         \caption{Size 5000}
     \end{subfigure}
\begin{subfigure}[b]{0.4\textwidth}
         \centering
         \includegraphics[trim={0 0.5cm 4cm 0.55cm},clip,width=\textwidth]{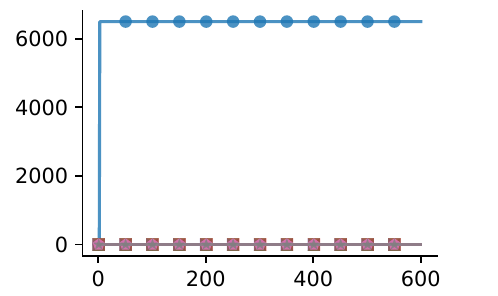}
         \caption{Size 10,000}
     \end{subfigure}
\begin{subfigure}[b]{0.4\textwidth}
         \centering
         \includegraphics[trim={0 0.5cm 4cm 0.55cm},clip,width=\textwidth]{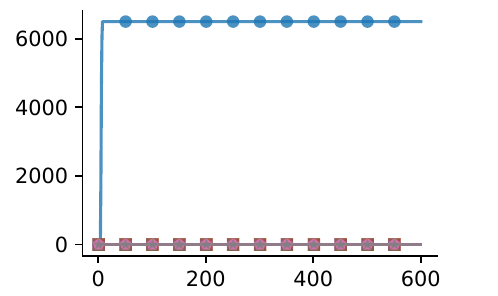}
         \caption{Size 20,000}
     \end{subfigure}

 \begin{subfigure}[b]{0.4\textwidth}
     \centering
     \includegraphics[trim={8cm 0 0 0.55cm},clip,width=0.6\textwidth]{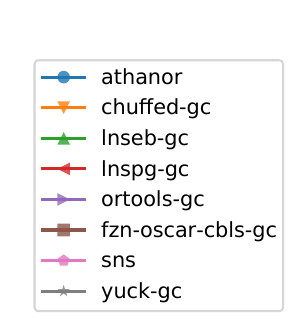}
 \end{subfigure}
\caption{Comparing all solvers \textbf{with the bin packing global constraint model} (except \athanor and SNS) on large instances of the Bin Packing problem (with 1000, 1500, 2000, 5000, 10,000, or 20,000 objects), using the scoring system described in \Cref{sec:expdetails}. Higher scores indicate better relative performance.}
\label{fig:experiments-large-bp-gc}
\end{figure}

\Cref{fig:experiments-large-bp-gc} shows the relative performance of the solvers on the Bin Packing problem as instance size is scaled up to 20,000 objects. We use the same instances as in \Cref{sec:experiments-scalability} up to size 2000, and the larger ones were generated with the same generator and parameters. 
The LNS solvers and Yuck scale better when using the specialised global constraint, but ultimately the outcome is the same as in \Cref{sec:experiments}: \athanor scales to larger instances than any of the other solvers. 

The results on the smallest instances in this experiment are quite different from those presented in \Cref{fig:experiments-athanor-vs-local-gc} and \Cref{fig:experiments-athanor-vs-systematic-gc}. This is caused by differences in the instance distribution (other than their size). Here the instances of size 1000 are drawn from one distribution (i.e.\ generated with one generator using one set of parameters). The normal size instances of Bin Packing are drawn from multiple distributions and have 60 to 1000 objects.

\begin{figure}[!h]
\centering
    \begin{subfigure}[b]{0.3\textwidth}
         \centering
         \includegraphics[trim={0 0.5cm 4cm 0.55cm},clip,width=\textwidth]{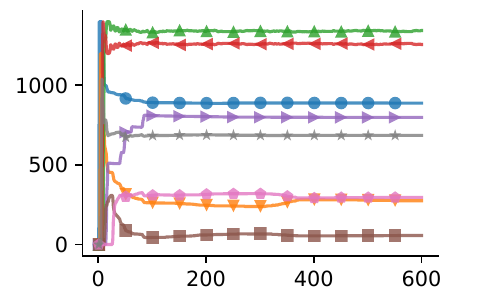}
         \caption{Size 10,000}
     \end{subfigure}
     \begin{subfigure}[b]{0.3\textwidth}
         \centering
         \includegraphics[trim={0 0.5cm 4cm 0.55cm},clip,width=\textwidth]{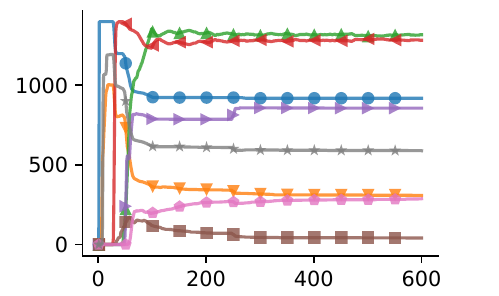}
         \caption{Size 20,000}
     \end{subfigure}
     \begin{subfigure}[b]{0.3\textwidth}
         \centering
         \includegraphics[trim={0 0.5cm 4cm 0.55cm},clip,width=\textwidth]{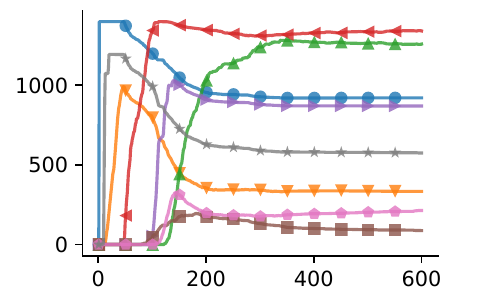}
         \caption{Size 30,000}
     \end{subfigure}

     \begin{subfigure}[b]{0.3\textwidth}
         \centering
         \includegraphics[trim={0 0.5cm 4cm 0.55cm},clip,width=\textwidth]{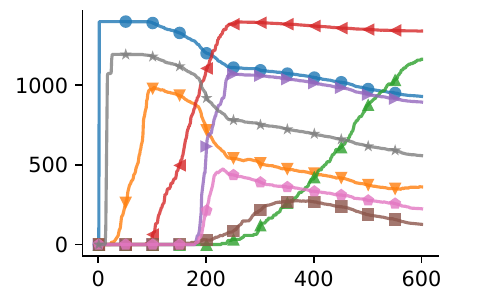}
         \caption{Size 40,000}
     \end{subfigure}
     \begin{subfigure}[b]{0.3\textwidth}
         \centering
         \includegraphics[trim={0 0.5cm 4cm 0.55cm},clip,width=\textwidth]{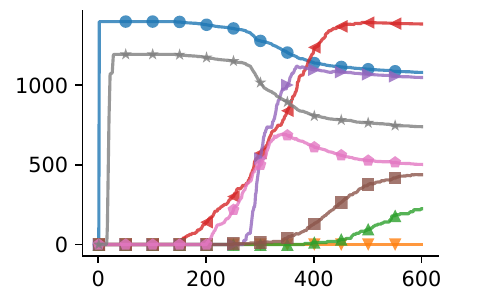}
         \caption{Size 50,000}
     \end{subfigure}
     \begin{subfigure}[b]{0.3\textwidth}
         \centering
         \includegraphics[trim={0 0.5cm 4cm 0.55cm},clip,width=\textwidth]{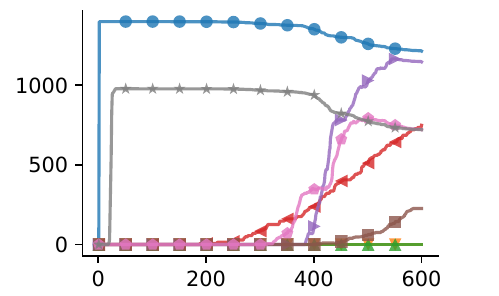}
         \caption{Size 60,000}
     \end{subfigure}

\begin{subfigure}[b]{0.3\textwidth}
         \centering
         \includegraphics[trim={0 0.5cm 4cm 0.55cm},clip,width=\textwidth]{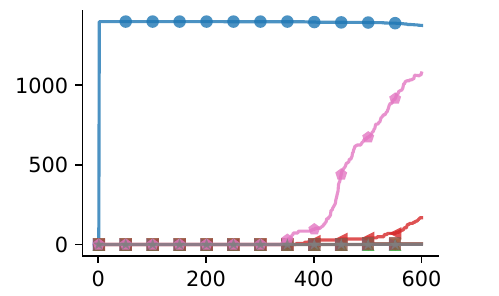}
         \caption{Size 70,000}
     \end{subfigure}
\begin{subfigure}[b]{0.3\textwidth}
         \centering
         \includegraphics[trim={0 0.5cm 4cm 0.55cm},clip,width=\textwidth]{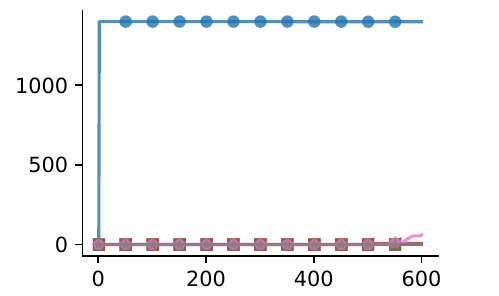}
         \caption{Size 80,000}
     \end{subfigure}
     \begin{subfigure}[b]{0.3\textwidth}
         \centering
         \includegraphics[trim={8.5cm 0 0 0.55cm},clip,width=0.55\textwidth]{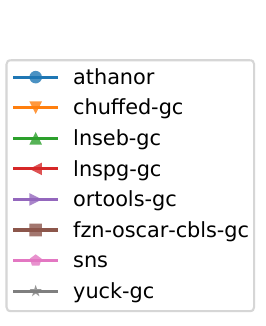}
     \end{subfigure}
\caption{Comparing all solvers \textbf{with the knapsack global constraint model} (except \athanor and SNS) on large instances of the Knapsack problem (with 10,000 to 80,000 objects), using the scoring system described in \Cref{sec:expdetails}. Higher scores indicate better relative performance.}
\label{fig:experiments-large-knap-gc}
\end{figure}

Knapsack is represented as a set of integers in \essence, so \athanor can take advantage of the variable-sized compound type and corresponding variable-sized expressions in the constraint and the objective function. As we scale up the number of objects to 80,000, \athanor scales better than all other solvers from size 60,000 upwards, as shown in \Cref{fig:experiments-large-knap-gc}. 

The knapsack global constraint enables the LNS solvers to perform very well at size 10,000, much better (relative to other solvers) than in \Cref{sec:experiments}. However as the size is increased the LNS solvers struggle to find a feasible solution. For instances of size 80,000, all solvers other than \athanor and SNS have a score of 0 after 600 seconds have elapsed. In contrast, in \Cref{sec:experiments} we reported that the LNS solvers perform well, even up to size 80,000. It seems the overhead of propagating the knapsack constraint is not worthwhile for the largest instances. 

Finally, several other solvers outperformed \athanor on the TSP when using the circuit constraint. There is no fundamental reason to expect \athanor to scale better than the other solvers. The \essence specification for TSP (\Cref{fig:tsp-spec}) does not contain any decision variables with variable-sized compound types: the only decision variable is a fixed-length sequence. Also, it does not contain any expressions that unroll dynamically (either in a constraint or the objective function). Therefore we have not performed a scalability experiment for TSP. The success of Yuck on the TSP suggests there may be neighbourhood structures or local search techniques that could be added to \athanor to improve its performance on the TSP and related problems. 

\subsection{Summary}

\athanor compares well to both systematic and local search solvers, even when the other solvers are using \textit{specialised} global constraints (which we define as global constraints that encapsulate all or a very substantial part of the problem). This is a remarkable result, given that the implementation of a specialised global constraint can include arbitrary problem-class-specific reasoning. 

The most acute examples are the Knapsack and Bin Packing problems. The knapsack global constraint encapsulates the entire Knapsack problem, while the bin packing constraint includes almost the entire problem. They both have sophisticated propagators for systematic CP solvers~\cite{katriel2007propagating,shaw2004constraint}. Even so, \athanor scales well and outperforms all other solvers on the largest instances of Knapsack and Bin Packing. On the TSP, \athanor is competitive but is not the leading solver, however for CVRP (another tour problem that uses the circuit global constraint) \athanor outperforms all other solvers. 

\section{Conclusions}\label{sec:conclusion}

In this work, we have proposed \athanor, a general-purpose constraint-based local search solver. Compared to existing constraint solvers, \athanor's novelty lies in its ability to search directly on the high-level description of a problem written in the \essence constraint modelling language. This key idea offers two unique advantages. Firstly, it allows the solver to effectively exploit the (high-level) structure of a given problem and automatically derive a rich set of neighbourhood structures for the search. Secondly, by working directly with variable-sized data structures and the use of dynamic memory allocation, \athanor is able to scale gracefully with the size of the given problem. The twin benefits of the proposed approach result in an effective and highly scalable solver, as demonstrated in our experiments across seven combinatorial optimisation problems. 
We compared \athanor with general purpose local search solvers (Choco LNS, \fznoscar, Yuck, and SNS) and with systematic solvers that have conflict learning (Chuffed and OR-Tools). The experimental results are complex, but in summary \athanor shows a clear advantage for five of the seven problem classes. The two exceptions are the Travelling Salesperson Problem (TSP) and the Progressive Party Problem (PPP). The \essence specification of TSP does not have favourable characteristics for \athanor (such as a nested domain, or a domain with a variable-sized compound type). In this case, \athanor is outperformed by solvers using a specialised global constraint.
The limitation in performance on PPP can be attributed to the simplicity of the neighbourhood structure selection mechanism inside the solver, which does not allow the search to recognise effective combinations of neighbourhood structures that need to be applied in sequence. 

There are several directions to expand on this work. 
The design of \athanor provides the opportunity to investigate other ways of designing neighbourhoods and neighbourhood structures, taking advantage of both the high-level structure and recursive nature of variables. Our experiments show that our current domains can already compete with other state of the art solvers, but we believe this area of research could lead to significant improvements over our current system.
Firstly, as described above, the employment of more sophisticated neighbourhood structure selection mechanisms can potentially help the solver to overcome the limitation in performance found in our experiments on PPP or similar problems, where the effectiveness of a neighbourhood structure can only be observed if it is used in combination with others. This topic has been intensively studied by the hyper-heuristic community (see, e.g. \cite{misir2012intelligent,kheiri2015sequence}). In addition to learning to select the best neighbourhood combinations on-the-fly, a recently emerging family of techniques is to learn selection strategies in a data-driven fashion via the use of deep reinforcement learning~\cite{yi2022automated}. Similar approaches have shown promising results in different areas, such as heuristic selection in planning solvers~\cite{speck2021learning}. 
A second direction for future research is on expanding on the neighbourhood templates. By investigating the effectiveness of various problem-specific neighbourhood templates commonly used in the metaheuristics community (e.g., \cite{sorensen2008multiple}) and integrating them into the neighbourhood template library of \athanor, we can potentially enhance the solver's ability in exploiting the inherent structure present in combinatorial optimisation problems. This may lead to significant improvement in the solver's performance. However, it is important to note that such expansion will also make the neighbourhood structure selection problem much more challenging, which will likely require developing more effective learning techniques. 
Finally, there are several design choices in \athanor that are currently set in an ad-hoc manner. A thorough investigation of those choices and their impact on the solver's performance, together with the development of effective algorithm configuration techniques~\cite{stutzle2019automated,biedenkapp2020dynamic,adriaensen2022automated} to automate such design choices are other important avenues for future work. 

\subsubsection*{Acknowledgements}

The experiments made use of Cirrus, a UK National Tier-2 HPC Service at EPCC (\url{http://www.cirrus.ac.uk}) funded by the University of Edinburgh and EPSRC (EP/P020267/1). Ian Miguel is funded by EPSRC grant EP/V027182/1, and Peter Nightingale is funded by EPSRC grant EP/W001977/1. Christopher Jefferson was funded by a Royal Society University Research Fellowship and Nguyen Dang was funded by a Leverhulme Early Career Fellowship during the time this work was conducted. We also thank Håkan Kjellerstrand for constraint models that were used in this paper.

\bibliography{bib}

\newpage

\appendix

\section{Example of triggers and incremental evaluation with the \lstinline|sum| operator}
\label{sec:trigger_sum_operator}

The \lstinline|sum| operator in \athanor has one operand, which must be a sequence of the integer nodes which are to be added together. Like the \lstinline|set| type, the sequence type makes use of an extended set of trigger notifications:

\begin{itemize}
\item \lstinline|valueAdded(index, x)|: \lstinline|x| has been added to the sequence at position \lstinline|index|, moving all later elements up one index.
\item \lstinline|valueRemoved(index,x)|: \lstinline|x|, which was at position \lstinline|index|, has been removed, moving all later elements down one index.
\item \lstinline|subsequenceChanged(startIndex,endIndex)|: One or more of the elements in the sequence between positions \lstinline|startIndex| (inclusive) and \lstinline|endIndex| (exclusive) have a new value.
\item \lstinline|positionsSwapped(index1,index2)|: elements at positions \lstinline|index1| and \lstinline|index2| have swapped positions.
\item \lstinline|memberHasBecomeUndefined(index)|: The value of the element at position \lstinline|index| has become undefined.
\item \lstinline|memberHasBecomeDefined(index)|: The value of the element at position \lstinline|index| has become defined.
\end{itemize}

The following example shows how the \lstinline|sum| operator is incrementally updated using the sequence notifications.  We assume that the \lstinline|sum| operator has already been fully evaluated.  The \lstinline|sum| operator is represented by the following state:

\begin{itemize}
\item $\mathit{value}$: The result of summing the integers in the sequence.
\item $\mathit{cmv}$ (cached member values):  a copy of the values in the sequence. Values currently undefined are stored as $0$.
\item $\mathit{undefined}$: The number of undefined values in the sequence.
\end{itemize}

The sum is then incrementally updated as changes arrive. The \emt|sum| implementation accepts the \emt|valueAdded|, \emt|valueRemoved|, \emt|subsequenceChanged| and \emt|positionsSwapped| triggers, and incrementally updates both the $\mathit{value}$ and $\mathit{cmv}$. When \emt|memberHasBecomeUndefined| is received, the $\mathit{value}$ is still kept up to date (except for the undefined value, which is treated as $0$), and $\mathit{undefined}$ is incremented. While $\mathit{undefined}>0$, the \emt|sum| node returns undefined as its value. \emt|memberHasBecomeDefined| decrements $\mathit{undefined}$, and when it reaches $0$ the \emt|sum| node returns $\mathit{value}$.

The only time a full re-evaluation of \textit{value}, \textit{cmv}, and \textit{undefined} are performed is when \emt|valueChanged|, \emt|hasBecomeUndefined| or \emt|hasBecomeDefined| are triggered, as these do not provide any incremental information.

\section{Value Representation and Incremental Hashing}\label{sec:appendix-hashing}

In this section we describe the internal representation and the hash function of each type supported by \athanor.
A hash function is a function that takes a string of arbitrary length and maps it to a \emph{hash} -- a fixed length string.
In \athanor, hash functions are used to quickly compare data structures by comparing their hashes rather than by recursively comparing the entire data structure. Comparing by hash can improve performance by orders of magnitude. However, hash functions have the limitation that multiple input values can have the same hash value -- this is termed a \textit{hash collision}. In general the utility of a hash function depends on the extent to which it satisfies multiple properties~\cite{Preneel2011}, but here we are only concerned with \textit{collision resistance}. Collision resistance is the difficulty of finding two strings $X$ and $Y$ such that $X \neq Y$ and $\hash{X} = \hash{Y}$. We need strong collision resistance in \athanor because hash values are used to compare values for equality: to maintain type correctness (with sets, injective functions, and partitions among others); and in several constraint types (\emt|allDifferent| and \emt|subset| among others). In this section we define a hash function for the types supported by \athanor{}, describe how it can be computed incrementally (i.e.\ updated efficiently when a change is made to a data structure), and discuss the consequences of a hash collision.

In the following subsections we define a hash function \hash{v}, where \(v\) is an expression of any supported \essence type, and show how the hash function is updated incrementally following a change \(\delta\). The hash function produces a 64-bit value. We make use of an existing hash function named \mixF, which has the property that two similar input strings have (with high probability) widely different hashes. MurmurHash3~\cite{murmurhash3} (128-bit) is used for the \mixF function, and its output is reduced to 64 bits with XOR.
The concrete types are dealt with simply. When \(v\) is an integer, \(\hash{v}=v\) and when \(v\) is Boolean, \(\hash{v}=toInt(v)\) (i.e.\ 0 or 1). Enumerated types are represented internally with integers and the integer hash function is used.

\subsection{Sets and Multisets} \label{sec:athanor-hashing-sets}

Sets are represented internally using an extensible array of references to the elements (in no particular order), and a hash set (i.e.\ a hash table containing only keys rather than key-value pairs) of the elements to enable fast membership tests. Insertion and deletion occur at the end of the array; an element to be deleted must be swapped with the last element prior to deletion. For a multiset, the hash set is replaced with a hash table mapping each element to the number of occurrences of that element.

We take the approach of hashing each element of the set or multiset, then combining the hashes of the elements into a hash for the container. The hashes are combined with a commutative function because sets and multisets are unordered containers and may be stored in any order. We use an associative and commutative binary operator $\mu$ to combine the hashes of the elements. This also allows for incremental updates to a hash.  When adding an element $i$ to the set $s$ that has a cached hash value $h$, the new hash $h'$ can be calculated as $h' \gets \mu\left(h,\hash{i}\right)$.  If an element $i_1$ is removed, the new hash $h'$ is calculated as $h' \gets \mu\left(h,(\mathrm{hash}(i_1))^{-1}\right)$, where \(x^{-1}\) is defined such that it will ``cancel out'' \(x\) in the hash.
The choice of the combining operator $\mu$ is important.  Consider a simple choice of addition:

\[
\hash{s} = \RangeExpand{\hash{s_1}}{\hash{s_2}}{\hash{s_n}}{+}
\]

Addition produces a hash function with a high probability of hash collisions.  As \athanor hashes the integer $i$ to $i$, the sets $\Set{2,8}$, $\Set{4,6}$ and  $\Set{1,2,3,4}$  would all hash to $10$. Our solution to this problem is based on Clarke et al.\ \cite{Clarke2003}: the hash of each element is hashed again with a function called \mixF{}, so that similar elements no longer have similar hashes. Therefore, $\hash{s}$ where $s$ is a set of elements $\RangeExpand{s_1}{s_2}{s_n}{,}$ is defined as:

\[
\hash{s} = \RangeExpand{\mix{\hash{s_1}}}{\mix{\hash{s_2}}}{\mix{\hash{s_n}}}{+}
\]

The hash function for multisets is identical to the one for sets.
Note that compared to a set, a multi-set allows multiple occurrences of the same element. As the hashes are added (after mixing), this can reduce the strength of the hash -- as we use 64-bit hashes, then all multi-sets which contain \(2^{64}\) identical values will hash to 0. However, memory constraints make this practically infeasible -- even if a multiset had $2^{32}$ occurrences of the same value, the hash value would still be able to encode $2^{64 - 32} = 2^{32}$ distinct values.

\subsection{Sequences and Tuples}

Sequences are represented with an extensible array of references to elements. In this case the order is preserved when adding or deleting elements at any index of the array.  Sequence hashing reuses the method for incrementally hashing sets by treating the sequence as a set of tuples $(i,j)$, where $j$ is a member in the sequence and $i$ is its index (position) in the sequence.  Assuming we have a method for hashing tuples (see below), we can calculate a new hash $h'$ from an existing hash $h$ for the following operations:

\begin{itemize}
\item Element $e$ added to end of sequence with index $i$:  $h' \gets h + \mix{\hash{(i,e)}}$
\item Element $e$ removed from end of sequence, its index was  $i$:  $h' \gets h - \mix{\hash{(i,e)}}$
\item Element $e$ with index $i$ changed for element $e'$:  $h' \gets h - \mix{\hash{(i,e)}} + \mix{\hash{(i,e')}}$
\end{itemize}

The hashing method for tuples is very simple and does not allow for incremental updates. The hashes of the members of the tuple are concatenated into a sequence, and the sequence is then hashed with MurmurHash3 (used as in the \mixF{} function). We have found that tuples in \essence are usually short, and the lack of incremental updates does not cause a performance problem in practice.  If necessary, the method of incrementally hashing sequences can also be applied to tuples.

\subsection{Functions}

A function consists of a defined set, a range set, and a mapping from elements of the first to elements of the second. The range is stored as an extensible array of references to elements. If the function is total and the defined set is a contiguous set of integers (or tuples of integers, each drawn from a contiguous set), then the defined set is not stored; instead indexes into the range array are calculated directly from the defined values. Otherwise, the defined set is stored in an extensible array and a hash table stores the mapping from defined values to the index of their image in the range array.
Functions are treated as a set of tuples (preimage, image) for the purpose of hashing. As a consequence, functions use the method for incrementally hashing sets.

\subsection{Partitions}

Given a partition with $n$ elements, the partition representation maintains an array of partition elements $e$ and an array of integers $p$ used to map each element $e[i]$ to a label $p[i]$. Those with the same integer label are in the same part (cell) of the partition.
For incremental hashing, a partition is treated as a set of sets.
In order to maintain the hash of each of the parts  (each of the inner sets in our set of sets), we use a third array $h_p$ of hashes:

\[
\forall i \in I. \quad
h_p[i] = \sum\limits_{j \in I, p[j] = i}{\mix{\hash{e[j]}}}
\]

Since we are treating each part of a partition as an individual set, we hash the part in the same way as a set.
We then define the hash of the entire partition similarly, treating the partition as a set of sets. Since in the previous step we have defined the hash of each part (or each inner set of our set of sets) we can apply the same mix/sum operations to combine the hashes, as follows.

\[
\hash{\mathrm{partition}} = \sum \limits_{i \in I}{\mix{h_p[i]}}
\]

When an element is moved from one part to another, the hash of the partition is updated incrementally by removing the hashes of both parts (\(p_1\) and \(p_2\)) from the hash of the partition, incrementally updating the hashes of both \(p_1\) and \(p_2\), then restoring the updated hashes back into the partition hash. The update is performed in constant time assuming that the hash and mix functions are constant time.

\subsection{Hash Collisions}

The path taken through the search space by \athanor{} can, in extremely rare cases, be affected by a hash collision. This can cause incorrect solutions, so all solutions must be verified before they are returned to the user. \athanor{} includes a function to verify that solutions are correct. This function has only ever found incorrect solutions when we purposefully reduce the size of the hash to 8 bits, to check it behaves correctly.
The problem arises when a constraint requires that two elements of a compound type are equal, or (in a negated context) that two elements of a compound type are different. The elements are checked for equality or disequality by comparing their hashes -- this greatly improves performance, but can lead to incorrect solutions if a hash collision occurs. 

\end{document}